\documentclass[lettersize,journal]{IEEEtran}
\usepackage{amsmath,amsfonts}
\usepackage{algorithmic}
\usepackage{algorithm}
\usepackage{array}
\usepackage{textcomp}
\usepackage{stfloats}
\usepackage{url}
\usepackage{verbatim}
\usepackage{graphicx}
\usepackage{caption}
\usepackage{subcaption}
\usepackage{cite}
\usepackage{orcidlink}
\usepackage{hyperref}
\usepackage{array,multirow}
\usepackage[nameinlink,noabbrev]{cleveref}
\usepackage{xcolor}
\usepackage{mathtools}
\usepackage{dsfont}
\usepackage{booktabs}
\usepackage{threeparttable}
\usepackage{setspace}
\usepackage{textcomp}
\usepackage{stfloats}
\usepackage{url}
\usepackage{verbatim}

\begin{document}

\title{VarCoNet: A variability-aware self-supervised framework for functional connectome extraction from resting-state fMRI}

\author{Charalampos Lamprou \orcidlink{https://orcid.org/0000-0002-6247-8689}, Aamna Alshehhi \orcidlink{https://orcid.org/0000-0003-1868-1003}, Leontios J. Hadjileontiadis \orcidlink{https://orcid.org/0000-0002-9932-9302}, and Mohamed L. Seghier \orcidlink{https://orcid.org/0000-0002-1146-8800}
\thanks{This paper was supported by Khalifa University of Science and Technology.}
\thanks{Manuscript received \today}
\thanks{C. Lamprou, L. J. Hadjileontiadis, M. L. Seghier, and A. Alshehhi are  with the Department of Biomedical Engineering and Biotechnology, and the Health Engineering Innovation Group (HEIG), Khalifa University, P.O. Box 127788 Abu Dhabi, UAE (e-mail: \{charalampos.lamprou, aamna.alshehhi, leontios.hadjileontiadis, mohamed.seghier \}@ku.ac.ae).}
\thanks{L. J. Hadjileontiadis is also with 
the Department of Electrical and Computer Engineering, Aristotle University of Thessaloniki, GR 54124, Thessaloniki, Greece (e-mail: leontios@auth.gr).}
}

\maketitle

\begin{abstract}
Accounting for inter-individual variability in brain function is key to precision medicine. Here, by considering functional inter-individual variability as meaningful data rather than noise, we introduce VarCoNet, an enhanced self-supervised framework for robust functional connectome (FC) extraction from resting-state fMRI (rs-fMRI) data. VarCoNet employs self-supervised contrastive learning to exploit inherent functional inter-individual variability, serving as a brain function encoder that generates FC embeddings readily applicable to downstream tasks even in the absence of labeled data. Contrastive learning is facilitated by a novel augmentation strategy based on segmenting rs-fMRI signals. At its core, VarCoNet integrates a 1D-CNN-Transformer encoder for advanced time-series processing, enhanced with a robust Bayesian hyperparameter optimization. Our VarCoNet framework is evaluated on two downstream tasks: (i) subject fingerprinting, using rs-fMRI data from the Human Connectome Project, and (ii) autism spectrum disorder (ASD) classification, using rs-fMRI data from the ABIDE I and ABIDE II datasets. Using different brain parcellations, our extensive testing against state-of-the-art methods, including 13 deep learning methods, demonstrates VarCoNet’s superiority, robustness, interpretability, and generalizability. Overall, VarCoNet provides a versatile and robust framework for FC analysis in rs-fMRI.
\end{abstract}


\section{Introduction} \label{intro}
Each brain exhibits distinct anatomical and functional traits shaped by genetic and environmental factors \cite{gu2014contributes,kosslyn2002bridging}. Consequently, brain structure and function vary across individuals, even among those with similar demographics (e.g., age, gender, handedness) \cite{seghier2018interpreting}. Functional inter-individual variability is depicted as differences across subjects in regional brain activations or inter-regional functional connectivity \cite{vogel2004neural,miller2012individual}, the latter typically  represented through functional connectomes (FCs) derived from resting-state fMRI (rs-fMRI) data. Such variability holds useful information about the individual's traits that should be accounted for when studying brain function. Indeed, understanding variability is key to advancing precision medicine  \cite{seghier2018interpreting}. However, popular FC analysis frameworks rarely consider variability across subjects as a source of meaningful information about brain function in group studies.

Functional variability can predict task performance at the individual level \cite{mennes2010inter,tavor2016task}, showing highly reliability over time \cite{miller2002extensive}. On a practical level, the capability to detect and characterize differences among individuals depends on the relative contribution of two primary components or sources: within-subject variation and inter-subject variation \cite{rex}. Within-subject variation pertains to the variability in brain function observed within the same individual at different times (e.g. in multi-session protocols and test-retest studies). The human brain is a dynamic system, continuously altering its functions in response to internal and external stimuli. Nevertheless, despite the constant changes, there is an equilibrium, maintained through homeostatic processes. This equilibrium enables the brain to preserve order despite its dynamic nature and allows brain measures to approach consistency over time \cite{laumann2017stability,poldrack2015long}. In other words, although the brain function of a subject may vary over time, it maintains specific patterns, allowing test results that are temporally independent to be similar, even though they are never identical \cite{tavor2016task}. Therefore, to achieve reproducibility and reliability, it is crucial to obtain representations from neurophysiological data that effectively describe this equilibrium and are robust to variations in brain function. To improve reliability, within- to between-subject variation ratio has to be kept small over repeated measurements.

\begin{figure*}[t!]
    \centering
    \includegraphics[scale = 0.58]{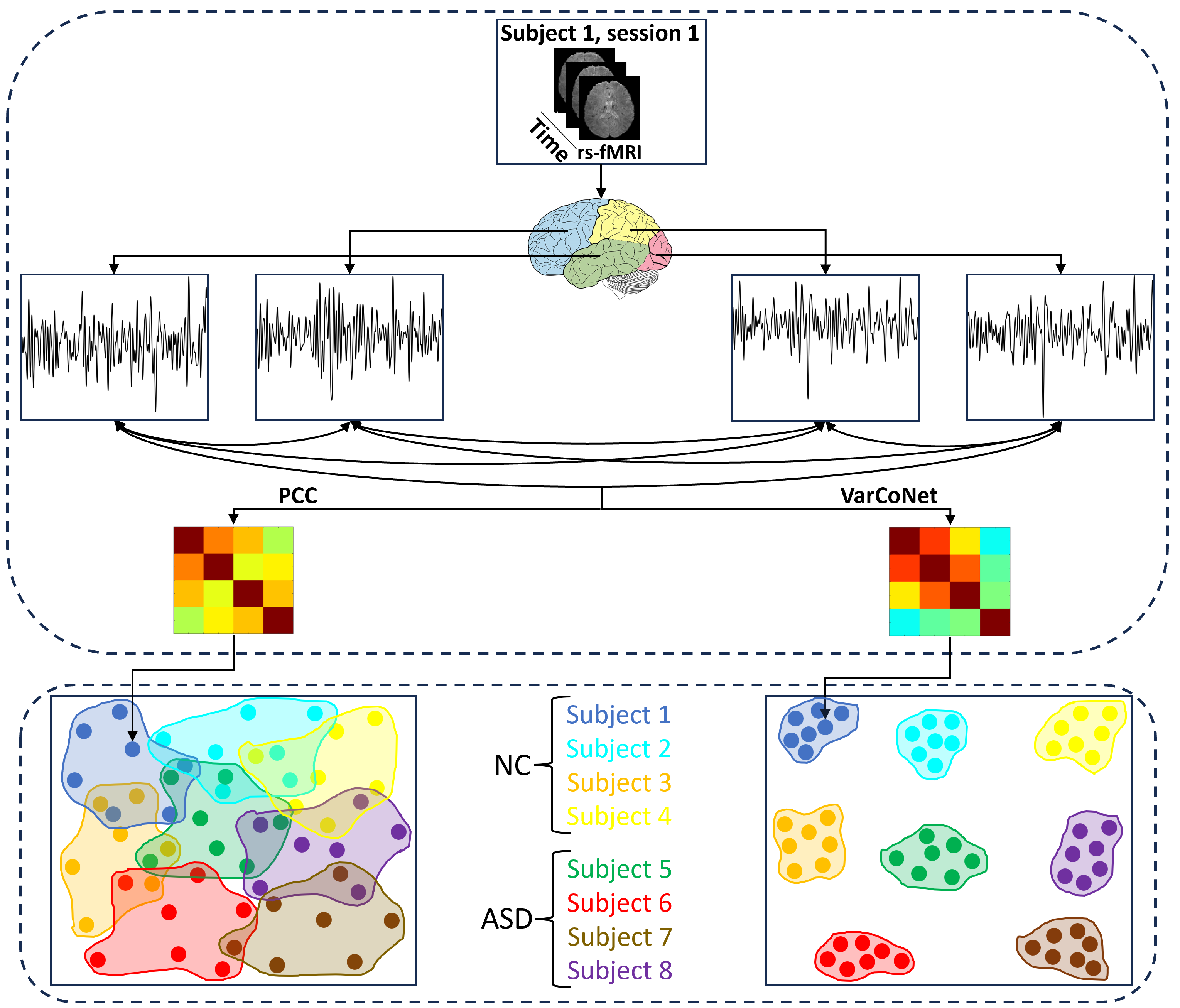}
     \hfill
    \caption{A hypothetical example illustrating the importance of reducing the ratio between intra- to inter-individual variability. The upper panel shows the computation of FCs from rs-fMRI data (subject 1, session 1) using PCC and VarCoNet, including parcellation, time-series extraction, and FC calculation. With a 4-region atlas, each FC yields six distinct values ($(4\times(4-1))/2$), which can be represented as a point in a 6-D scatter plot. For simplicity, FCs are shown as points in a reduced 2-D space. The lower panel compares scatter plots of PCC-based (left) and VarCoNet-based (right) FCs. Colors denote subjects, while multiple points per color indicate sessions. Shaded areas reflect intra-individual variability; distances between shaded areas reflect inter-individual variability. Assuming subjects 1–4 are neurotypicals and subjects 5–8 have ASD, the figure highlights how accounting for variability facilitates clearer decision boundaries.}
    \label{intra-subject_var}
\end{figure*}

Indeed, within-subject variation determines how much of the observed inter-subject variation can be explained \cite{rex} and directly impacts analyses such as inter-individual variability characterization, brain function assessment, brain network mapping, and brain disorder detection. To illustrate the impact of intra- and inter-subject variation on brain disorder classification accuracy, a key aspect of precision medicine, we present a hypothetical scenario with eight subjects: four neurotypical controls (NC) and four with Autism Spectrum Disorder (ASD) (see \autoref{intra-subject_var}). The upper panel shows FC computation for Session 1 of Subject 1 using a 4-region atlas, yielding four time series. FCs are derived with Pearson’s correlation coefficient (PCC) and VarCoNet, yielding $4\times4$ symmetric FC matrices. FCs are estimated for each subject and session. Then, FCs are projected on a 2D-scatter plot for visualisation purporses (lower panel of \autoref{intra-subject_var}: PCC (left) and VarCoNet (right). Each point on the 2D-scatter plot denotes one session, and each subject is shown in a different color. Shaded areas capture intra-subject variability; distances between them reflect inter-subject variability. The PCC scatter plot shows high intra-subject and low inter-subject separation, while VarCoNet shows the opposite pattern. Because most brain disorder datasets include only one session per subject, classification effectively corresponds to selecting one random point per subject. In the PCC case, this leads to unstable and inconsistent decision results, since different sessions (points of the same color) would result in significantly different decision boundaries. In contrast, in the VarCoNet case, boundaries remain stable across selections. This simplified hypothetical example shows how reducing the intra- to inter-individual variability ratio improves classification reliability in brain disorders.

However, collecting enough rs-fMRI data to estimate intra-subject variation across several sessions is  impractical. An alternative framework that only requires two rs-fMRI sessions is subject fingerprinting, which can provide a reliable estimate of intra-subject variation in FC \cite{finn2015functional}. More specifically, in subject fingerprinting, identification is performed using pairs of FC features derived from rs-fMRI scans collected on two different visits (e.g. Time 1 and Time 2). Each subject’s feature from Time 1 is then compared against all subjects' features from Time 2. If the highest similarity is found between the same subject’s features, the identification is considered correct. The identification rate is then computed as the ratio of correctly identified subjects to the total number of subjects. Since low intra-subject variation is crucial for high identification rates, subject fingerprinting serves as a robust method for evaluating the stability of brain function encoding techniques in the presence of intra-subject variability in multi-session multi-subject rs-fMRI studies. Similarly, improving FC-based brain disorder classification requires better characterization of functional inter-individual variability while minimizing intra-subject variability in test-retest scenarios. However, for both applications, i.e. subject fingerprinting and brain disorder classification, FC estimation must be robust to differences in signal length, in particular when processing unlabeled rs-fMRI datasets with variable durations \cite{birn2013effect}.       

In that context, we present VarCoNet, a novel deep learning (DL) framework designed to enhance FC estimation. Following our recent prior work \cite{lamprou2025Improved}, VarCoNet leverages contrastive self-supervised learning (SSL) to train a combined 1D-CNN–Transformer encoder that processes rs-fMRI time-series without requiring labeled data. Rather than computing FCs directly from raw time-series, VarCoNet derives FCs by assessing the similarity between the learned embeddings. To enable contrastive learning, augmented views are generated by segmenting the input signals. For each subject in a training batch, the model increases the similarity between FCs computed from two different segments of that subject’s data, while decreasing similarity with FCs from other subjects. This core mechanism of contrastive learning naturally aligns with the key principles of inter-individual variability analysis—namely, reducing intra-subject variation while maintaining higher inter-subject variation.

The main novelty of our work lies in the optimal design and implementation of all VarCoNet modules, including in particular the optimization and augmentation processes. Specifically, VarCoNet architecture and training parameters were optimized using a Bayesian optimization scheme, with an objective function that jointly maximized fingerprinting accuracy and minimized sensitivity to signal duration. Likewise, the augmentation process involves positive samples created by segmenting time-series and negative samples by contrasting segments from different subjects within a batch. Compared to previous studies employing similar strategies \cite{wang2023unsupervised}, VarCoNet’s augmentation introduces three key improvements: (1) segment lengths are sampled from a continuous range of physiologically plausible values, rather than fixed lengths, improving robustness to signal duration; (2) the minimum and maximum lengths are carefully chosen to ensure sufficient diversity without causing large fluctuations in fingerprinting accuracy; and (3) segmentation windows are placed at random positions rather than fixed ones, promoting more diverse training samples and better exploiting the hybrid 1D-CNN–Transformer architecture given the dynamic nature of FC.

The performance of VarCoNet is evaluated on two key tasks: subject fingerprinting, to assess its ability to enhance inter-individual variability characterization using rs-fMRI data from the Human Connectome Project (HCP), and ASD classification using rs-fMRI data from the Autism Brain Imaging Data Exchange (ABIDE) I and II, where ABIDE II is used exclusively for external testing. Performance is benchmarked against state-of-the-art brain activity encoding methods. Subject fingerprinting is evaluated across a range of signal lengths commonly found in multi-site rs-fMRI datasets. ASD classification is assessed using repeated K-fold cross-validation and external testing, ensuring a robust and generalizable evaluation. Our findings show that VarCoNet consistently outperforms baseline methods on both tasks. Moreover, explanatory analyses identify key functional connections associated with ASD detection, providing interpretability into the model’s learned representations and further supporting its clinical relevance.

\section{Related work} \label{related}

\subsection{Subject fingerprinting}
The early work of Finn et al. \cite{finn2015functional} demonstrated that rs-fMRI FC, computed using Pearson’s correlation coefficient (PCC), is reliable for subject fingerprinting, which provided a practical demonstration that functional inter-individual variability contains meaningful information about individual's brain function. Since then, researchers have sought to enhance functional subject fingerprinting through novel connectivity metrics \cite{jin2019extracting}, dimensionality reduction techniques \cite{abbas2020geff}, and more recently, DL \cite{cai2021functional,lu2024brain,lee2024discovering}.

To eliminate indirect effects in connectivity matrices, Jin \textit{et al.} \cite{jin2019extracting} proposed direct FC (dFC), an alternative to conventional PCC-based FC, computed using the silencing method \cite{barzel2013network}, which suppresses indirect interactions.

Aiming to enhance subject-specific representation, Abbas \textit{et al.} \cite{abbas2020geff} introduced GEFF, a graph embedding technique that applies PCA to vectorized PCC-based FCs, projecting them into an eigenspace where fingerprinting is performed by comparing new FCs from the same subjects.

To capture individual variability from fMRI time-series, Cai \textit{et al.} \cite{cai2021functional} designed a fingerprinting pipeline involving an autoencoder that reconstructs the input signal; residuals obtained by subtracting reconstructions from originals are then used to compute PCC-based FCs, which are refined through sparse dictionary learning using the K-SVD algorithm \cite{aharon2006ksvd}.

In a similar approach, Lu \textit{et al.} \cite{lu2024brain} applied a variational autoencoder (VAE) to PCC-based FCs and extracted residual FCs for further refinement using the K-SVD-based sparse dictionary learning module \cite{aharon2006ksvd}.

Finally, to explore the robustness of PCC-based FCs over extremely short rs-fMRI segments, Lee \textit{et al.} \cite{lee2024discovering} trained a multi-layer perceptron (MLP) on vectorized FCs, including segments as short as 15 seconds.

Although all the aforementioned methods are designed for subject fingerprinting, they exhibit key methodological differences. For instance, the approaches in \cite{abbas2020geff,lu2024brain,lee2024discovering} process PCC-based FCs, while the one in \cite{cai2021functional} processes parcellated fMRI time-series. Another distinction concerns how these methods perform fingerprinting. Based on their fingerprinting approach, they can be categorized into three groups: (1) Purely mathematical methods that do not require training, such as PCC-based FCs \cite{finn2015functional} and dFC \cite{jin2019extracting}, (2) trainable methods that, once trained, can generalize to fingerprinting new subjects without requiring additional fMRI data from them \cite{cai2021functional,lu2024brain}, and (3) data-dependent methods that require at least one fMRI run from test subjects to be included in the training set \cite{abbas2020geff,lee2024discovering}. Because the latter category requires several fMRI runs, methods that fall in that category were excluded as a benchmark in this study.

\subsection{FC-based classification and other downstream tasks}
There is a growing interest for DL models to process fMRI data for various downstream tasks, including brain disorder classification, gender classification, brain state detection, and age prediction. 

To enhance classification performance by refining extracted features from parcellated rs-fMRI, Riaz \textit{et al.} \cite{riaz2020deepfmri} proposed DeepFMRI, a DL model that combines a 1D-CNN-based feature extractor with a functional connectivity network prior to classification. The model demonstrated promising results on attention deficit hyperactivity disorder (ADHD) detection.

To effectively capture spatiotemporal dependencies in volumetric fMRI data, Nguyen \textit{et al.} \cite{nguyen2020attend} introduced BAnD, a DL framework integrating a 3D-ResNet-18 that processes volumetric fMRI data, summarizing information and extracting meaningful time-series. These time-series are then concatenated with a classification token and passed through a transformer encoder. Finally, the pooled classification token is used for task classification. BAnD was trained and tested on task state decoding yielding promising results.

Seeking an end-to-end architecture for rs-fMRI encoding and FC-based analysis, Kan \textit{et al.} \cite{kan2022fbnetgen} introduced FBNetGen, which integrates a time-series encoder with a graph neural network (GNN) for downstream prediction tasks. In this framework, rs-fMRI time-series are first processed using either a 1D-CNN or a gated recurrent unit (GRU). FC is then computed by taking the dot product between all possible pairs of ROIs. The resulting FC matrix is subsequently fed into a GNN, whose output is used for downstream predictions. FBNetGen was evaluated on gender classification

To scale fMRI encoding with limited supervision, Thomas \textit{et al.} \cite{thomas2022self} applied multiple SSL strategies, including autoencoding, BERT, and causal sequence modeling, to train Transformer models on 11,980 fMRI runs from 34 datasets. When fine-tuned, the pretrained models, especially those using causal sequence modeling (CSM), generalized well to unseen datasets and outperformed models trained from scratch on mental state detection.

Inspired by the success of Transformers across diverse data types, including graphs, Kan \textit{et al.} \cite{kan2022brain} introduced the Brain Network Transformer (BNT) for brain network analysis. In this framework, brain networks are modeled as graphs where nodes are defined by an atlas, node features represent the connectivity profiles of ROIs, and pairwise connection strengths are learned through attention weights. The attention mechanism is designed to capture cross-individual patterns relevant for downstream tasks. BNT was evaluated on ASD classification and demonstrated promising performance.

To reduce dependence on labeled data, Wang \textit{et al.} \cite{wang2023unsupervised} introduced UCGL, an unsupervised contrastive graph learning framework based on SimCLR \cite{chen2020simple}. Two augmented views of each fMRI signal are generated via temporal cropping (first 90\% and last 90\%) and encoded with a spatio-temporal graph convolutional network (ST-GCN). Each view is modeled as a spatio-temporal graph, where nodes represent brain regions at specific time points, temporal edges capture dynamics across time, and spatial edges, weighted by PCC, capture FC. UCGL maximizes agreement between views of the same sample while minimizing agreement across samples. A fine-tuning stage then adds a classification layer and trains the full model with labeled data. The framework was evaluated on ASD, Alzheimer’s disease, and MDD classification.

Another contrastive learning-based method, A-GCL, was proposed by Zhang \textit{et al.} \cite{zhang2023gcl}. It operates directly on PCC-based FCs matrices derived from full-length rs-fMRI. For each subject, a graph is constructed where nodes represent ROIs with features defined by amplitude of low-frequency fluctuations (ALFF) in three frequency bands, and edges are weighted by PCC. A graph augmentation module perturbs the structure by using a graph isomorphism network (GIN) to generate new node features, which drive stochastic edge dropping to form an augmented graph. The model is then trained with a contrastive objective to maximize agreement between original and augmented graphs using SimCLR. After training, embeddings extracted by the GIN are used to train and test a support vector machine classifier. A-GCL was evaluated on ASD and ADHD classification.

To improve local temporal representation learning in fMRI data, Bedel \textit{et al.} \cite{bedel2023bolt} developed BolT. BolT employs a cascade of transformer encoders with a novel fused window attention mechanism. Encoding is performed on temporally overlapping windows within the time-series to capture local representations. BolT was tested on gender detection, task detection, and ASD classification, outperforming several baselines and achieving state-of-the-art performance.

To jointly exploit temporal and connectivity-based information, Meng \textit{et al.} \cite{meng2024cvformer} proposed CvFormer, a Cross-View Transformer architecture that processes rs-fMRI time-series and PCC-based FCs in parallel using separate encoders, followed by cross-view modules for joint feature integration. CvFormer delivered strong results on both AD and ASD classification.

To improve interpretability and performance of rs-fMRI-based brain disorder classification, Zheng \textit{et al.} \cite{zheng2025brainib} proposed BrainIB, a novel GNN framework to analyze PCC-based FCs by leveraging the information bottleneck principle. BrainIB consists of three modules: a subgraph generator which takes as input the original graph and ouputs a reduced sub-graph, a GIN encoder, and a mutual information (MI) estimation module responsible for minimizing the MI between the embeddings of the original reduced graphs, while maximizing the MI between the embeddings of the reduced graph and the labels. BrainIB was evaluated on ASD and MDD classification.

Another SimCLR-based method, GCDA, was proposed by Wang \textit{et al.} \cite{wang2025self}. Each subject is represented as a graph where nodes are ROIs with connectivity-profile features, edges are weighted by PCC, and adjacency is thresholded at 0.3. An augmented view is generated with a graph diffusion augmentation (GDA) module, which perturbs the graph via Markov transition noise and reconstructs it with a Transformer-based denoising network. Original and augmented graphs are then encoded with a GIN and trained using a contrastive objective. A fine-tuning stage then adds a classification layer and trains the full model with labeled data. GCDA was evaluated on ASD and MDD classification.

The aforementioned methods exhibit key methodological differences. For instance, the models in \cite{riaz2020deepfmri,kan2022fbnetgen,bedel2023bolt,meng2024cvformer,thomas2022self} accept as inputs parcellated fMRI time-series, while the model in \cite{nguyen2020attend} are trained on fMRI volumes. Finally, the models in \cite{kan2022brain,wang2023unsupervised,zhang2023gcl,zheng2025brainib,wang2025self} are trained on PCC-based FCs. Furthermore, these models are evaluated on different downstream tasks. However, they are still relevant to the scope of our study for comparison purposes because their primary objective is to improve fMRI encoding, with downstream tasks serving as benchmarks.

\subsection{Research gaps and contributions} \label{research_gaps}
VarCoNet addresses the following key gaps in fMRI encoding.
\begin{itemize}
    \item \textbf{Inefficient handling of variability in acquisition duration:} Existing methods often overlook variability in acquisition duration, a common challenge in multi-site datasets. PCC-based FC approaches discard temporal dynamics \cite{kan2022brain,kan2022brain,zhang2023gcl,zheng2025brainib,wang2025self}, while some signal-level methods crop all recordings to the shortest length \cite{riaz2020deepfmri}, leading to substantial information loss. SimCLR-based frameworks typically split each signal into two fixed halves \cite{wang2023unsupervised}, which fails to capture dataset heterogeneity. VarCoNet addresses this issue with a flexible augmentation strategy where both segment length and position are treated as random variables. The minimum and maximum segment lengths are carefully selected to remain physiologically plausible while reflecting realistic acquisition durations.

    \item \textbf{Limited integration of local and global temporal features:} Existing models utilize 1D-CNNs or gated recurrent units (GRUs) \cite{riaz2020deepfmri,kan2022fbnetgen}, which can be insufficient for capturing both short- and long-range temporal dependencies in fMRI data. VarCoNet addresses this by integrating a 1D-CNN with a Transformer, enabling the model to extract local patterns via convolution and global dependencies via attention.

    \item \textbf{Inefficient attention computation on raw time-series:} Existing methods rely solely on Transformers \cite{nguyen2020attend,thomas2022self,malkiel2022self,bedel2023bolt,meng2024cvformer} for fMRI time-series encoding. However, Transformers are designed for NLP, where words are semantically distinct tokens. In contrast, individual fMRI time points lack inherent meaning. Applying attention directly to raw time-series can therefore be inefficient. VarCoNet resolves this by introducing a 1D-CNN before the Transformer, allowing the model to form more meaningful intermediate representations on which attention can operate effectively.

    \item \textbf{Overfitting to task-specific labels:} The majority of methods for fMRI time-series encoding are based on supervised learning \cite{riaz2020deepfmri,bedel2023bolt,meng2024cvformer,kan2022fbnetgen,nguyen2020attend}. However, supervised models may suffer from “attentive overfitting,” a phenomenon where models overfit to label-specific features and ignore broader patterns \cite{zagoruyko2016paying}. On the other hand, VarCoNet, based on contrastive learning, mitigates this by learning task-agnostic representations. Importantly, the contrastive learning approach used here is physiologically interpretable and theoretically grounded, reinforcing the usefulness of accounting for inter-individual variability for precision medicine.
\end{itemize}

Overall, VarCoNet, as an SSL-based framework, can be easily adapted for diverse tasks operating as a general multi-purpose brain function encoder, extracting meaningful and interpretable representations from rs-fMRI data.

\section{Method} \label{methods}
\begin{figure*}[t!]
    \centering
    \includegraphics[width = \textwidth]{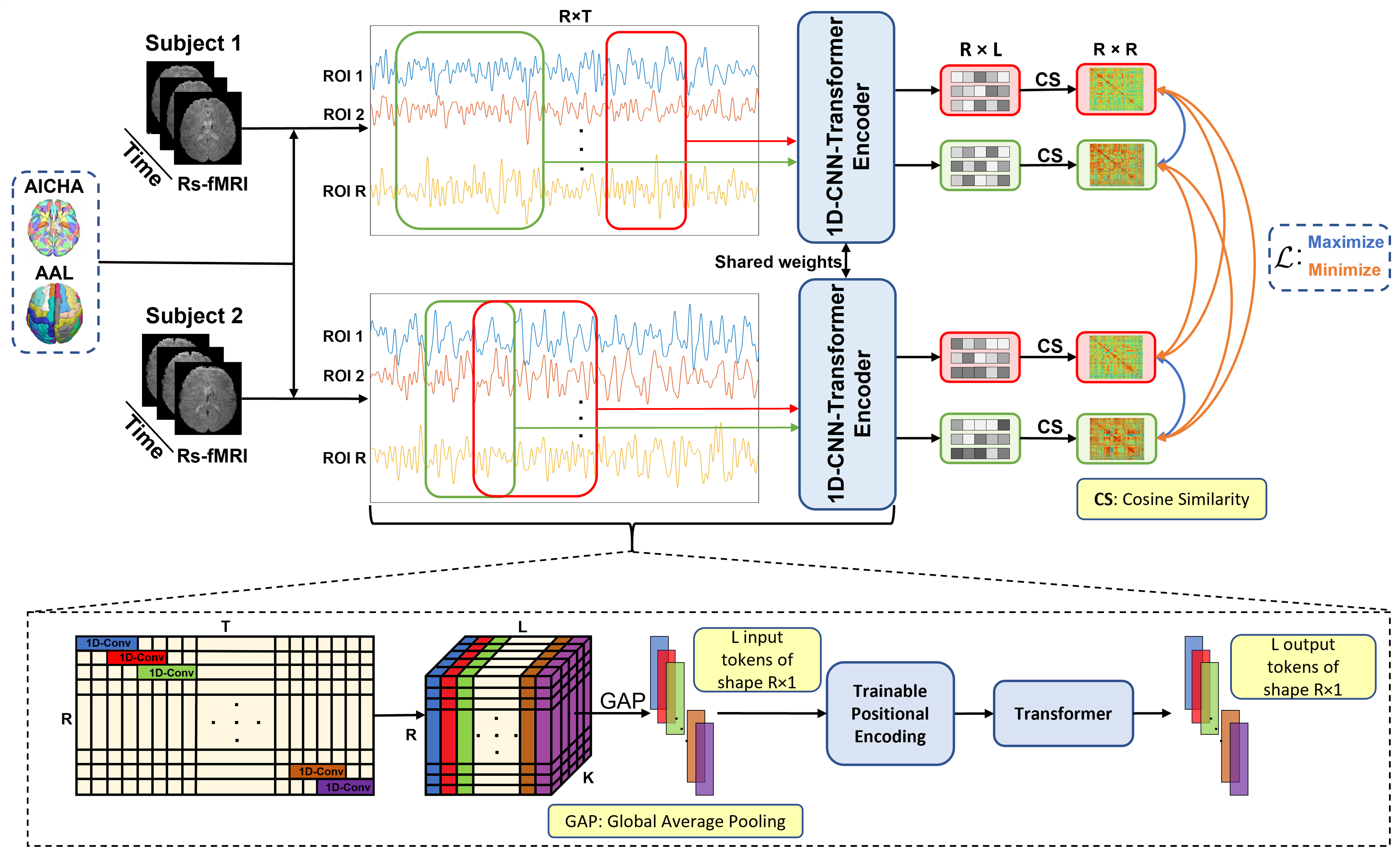}
     \hfill
    \caption{Block diagram of VarCoNet's contrastive training. For illustration purposes, a batch of two subjects is shown. After parcellation using one of two atlases, each rs-fMRI signal is augmented to generate a pair of views. These views are processed by a shared 1D-CNN-Transformer encoder, producing embeddings of size $R \times L$, where $R$ is the number of ROIs and $L$ denotes the number of tokens. Cosine similarity is then computed among the ROIs to obtain the VarCoNet-based FCs as $R \times R$ matrices. These matrices are vectorized by discarding the lower triangular part and undergo a contrastive process that favors similarity between FCs of the same subject while minimizing similarity between FCs of different subjects. The lower part of the figure provides a detailed view of embedding computation. Specifically, the 1D-CNN slides over each ROI’s time-series of length $T$, extracting $L$ tokens of shape $R \times K$, where $K$, the number of kernels. Global average pooling reduces each token to shape $R \times 1$. Positional encodings are then added before the Transformer processes the token sequence.}
    \label{block_diagram}
\end{figure*}

The framework of our proposed VarCoNet is illustrated in \autoref{block_diagram}. Parcellated rs-fMRI time-series are fed into the VarCoNet framework to produce learned FCs (VarCoNet-based FCs) that are subsequently used for subject fingerprinting and classification. 

\subsection{Contrastive learning}
Contrastive-based SSL enables models to capture intricate data structures without relying on labeled data, addressing a key limitation of large-scale supervised learning (SL) models, particularly in medical applications. In this work, contrastive training is implemented using the SimCLR \cite{chen2020simple} framework, as depicted in \autoref{block_diagram}. Given a batch of $N$ samples, two augmented views are generated for each sample. SimCLR optimizes representations by increasing similarity between embeddings of the same sample while reducing similarity among different samples. This is achieved through the following loss function.

Denoting cosine similarity between vectors $\mathbf{u}$ and $\mathbf{v}$ as $sim(\mathbf{u},\mathbf{v}) = \mathbf{u} \cdot \mathbf{v} / |\mathbf{u}| |\mathbf{v}|$, the loss for a positive pair $(i, j)$ is formulated as:
\begin{equation} \label{simclr}
    \ell_{i,j} = -\log \frac{\exp(\text{sim}(\mathbf{z}_i, \mathbf{z}_j) / \tau)}{\sum_{k=1}^{2N} \mathds{1}_{[k \neq i]} \exp(\text{sim}(\mathbf{z}_i, \mathbf{z}_k) / \tau)},
\end{equation}
where $\mathds{1}_{[k \neq i]} \in [0, 1]$ is an indicator function set to 1 if k$\neq$i and zero otherwise, and $\tau$ a temperature parameter. 

\subsection{VarCoNet}
The implementation of VarCoNet consists of three key components: data augmentation, a 1D-CNN-Transformer encoder, and hyperparameter optimization.

\subsubsection{Data augmentation}
An effective augmentation strategy is crucial for successful contrastive learning. It is essential to ensure that the two augmented views of the same sample retain similar information. In VarCoNet, rs-fMRI signals are augmented by segmenting them into multiple segments. This induces the encoder to produce similar embeddings for segments of the same signal while ensuring dissimilarity between embeddings from different signals.

There are different ways to segment the rs-fMRI signals. In our preliminary implementation \cite{lamprou2025Improved} following the work of \cite{peng2022gate}, two segmentation strategies were used: (1) Segments of different lengths ($L_1$ and $L_2$) centered at the same time point, and (2) Segments of equal length ($L$) with randomly selected centers. All segment lengths ($L_1$, $L_2$ and $L$) were chosen from a fixed set of five possible values. Here, our new implementation illustrated in \autoref{block_diagram} adopts a more efficient unified segmentation approach. Specifically, both the segment lengths ($L_1$ and $L_2$) and their centers are randomly selected at each epoch, increasing variation between segments for the encoder while reducing computational complexity. Moreover, the segment lengths $L_1$ and $L_2$ are allowed to take any value within a predefined range [$L_{min}$, $L_{max}$]. These limits are carefully chosen to optimize contrastive learning efficiency, as detailed in Section \ref{impdets}.

\subsubsection{rs-fMRI encoder}
The core component of VarCoNet is its encoder, which combines a 1D-CNN with a Transformer encoder, as illustrated in \autoref{block_diagram}, following a similar rationale to wav2vec \cite{baevski2020wav2vec}, a successful model developed by Facebook AI for audio signal processing. As mentioned above, individual time points in continuous rs-fMRI signals lack semantic meaning, making direct attention computation suboptimal. To address this, the 1D-CNN acts as a feature extractor, sliding across the time-series to produce meaningful tokens that are then processed by the Transformer encoder.   

Given an input $x$ of shape $R \times T$, where $R$ represents the number of ROIs and $T$ denotes the number of time points, an instance normalization layer first normalizes each sample within the batch. The 1D-CNN then processes the time-series of each ROI independently, producing an output of shape $R \times L \times K$, where $K$ is the number of kernels and $L$ is the reduced time dimension due to convolutional operations ($L < T$). Then, global average pooling is applied along the third dimension, yielding a representation of shape $R \times L$. Each of the $L$ columns in this representation serves as a token for the Transformer encoder. Furthermore, trainable positional encodings are added to preserve temporal information. The Transformer processes these inputs and yields output representations of shape $R \times L$. Finally, VarCoNet-based FC matrices of size $R \times R$ are computed using cosine similarity. To eliminate redundancy of information due to symmetry, the lower triangular part of each VarCoNet-based FC matrix is removed, resulting in a final embedding vector of shape $1 \times R(R-1)/2$.

\begin{figure*}[t!]
    \centering
    \includegraphics[width = 0.8\textwidth]{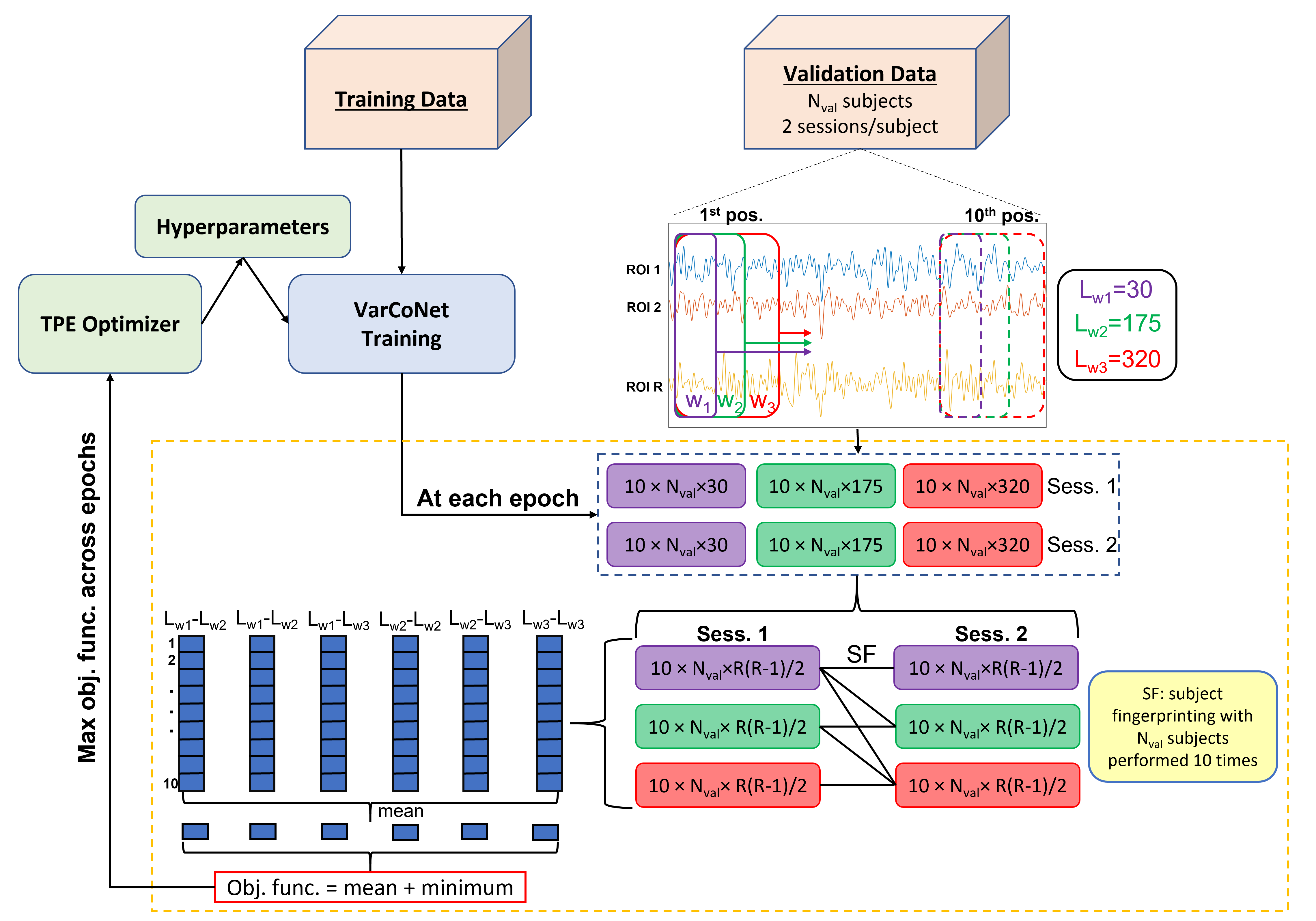}
     \hfill
    \caption{Block diagram of the subject fingerprinting-based objective function used in Bayesian optimization. Given a set of hyperparameters proposed by the optimizer, VarCoNet is initialized and trained on the training data. At each epoch, rs-fMRI signals from the validation set ($N_{val} = 200$ subjects) are segmented using sliding windows of lengths $L_{w1}$, $L_{w2}$, and $L_{w3}$ (10 segments per length). VarCoNet extracts vectorized FC embeddings of shape $1 \times R(R-1)/2$, where $R$ is the number of ROIs. Subject fingerprinting is then performed across all six segment-length combinations ($L_{w1}$-$L_{w1}$, $L_{w1}$-$L_{w2}$, ..., $L_{w3}$-$L_{w3}$) using data from both sessions. Each combination yields 10 fingerprinting scores, which are averaged to obtain six final scores. The objective function is computed as the harmonic mean of the average and minimum of these six scores. The highest objective function value across all training epochs is used to update the optimizer.}
    \label{fing_block_diagram}
\end{figure*}

\subsubsection{Bayesian hyperparameter optimization} \label{hyperparamopt}
Bayesian optimization is a powerful approach for tuning the hyperparameters of DL models \cite{snoek2015scalable}. By efficiently exploring a large hyperparameter space, it enhances model performance while reducing the computational cost of hyperparameter selection. In this study, the Tree-structured Parzen Estimator (TPE) is employed to optimize key hyperparameters of the encoder over 125 trials, including: (1) number of Transformer encoder layers $N_{layers}$, (2) number of attention heads per layer $N_{heads}$, (3) feedforward dimension $FF_{dim}$, (4) batch size (5) temperature parameter $\tau$ (see Equation \ref{simclr}), and (6) learning rate $lr$.

At each iteration of the TPE, the aforementioned hyperparameters are set to specific values. VarCoNet is then trained using those hyperparameters and at each epoch it is evaluated based on subject fingerprinting performance using a validation set of $N_{\text{val}} = 200$ subjects, each with two scanning sessions. From each validation, 30 segments of the rs-fMRI signal of lengths $L_{w1}=30$ (45 s with a TR of 1.5 s), $L_{w2}=175$ (4.375 mins with a TR of 1.5 s), and $L_{w3}=320$ (8 mins with a TR of 1.5 s) samples (time points) are extracted, with 10 segments per length exhibiting minimum overlap (or no overlap for segments of length 30). 

Subject fingerprinting is then performed under six different combinations of segment lengths: (1) 30 samples for session 1 and session 2, (2) 175 samples for session 1 and session 2, (3) 320 samples for session 1 and session 2, (4) 30 samples for session 1 and 175 for session 2, (5) 30 samples for session 1 and 320 for session 2, and (6) 175 samples for session 1 and 320 for session 2. For each combination, fingerprinting rates are computed and averaged across the 10 segments. The optimization objective is defined as the sum of the average and the minimum of these six fingerprinting rates. Since this evaluation is conducted after each epoch, a single optimizer iteration involves as many evaluations as the number of epochs. To update the TPE, the value from the epoch with the best validation score is used. This process, i.e. extracting segments of different lengths from the validation data, obtaining VarCoNet-based FCs, performing intra- and inter-length subject fingerprinting, and using the resulting fingerprinting rates to compute the objective function, is detailed in \autoref{fing_block_diagram}.

The chosen objective function (i.e. the harmonic mean of
the average and minimum performance) is designed to enhance overall performance while ensuring that the lowest performance remains high. This approach enhances the model's robustness across different signal lengths, which is crucial when working with multi-site datasets that might have variable acquisition durations.

\subsection{Implementation details}
\subsubsection{Datasets and preprocessing} \label{datadescription}
\textbf{Human Connectome Project (HCP).} The HCP dataset is a widely-used publicly available dataset comprising rs-fMRI data from 1,113 participants (ages 22–36), collected using a 3T Siemens Connectome-Skyra scanner \cite{van2013wu}. Each subject underwent two scanning sessions on separate days, with each session consisting of two runs: one employing echo-planar-imaging (EPI) with right-to-left phase encoding and the other with left-to-right phase encoding. To avoid session bias, data were randomly selected from both phase encodings. Each rs-fMRI run lasted 14.33 minutes and was acquired using an EPI sequence with a TR of 0.72 s and a TE of 33.1 ms, resulting in 1,200 volumes per run. During scans, participants were instructed to keep their eyes open, fixate on a crosshair, and remain still. The data utilized in this study underwent preprocessing with the HCP minimal preprocessing pipeline \cite{glasser2013minimal}, which includes artifact removal, motion correction, and registration to standard space. Additionally, we applied bandpass filtering ([0.01–0.1] Hz). In addition, a T1-weighted MPRAGE structural scan was obtained in each session (acquisition time: 7.6 minutes; TR: 2.4 s; TE: 2.14 ms; TI: 1 s; isotropic spatial resolution: $0.7 \times 0.7 \times 0.7$ $mm^3$). Out of 1,113 HCP subjects, 1,092 subjects were included in this study, after excluding subjects with corrupted data. 

\begin{table}[t!]
\setlength\extrarowheight{1pt}
\centering
\caption{Demographic information of utilized subjects in HCP, ABIDE I, and ABIDE II datasets. Acq. Dur.: acquisition duration.}
\label{data_demographics}
\resizebox{\columnwidth}{!}{\begin{tabular}{cccccc}
\hline \hline
Dataset & Subjects & Recordings & Gender (F/M) & Age & Acq. Dur (mins) \\ \hline
HCP      & 1,092 & 2,117 & 593/499 & 28.87 $\pm$ 3.58 & 14.33  \\
ABIDE I  & 500 NC \& 446 ASD & 995 & 155/791 & 16.13 $\pm$ 7.49 & 6.47 $\pm$ 1.79  \\
ABIDE II & 323 NC \& 256 ASD & 730 & 151/428 & 13.89 $\pm$ 8.13 & 6.67 $\pm$ 1.97 \\
\hline \hline
\end{tabular}}
\end{table}

\textbf{Autism Brain Imaging Data Exchange I (ABIDE I).} The ABIDE I dataset comprises rs-fMRI data from 17 international sites, acquired using different scanners and imaging protocols \cite{di2014autism}. It includes 539 individuals diagnosed with Autism Spectrum Disorder (ASD) and 573 neurotypical controls (NCs), with ages ranging from 7 to 64 years and acquisition durations ranging from 3.32 mins to 10 mins. Several preprocessed versions of the dataset are publicly available, each generated using different preprocessing pipelines. For this study, we used data processed with the fMRIPrep pipeline \cite{esteban2019fmriprep}, which includes rs-fMRI reference image estimation, head-motion and slice-timing correction, and susceptibility distortion correction. The rs-fMRI volumes were then normalized to the MNI152 NLIN 2009b template. ICA-AROMA \cite{pruim2015ica} was applied to remove motion-related artifacts, after which cerebrospinal fluid and white matter signals were regressed out. Finally, the signals were bandpass-filtered to the typical [0.01–0.1] Hz range. Out of 1,112 ABIDE I subjects, 946 subjects were included in this study, after excluding subjects with missing information.

\textbf{Autism Brain Imaging Data Exchange II (ABIDE II).} The ABIDE II dataset consists of rs-fMRI scans from 19 international sites, collected using diverse scanners and imaging protocols \cite{di2017enhancing}. It includes 521 individuals with ASD and 593 NCs, with ages ranging from 5 to 64 years and scan durations between 5.14 and 16.21 minutes. As with ABIDE I, we used data preprocessed with the fMRIPrep pipeline \cite{esteban2019fmriprep}, applying the same preprocessing steps. Of the 1,114 available subjects, 730 were included in this study. Subjects with missing functional scans or incomplete metadata were excluded.

\autoref{data_demographics} summarizes the number of subjects and recordings used from the HCP, ABIDE I, and ABIDE II datasets, along with key demographic details and acquisition duration.

Data from the HCP, ABIDE I and ABIDE II datasets were parcellated using two popular brain parcellations: the 166-ROI AAL3 \cite{rolls2020automated} and the 384-ROI AICHA \cite{joliot2015aicha} atlases. These atlases were selected to assess the impact of parcellation resolution: AAL3 provides a lower-resolution anatomical parcellation with fewer ROIs, whereas AICHA offers a much detailed functional parcellation. 

To ensure temporal consistency across datasets, all rs-fMRI data were resampled to a uniform TR of 1.5 s. After resampling, ABIDE I data varied in length from 120 to 320 time points. For acquisitions at one of the site (University of Michigan) that exceeded eight minutes, data were cropped to match this duration. These data manipulations were performed here for consistency purposes across the three datasets, as VarCoNet can handle any rs-fMRI durations and temporal resolutions. 

\subsubsection{Hyperparameter settings} \label{impdets}
The 1D-CNN in the model architecture consists of a single convolutional layer with 16 kernels, a kernel size of 8, and a stride of 4. The kernel size was chosen to match the typical duration of the hemodynamic response function (approximately 12 s \cite{VOSS2016187}, corresponding to 8 samples with a TR of 1.5 s). This duration of 12 s should suffice to capture the major temporal dynamics of a typical hemodynamic response, ignoring between-ROI variations in neurovascular coupling and the small signal change during the undershoot phase \cite{TAYLOR2018322,MIEZIN2000735}.  The stride of 4 ensures that neighboring tokens are sufficiently different, reducing redundancy in the attention operations performed by the Transformer. Regarding the Transformer architecture and the critical hyperparameters temperature $\tau$ and learning rate $lr$, the Bayesian optimizer produced the results shown in \autoref{BOparams}.

\begin{table}[t!]
\setlength\extrarowheight{1pt}
\centering
\caption{Hyperparameters Obtained Through Bayesian Optimization.}
\label{BOparams}
\resizebox{\columnwidth}{!}{\begin{tabular}{ccccccc}
\hline \hline
Atlas & $N_{layers}$ & $N_{heads}$ & Batch size & $FF_{dim}$ & $lr$ & $\tau$ \\ \hline
AAL3 & 1 & 1 & 64 & 2048 & $2.375\times 10^{-4}$ & 0.054 \\
AICHA & 1 & 2 & 64 & 512 & $1.369\times 10^{-4}$ & 0.054 \\
\hline \hline
\end{tabular}}
\end{table}

For the data augmentation hyperparameters $L_{min}$ and $L_{max}$, their selection was guided by the varying sequence lengths in the ABIDE I dataset. Specifically, $L_{max}$ was set to 320 samples (8 minutes) to match the longest sequence in ABIDE I. The choice of $L_{min}$ was constrained by two factors: it had to be below 120 samples (3 minutes) to enable effective augmentation, but it also needed to be sufficiently large to prevent excessive disparity between the shortest and longest sequences, which could hinder contrastive learning. To determine an appropriate $L_{min}$, we trained models using different values of $L_{min}$ while keeping other hyperparameters fixed (as shown in \autoref{BOparams}). Subject fingerprinting was then evaluated using sequences of length $L_{w1}=L_{min}$, $L_{w3}=L_{max}$, and $L_{w2}=(L_{min} + L_{max})/2$, as described in Section \ref{hyperparamopt}. The goal was to identify an $L_{min}$ that maintained balanced fingerprinting performance across all six test conditions (see Section \ref{hyperparamopt}). Our results indicated that $L_{min} = 80$ samples provided stable fingerprinting rates. To facilitate training, all inputs are zero-padded to $L_{max} = 320$ samples.

Since both the HCP and ABIDE I datasets include subjects with multiple recordings (nearly all subjects in HCP and 27 subjects in ABIDE I), it is important to exclude data repetitions from the same subject in the same batch to ensure unbiased contrastive process. Furthermore, for ASD classification, all subjects with multiple recordings are used exclusively for training to prevent information leakage between training, validation, and testing. While this does not affect VarCoNet, which is trained solely via SSL, the presence of repetitions might introduce bias in the training of the linear classification layer.

For subject fingerprinting, we adopt a procedure similar to that of \cite{finn2015functional} (see Section \ref{evaluation_fing}). This approach relies solely on similarity among FCs (the output of VarCoNet) and does not require additional modules beyond VarCoNet’s trained encoder. In contrast, ASD classification necessitates training a fully connected network to process VarCoNet’s output. Notably, while VarCoNet’s encoder remains fully self-supervised, the fully connected module is trained in a supervised manner.  

To evaluate VarCoNet’s encoding capabilities, we employ a single linear layer with an input size matching the embeddings and an output size of two (corresponding to the two classes). This layer is followed by a Softmax function. Given separate training, validation, and test sets, the validation and test sets are entirely excluded from the contrastive training process. At each epoch of contrastive learning, we extract embeddings (vectorized FCs) from the training and validation sets and use them to train the linear classification layer. The classification validation loss is monitored, and the minimum validation loss along with the corresponding weights of the linear layer are stored for each contrastive epoch. Finally, the contrastive epoch that yielded the minimum validation loss is identified, and the weights of both VarCoNet and the linear layer at that epoch are restored for making predictions on the test set.

All experiments were implemented in PyTorch and executed on a single NVIDIA RTX 6000 Ada Generation GPU with 48 GB of VRAM. Our scripts, including rs-fMRI postprocessing, VarCoNet training, and testing, were written in Python (3.11.8) and are publicly available at \url{https://github.com/CharLamp10/VarCoNet-V2}. To ensure reproducibility, we provide clear instructions on accessing the preprocessed rs-fMRI data, along with the subject IDs for the training, validation, and testing sets. VarCoNet was trained using the Adam optimizer with the learning rate and batch size specified in \autoref{BOparams}, and a linear warmup cosine annealing learning rate scheduler with 10 warmup epochs. 

\subsection{Competing methods}
We compare VarCoNet with two DL-based subject fingerprinting methods and the conventional PCC benchmark, as well as with 11 DL-based methods for ASD classification. For methods with publicly available code, we use the original implementations, modifying only the data preparation process to ensure all methods are trained, validated, and tested on the same subjects. For methods without available code, we provide our own implementations, which we release alongside VarCoNet, closely following the details in the original manuscripts. In cases where models were implemented from scratch, we optimized key hyperparameters, such as the learning rate and the batch size, and monitored the training loss to ensure convergence. A brief description of the baseline methods is provided below.

\subsubsection{Subject fingerprinting baselines}
\begin{itemize}
\item \cite{finn2015functional}: The conventional PCC-based FC method is used as a benchmark for subject fingerprinting, providing a comparison against VarCoNet and other DL-based approaches.

\item \cite{cai2021functional}: This method combines an autoencoder for reconstructing parcellated fMRI time-series with the K-SVD algorithm. The autoencoder is implemented from scratch, while K-SVD is adapted from \url{https://github.com/syanga/ksvd-sparse-dictionary}. To accelerate computations, we convert the original NumPy-based implementation to PyTorch for GPU compatibility.

\item \cite{lu2024brain}: This method integrates an autoencoder that reconstructs PCC-based FCs with the K-SVD algorithm. The autoencoder is implemented from scratch, while K-SVD is adapted from \url{https://github.com/syanga/ksvd-sparse-dictionary}. To enhance computational efficiency, we convert the NumPy-based implementation to PyTorch for GPU acceleration.
\end{itemize}

\subsubsection{ASD classification baselines}
\begin{itemize}
\item \cite{riaz2020deepfmri}: DeepFMRI consists primarily of 1D-CNN and linear layer computations. As the method's code is not publicly available, we implemented it from scratch.

\item \cite{nguyen2020attend}: For BAnD, we use the implementation provided by the authors at \url{https://github.com/LLNL/BAnD}. 

\item \cite{kan2022fbnetgen}: For FBNETGEN, we use the implementation provided by the authors at  \url{https://github.com/Wayfear/FBNETGEN}.

\item \cite{thomas2022self}: As discussed in Section \ref{related}, this work investigates several SSL strategies, with causal sequence modeling achieving the best performance. To compare with VarCoNet, we fine-tune the Transformer model trained on thousands of fMRI scans using the causal sequence modeling framework, following the authors’ instructions and publicly available code at \url{https://github.com/athms/learning-from-brains}. This is the only method for which we do not use the AAL3 and AICHA atlases. Since the pretrained model requires input data with the exact same shape as used during training, we adopt the DiFuMo atlas to ensure compatibility with the original implementation.

\item \cite{kan2022brain} For Brain Network Transformer, we use the implementation provided by the authors at \url{https://github.com/Wayfear/BrainNetworkTransformer}.

\item \cite{wang2023unsupervised} For UCGL, we use the implementation provided by the authors at \url{https://github.com/xiaochuanwang-1005/UCGL}.

\item \cite{zhang2023gcl} For A-GCL, we use the implementation provided by the authors at \url{https://github.com/qbmizsj/A-GCL}.

\item \cite{bedel2023bolt}: For BolT, we use the implementation provided by the authors at \url{https://github.com/icon-lab/BolT}.

\item \cite{meng2024cvformer}: CvFormer primarily consists of Transformer and linear layers. As no official code is publicly available, we implement it from scratch.

\item \cite{zheng2025brainib}: For BrainIB, we use the implementation provided by the authors at \url{https://github.com/SJYuCNEL/brain-and-Information-Bottleneck}.

\item \cite{wang2025self}: For GCDA, we use the implementation provided by the authors at \url{https://github.com/xiaochuanwang-1005/GCDA}.
\end{itemize}

\subsection{Evaluation strategies} \label{evaluation}
\subsubsection{Subject fingerprinting evaluation} \label{evaluation_fing}
For subject fingerprinting, we first split the HCP data into training, validation, and test sets. Since fingerprinting rates strongly depend on the number of subjects—given that for $N$ subjects, there are $N-1$ negatives and only one positive—we use data from 200 subjects for validation and 393 subjects for testing, both exceeding the 126 subjects used in \cite{finn2015functional}. The remaining 931 recordings, comprising 67 from distinct subjects and 864 from 432 subjects, are used for training. To maintain the contrastive learning process, during training, recordings from the same subject were excluded from the same batch.

At each epoch of VarCoNet's contrastive training, fingerprinting performance is evaluated on the validation set using the same metric employed in Bayesian hyperparameter optimization (see Section \ref{hyperparamopt} and \autoref{fing_block_diagram}), with the only difference being that the signal lengths $L_{w1}$ and $L_{w2}$ are set to 80 and 200, respectively, as explained in Section \ref{impdets}. In brief, this metric is the harmonic mean of the average and the minimum fingerprinting rates across six distinct fingerprinting procedures, derived from different combinations of three signal lengths (80, 200, and 320 samples). For each length, 10 segments are extracted per signal using a sliding window optimized for maximal distance between windows, and the mean value is used to compute the aforementioned metric. The model weights from the epoch with the highest validation score are restored for testing. In testing, the same procedure is applied, except that the six fingerprinting rates are reported separately (mean and standard deviation across the 10 segments) rather than being summarized into a single metric.

To perform subject fingerprinting, we follow the approach of \cite{finn2015functional}. Specifically, we use the trained encoder to extract VarCoNet-based FCs from each session of each subject. For $N$ subjects, this results in two matrices of shape $N \times R(R-1)/2$, corresponding to the two sessions. Each row contains the flattened upper triangular part of the VarCoNet-based FC. Let $A_1$ and $A_2$ represent the matrices corresponding to the first and second sessions, respectively. We then compute the PCC between each row in $A_1$ and all rows in $A_2$. If the maximum PCC occurs between the two sessions of the same subject, the identification is considered successful. The process is repeated with $A_2$ as the anchor. Finally, the percentage of correct identifications is reported. Additionally, we provide the correlation matrix between $A_1$ and $A_2$, which shows the similarities between the first and second sessions across all subjects. This allows us to assess whether the obtained identification scores are primarily driven by low intra-subject variability (high correlation between the two sessions of the same subject) or high inter-subject variability (low correlation between the two sessions of different subjects).

\subsubsection{ASD classification evaluation} \label{evaluation_asd}
For ASD classification, we ensure a thorough and fair comparison among competing methods by employing repeated K-fold cross-validation (CV), widely recognized as the gold standard for evaluating machine learning and DL models \cite{raschka2018model}, particularly when model selection is involved. Here, the best model in each fold is selected based on the epoch with the lowest validation loss. As noted in \cite{raschka2018model,kohavi1995study}, 10-fold CV provides an optimal balance between bias and variance, leading us to set $K = 10$. Furthermore, based on \cite{raschka2018model,efron1994introduction}, 50 to 200 performance samples are generally sufficient for reliable estimation. Thus, we conduct 10 independent 10-fold CV runs, yielding 100 performance samples, and report the mean and standard deviation. In each CV iteration, nine folds are used for training, while the remaining fold is used for testing. To restore the model at the epoch of minimum validation loss, the training set is further divided into 85\% training and 15\% validation.

Beyond CV, we incorporate an external test set, following best practices recommended by \cite{colliot2023machine} for brain disorder classification. Specifically, we hold separately all rs-fMRI data from the Caltech site in the ABIDE I dataset, comprising 37 subjects (18 ASD, 19 NC), to form an independent test set. These subjects are not included in the 10-fold cross-validation. After completing cross-validation, the remaining ABIDE I data are split into 90\% for training and 10\% for validation to train VarCoNet and determine the epoch of minimum validation loss. The Caltech data are used solely for final testing.

To ensure robustness, this procedure is repeated 10 times using different random seeds, minimizing the chance that results are influenced by particularly favorable or unfavorable validation splits. Subsequently, both VarCoNet and the best competing method are evaluated on the ABIDE II dataset using the same 10 models (corresponding to the 10 random seeds) previously used for classification of data from the Caltech site.

We also assess prediction stability between VarCoNet and the best competing method. Since the rs-fMRI data collected at the Caltech site are 5 minutes long, we use sliding windows of 3 and 4 minutes to extract three segments per window length, taken from the beginning, middle, and end of the time-series, yielding six cropped signals per subject. The durations of 3 and 4 minutes were chosen as they represent plausible acquisition lengths in real-world rs-fMRI studies. This analysis aims to test whether a model’s prediction for a given subject changes when only a segment of the full signal is used, effectively simulating slightly shorter acquisitions starting at different time points. Ideally, a robust model should maintain consistent predictions across all cropped segments, as the data originate from the same subject and are of realistic duration.

To quantify stability, we record how often the model's decision changes when using these partial signals. This is repeated across all 10 trained models used in the evaluation on the data from the Caltech site, and we report the percentage of instances where prediction inconsistency occurs.

In all cases, we use fixed random seeds to ensure that all methods are trained, validated, and tested on identical data splits.

\begin{figure}
    \centering
    \begin{subfigure}[b]{\columnwidth}
        \centering
       \includegraphics[width=\columnwidth]{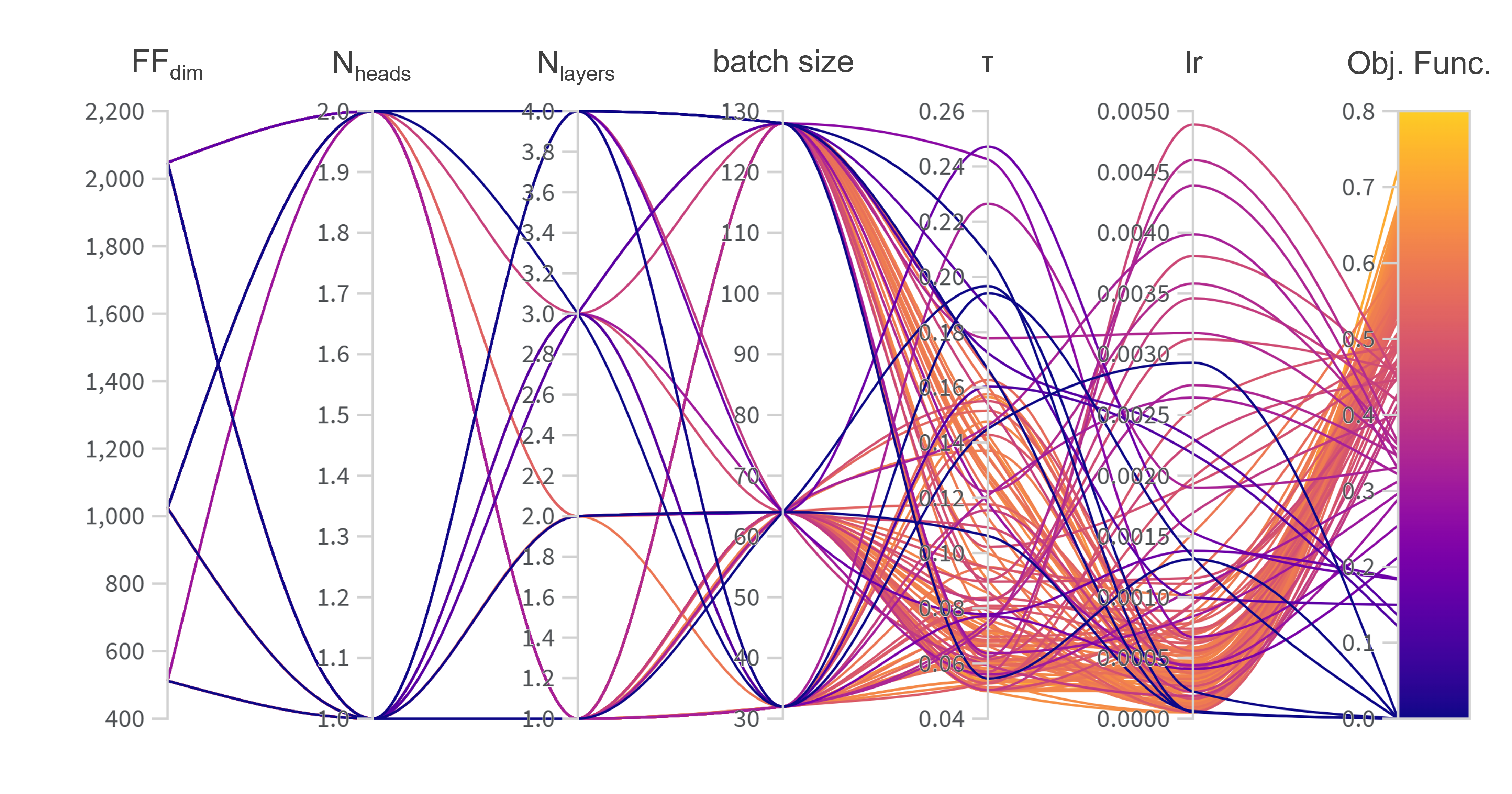}
       \caption{AAL atlas}
       \label{bo_plot_aal} 
    \end{subfigure}
    \centering
    \begin{subfigure}[b]{\columnwidth}
        \centering
       \includegraphics[width=\columnwidth]{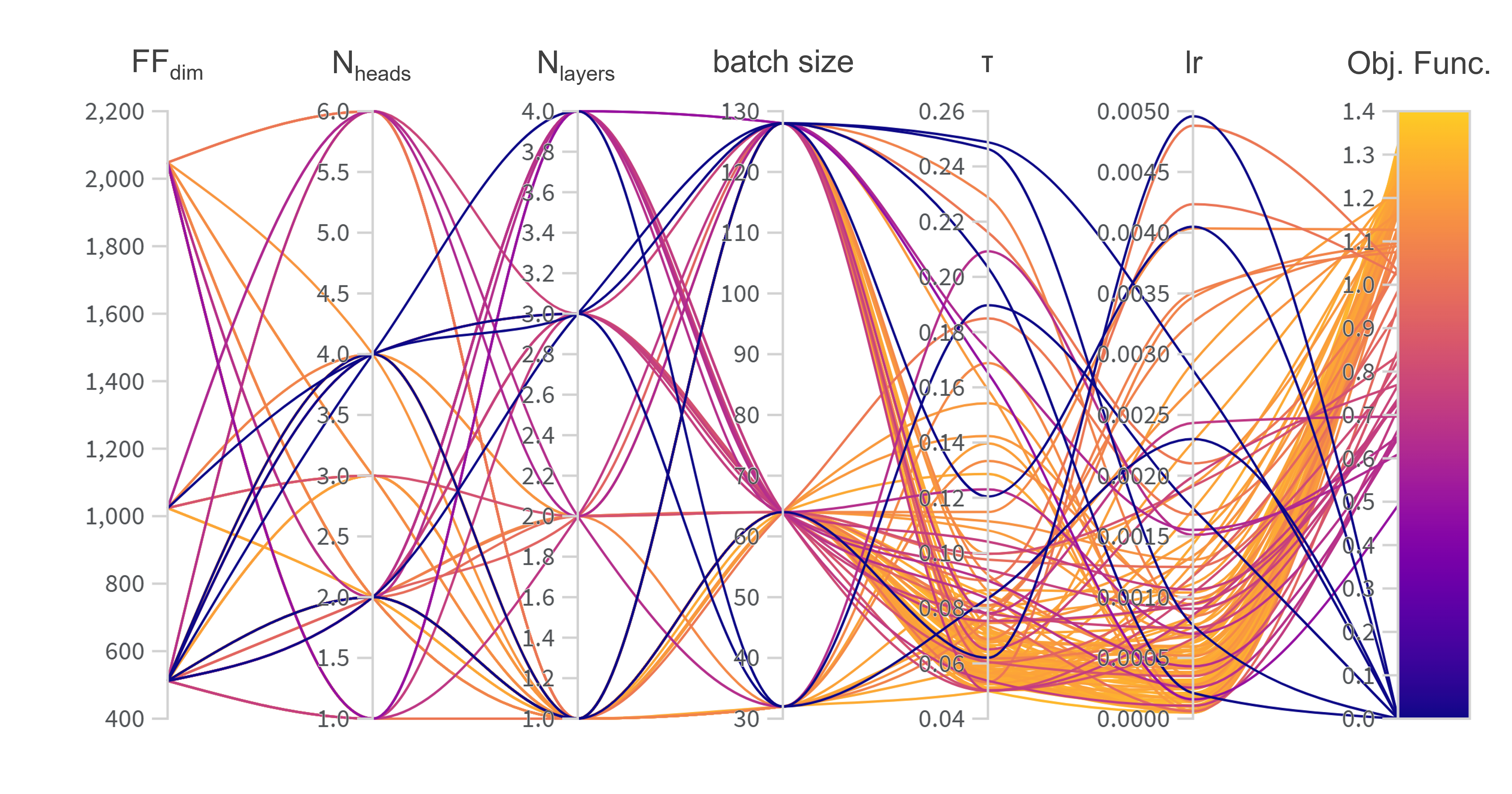}
       \caption{AICHA atlas}
       \label{bo_plot_aicha}
    \end{subfigure}
    \caption{Hyperparameter values for each Bayesian optimization trial and the corresponding objective function value. Each of the 125 runs, represented by a curved line, corresponds to a unique combination of $FF_{dim}$, $N_{heads}$, $N_{layers}$, batch size, learning rate ($lr$), and temperature parameter ($\tau$).}
    \label{bo_plots}
\end{figure}

For model evaluation, we use binary cross-entropy (BCE) loss, the area under the receiver operating characteristic curve (AUC), and the F1-score, three widely used metrics for assessing classification performance. Although BCE is not typically used as a standalone performance metric due to its unbounded nature, it is particularly effective for comparing models evaluated on the same test sets. This is because it directly reflects how well-calibrated the model’s probabilistic outputs are, favoring confidence scores that align with the true class probabilities. AUC quantifies the model’s ability to discriminate between ASD and NC subjects across all possible classification thresholds, making it robust to class imbalance. The F1-score, computed at the optimal decision threshold determined using Youden’s index, provides a balanced measure of precision and recall, offering a meaningful summary of classification performance.

\section{Results}

\subsection{VarCoNet hyperparameters}
\autoref{bo_plot_aal} and \autoref{bo_plot_aicha} show the combination of hyperparameters that maximized the objective function for the AAL and AICHA atlas, respectively, revealing several insights into VarCoNet’s behavior. For instance, the optimal number of Transformer layers is consistently found to be one (the minimum possible) suggesting that VarCoNet does not benefit from deeper attention processing. In fact, additional layers may lead to unnecessary over-processing of the input. Focusing on hyperparameters that directly affect contrastive learning, specifically, batch size and the temperature parameter $\tau$, we observe that the Bayesian optimizer consistently selects low values of $\tau$ as optimal. For batch size, a value of 64 consistently yields the best performance across both atlases. Moreover, the optimal $\tau$ is found to be approximately 0.054. Such a small value leads to highly peaked exponentials in Equation \ref{simclr}, thereby placing greater emphasis on hard negatives (samples that are not true pairs but lie close in representation space). Additionally, a smaller $\tau$ forces positive pairs to be mapped very close to each other, enhancing the model's ability to capture subject-specific traits, a desirable outcome for subject fingerprinting. While small $\tau$ values are known to produce large gradient magnitudes even for minor sample differences, we did not observe any instability during training in our experiments.

\subsection{VarCoNet convergence}
\begin{figure}
    \centering
    \begin{subfigure}{\columnwidth}
        \centering
        \includegraphics[width=\columnwidth]{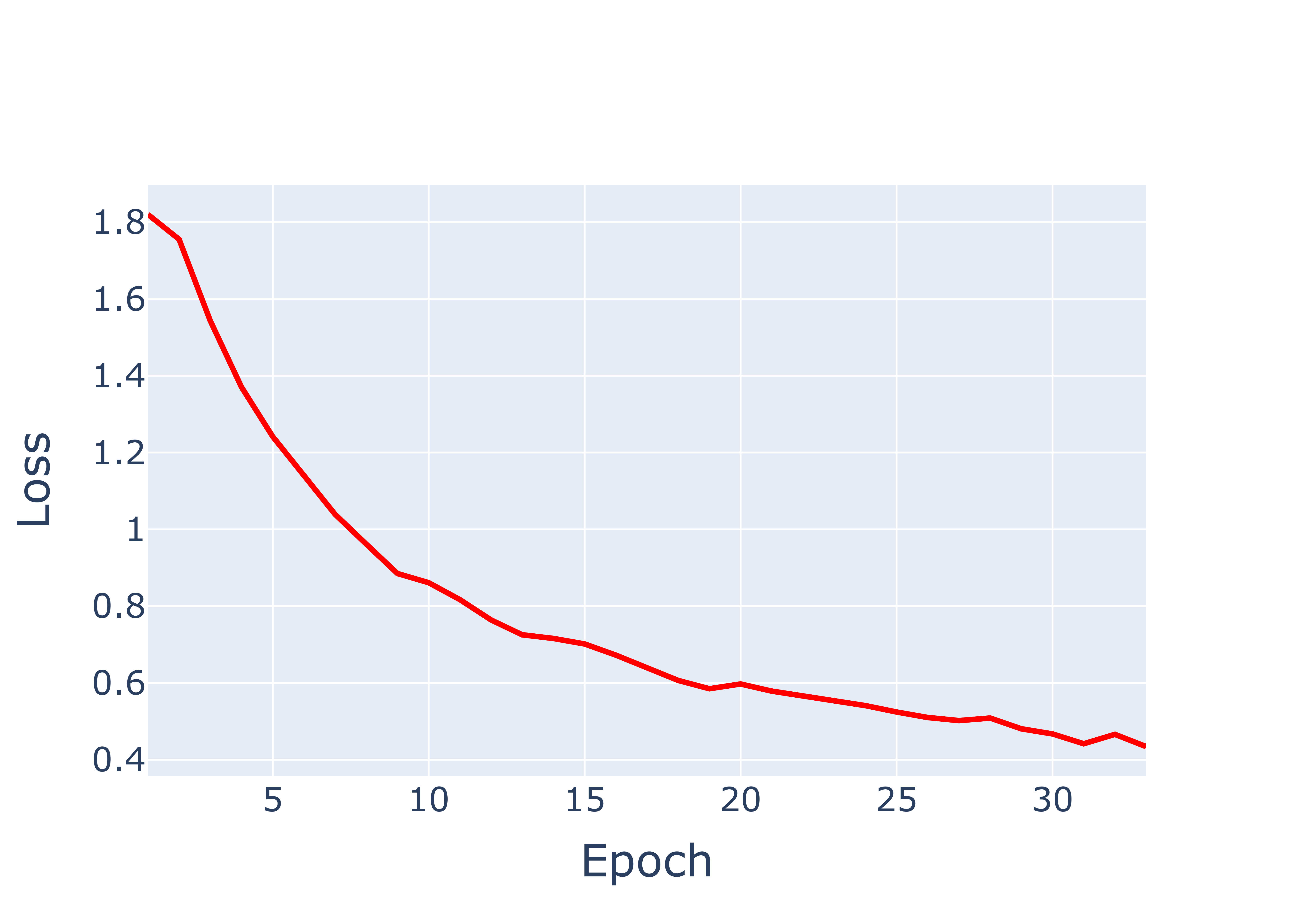}
        \caption{AAL3}
    \end{subfigure}
    \centering
    \begin{subfigure}{\columnwidth}
        \centering
        \includegraphics[width=\columnwidth]{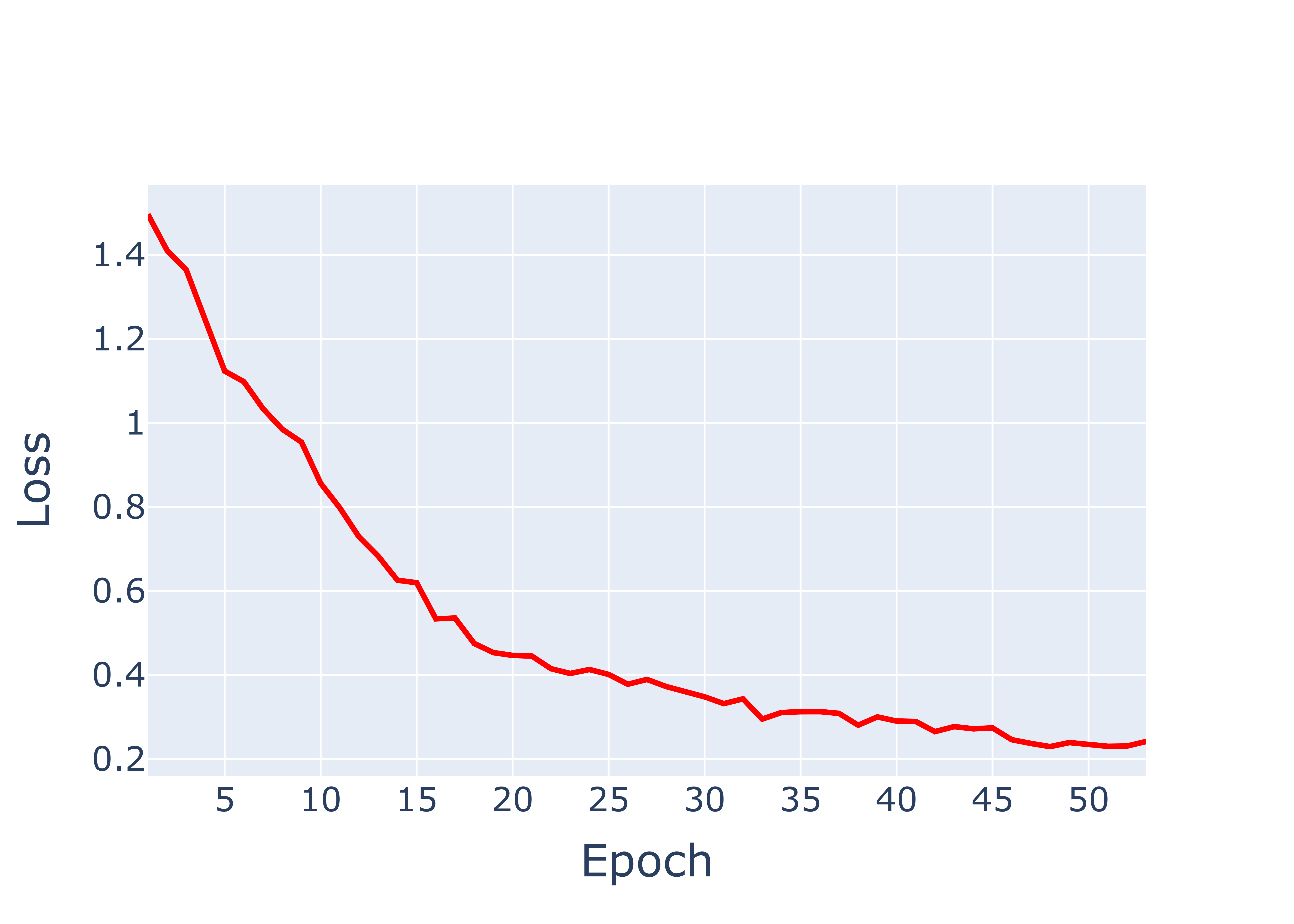}
        \caption{AICHA}
    \end{subfigure}
    \caption{Contrastive training loss of VarCoNet trained on 930 rs-fMRI scans from the HCP dataset, using the AAL3 (a) and AICHA (b) atlases for parcellation.}
    \label{training_loss}
\end{figure}
\autoref{training_loss} shows the SimCLR contrastive loss over training epochs for VarCoNet, trained on 930 rs-fMRI scans from the HCP dataset using both the AAL3 and AICHA atlases. The training curves exhibit a consistently smooth and decreasing loss, suggesting stable optimization and successful convergence. However, in contrastive SSL, convergence of the training loss does not always imply that the model has learned meaningful or transferable representations. To evaluate this, we assess the model's performance on two downstream tasks: subject fingerprinting and ASD classification, as presented below.

\subsection{Subject fingerprinting performance}
\begin{figure}
    \centering
    \begin{subfigure}{\columnwidth}
        \centering
        \includegraphics[width=\columnwidth]{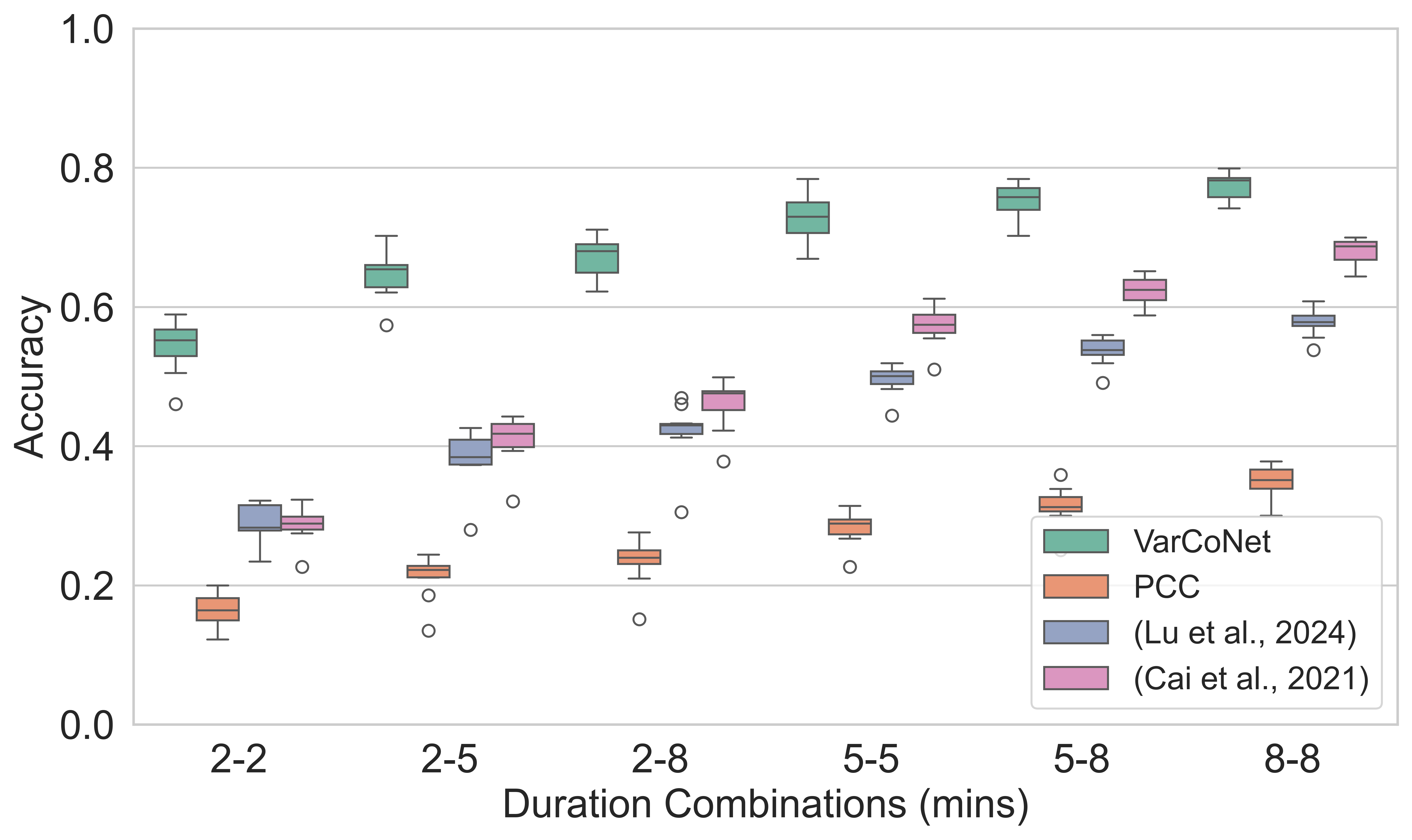}
        \caption{AAL3}
    \end{subfigure}
    \centering
    \begin{subfigure}{\columnwidth}
        \centering
        \includegraphics[width=\columnwidth]{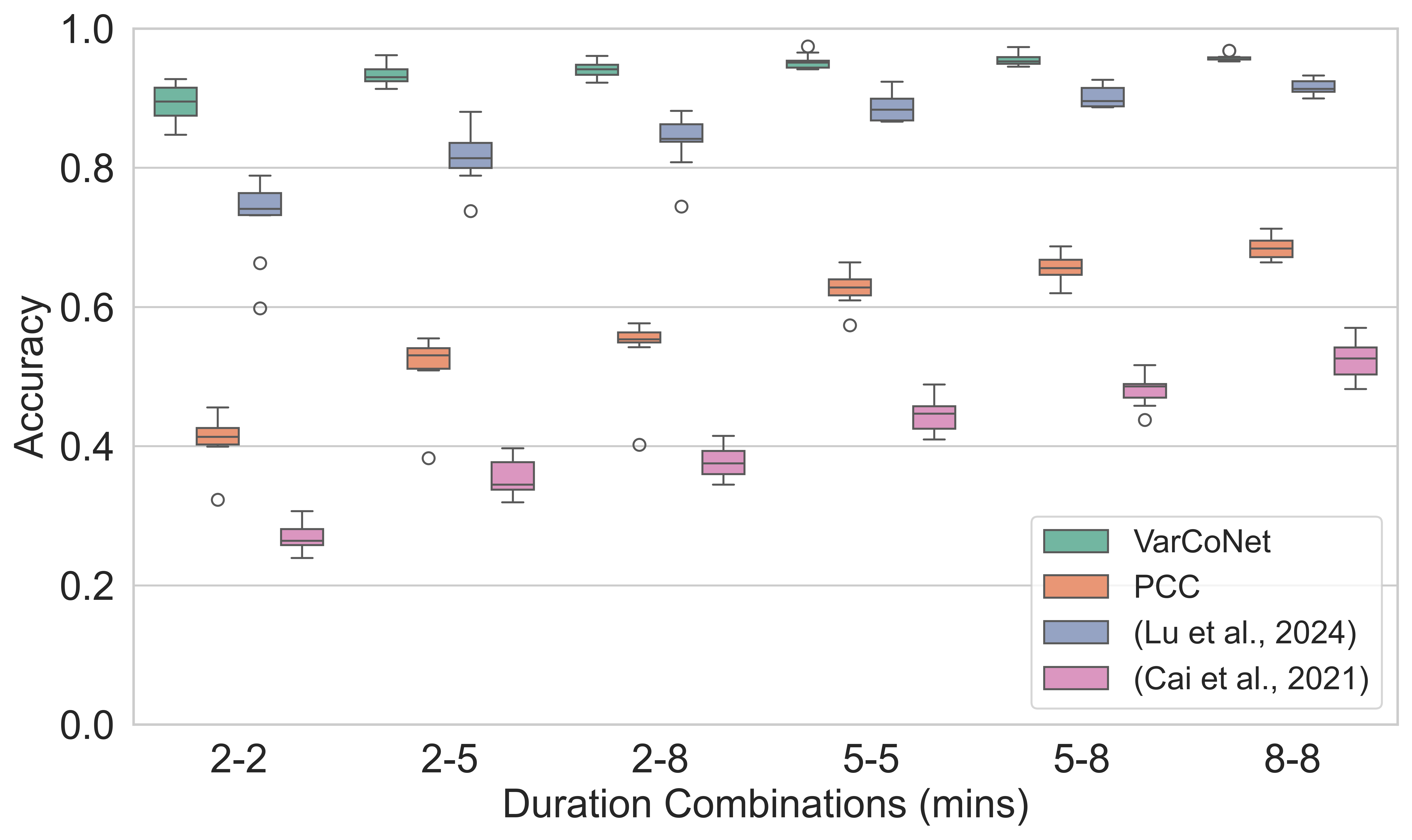}
        \caption{AICHA}
    \end{subfigure}
    \caption{Subject fingerprinting accuracy of VarCoNet compared to PCC and competing DL-based methods across varying combinations of rs-fMRI durations from the first and second sessions.}
    \label{fingerprinting_results}
\end{figure}

\begin{figure*}
    \centering
    \begin{subfigure}{0.24\textwidth}
        \includegraphics[width=\textwidth]{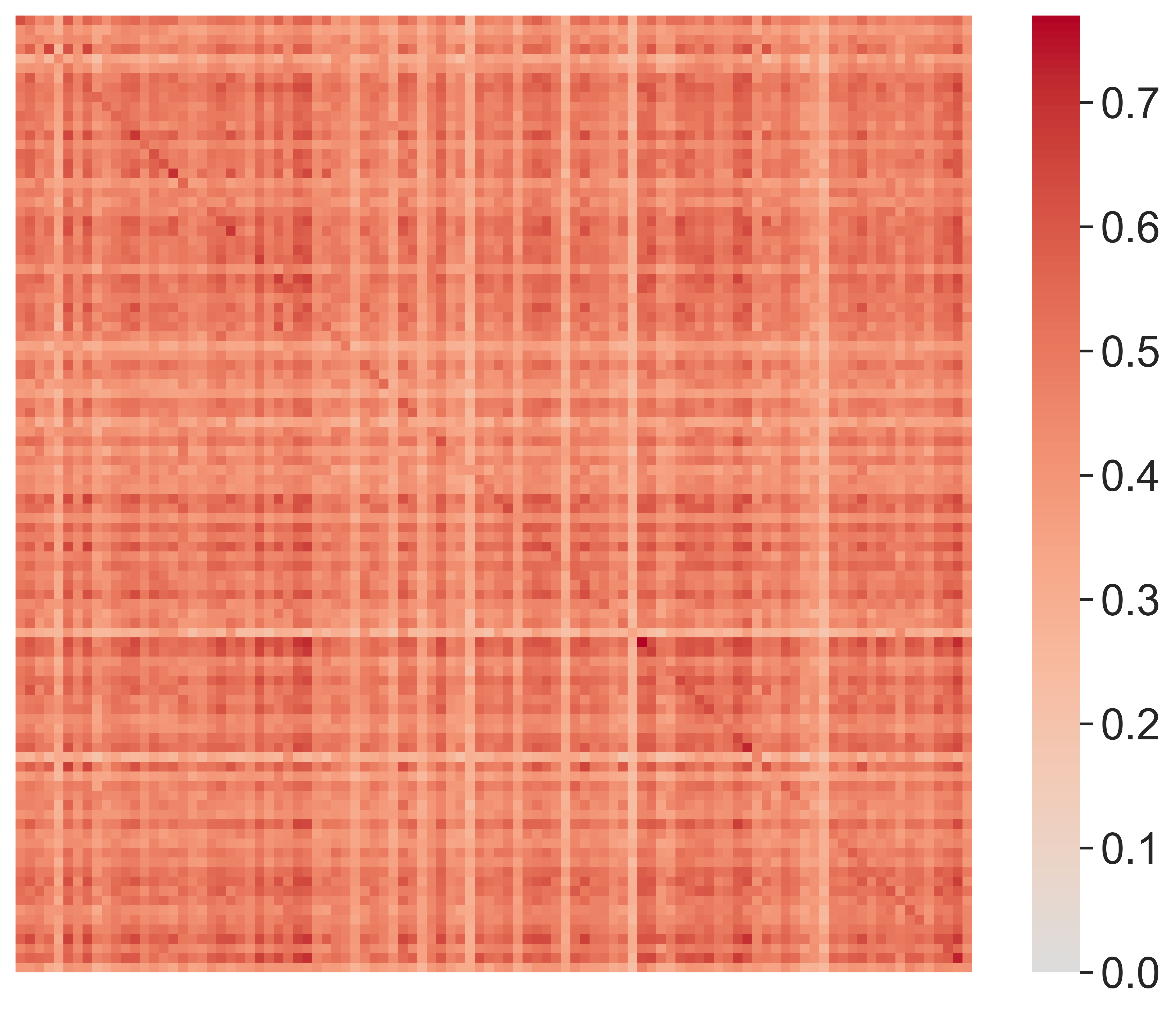}
        \caption{PCC, AAL}
        \label{intra_inter_subj_sims_pcc_aal}
    \end{subfigure}
    \begin{subfigure}{0.24\textwidth}
        \includegraphics[width=\textwidth]{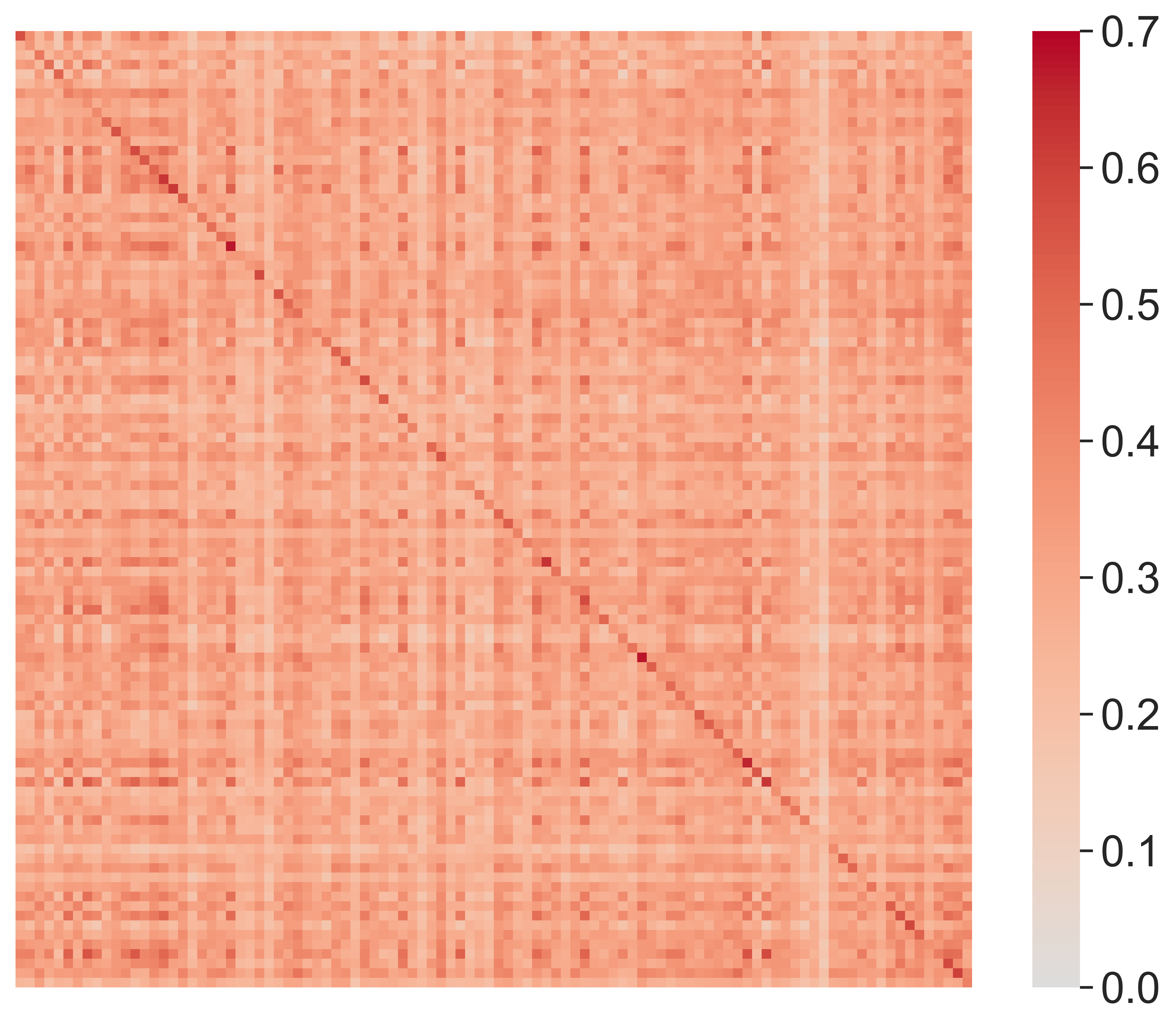}
        \caption{PCC, AICHA}
        \label{intra_inter_subj_sims_pcc_aicha}
    \end{subfigure}
    \begin{subfigure}{0.24\textwidth}
        \includegraphics[width=\textwidth]{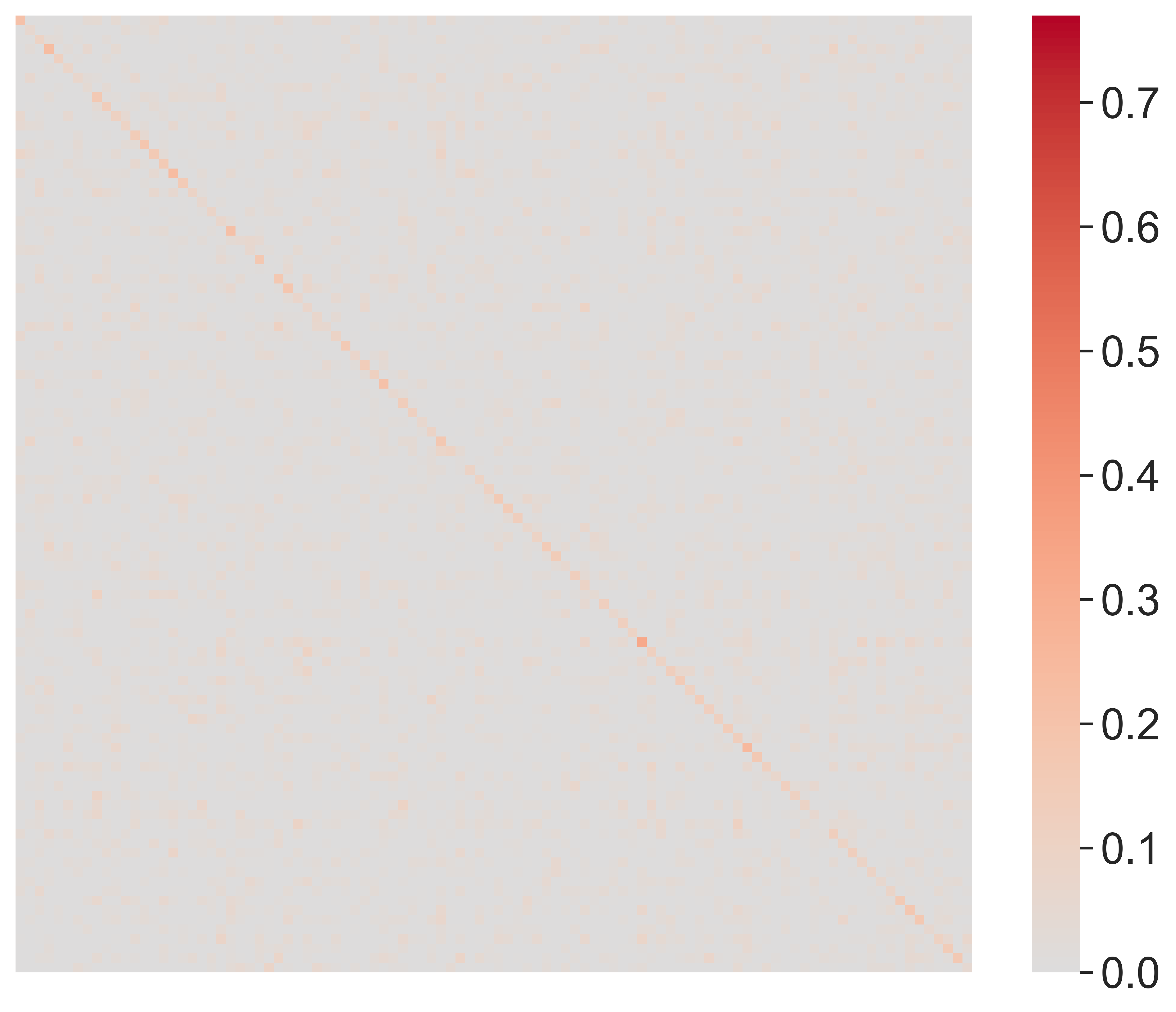}
        \caption{\cite{cai2021functional}, AAL}
        \label{intra_inter_subj_sims_ae_aal}
    \end{subfigure}
    \begin{subfigure}{0.24\textwidth}
        \includegraphics[width=\textwidth]{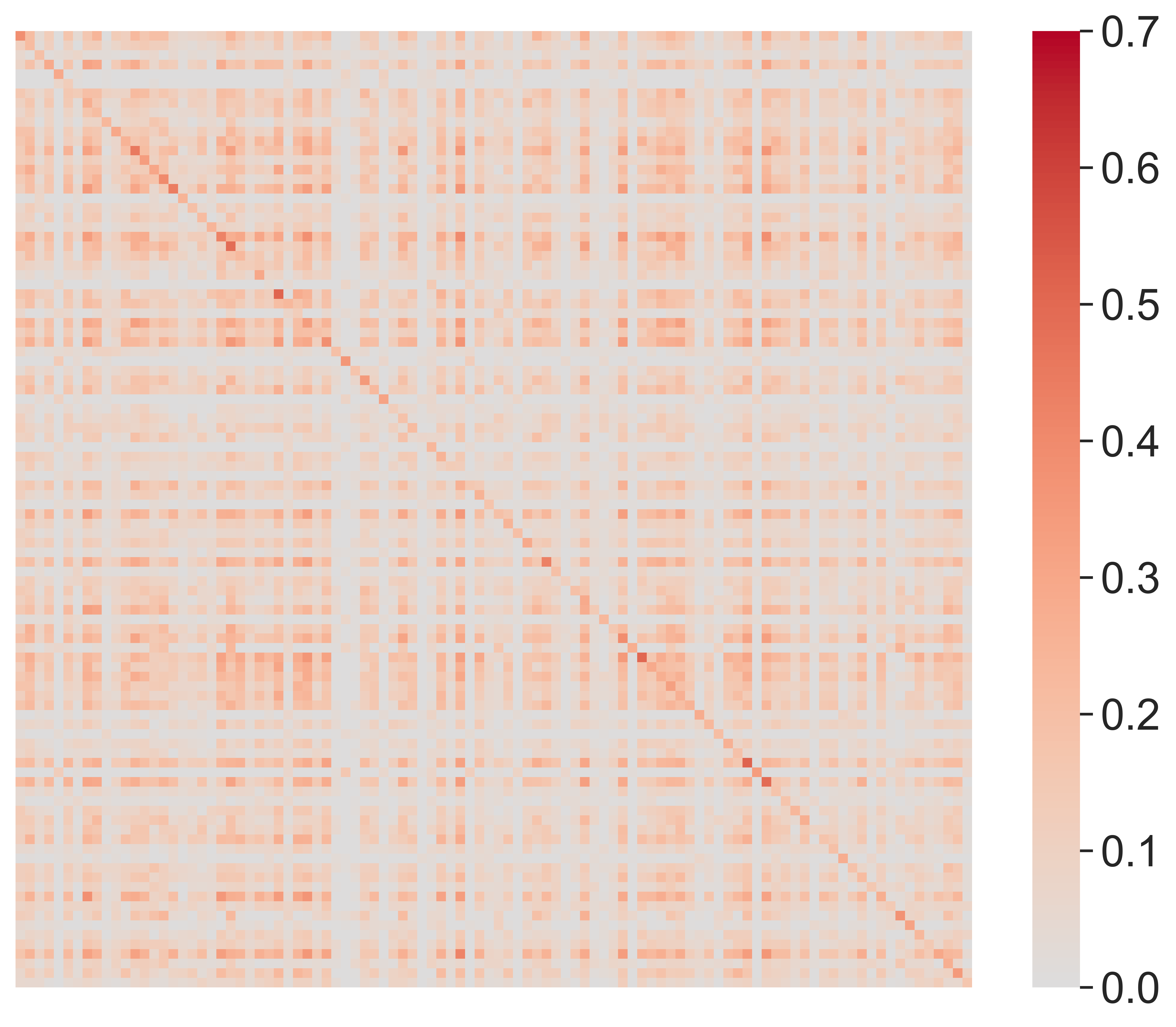}
        \caption{\cite{cai2021functional}, AICHA}
        \label{intra_inter_subj_sims_ae_aicha}
    \end{subfigure}
    \centering
    \begin{subfigure}{0.24\textwidth}
        \includegraphics[width=\textwidth]{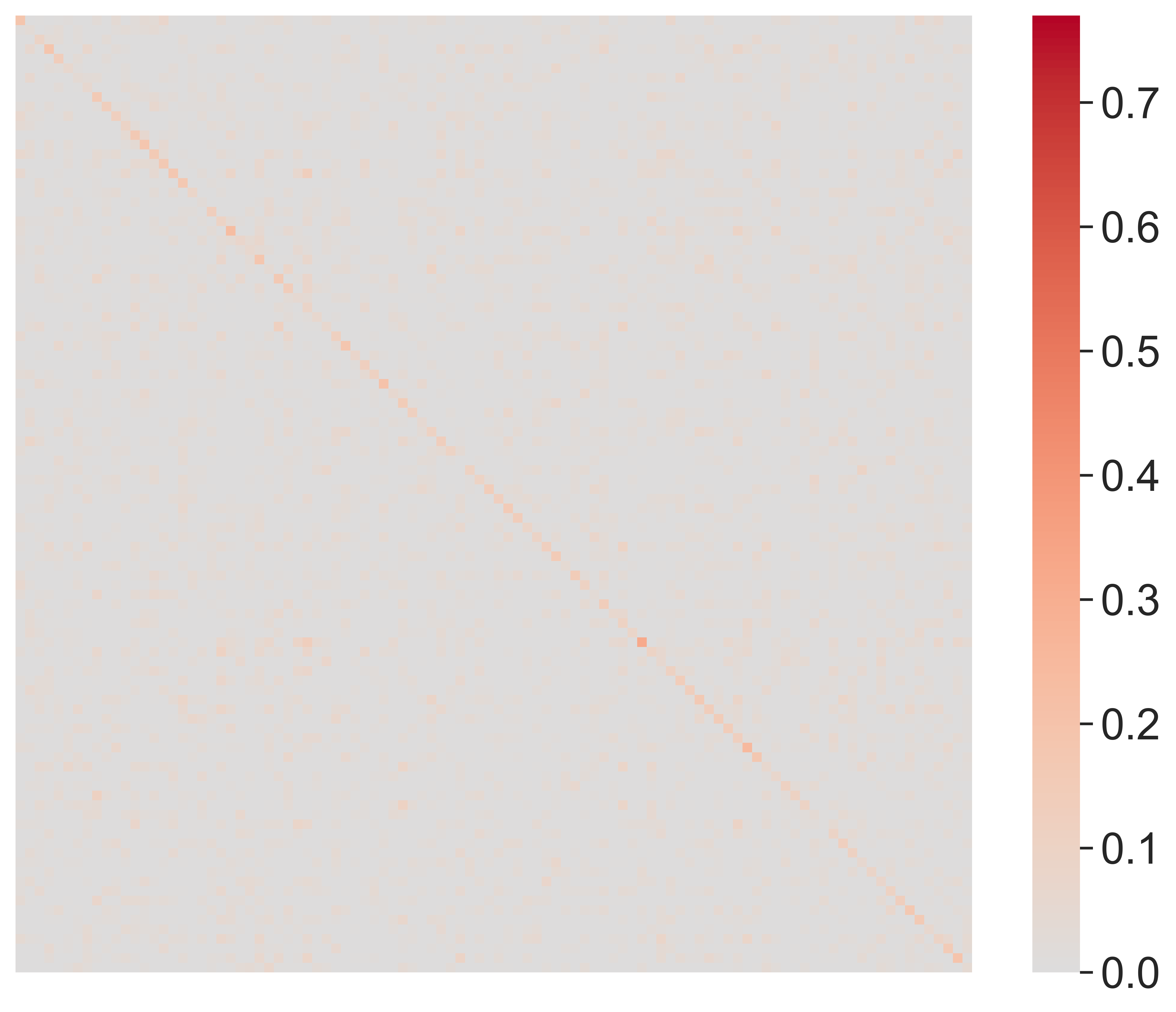}
        \caption{\cite{lu2024brain}, AAL}
        \label{intra_inter_subj_sims_vae_aal}
    \end{subfigure}
    \begin{subfigure}{0.24\textwidth}
        \includegraphics[width=\textwidth]{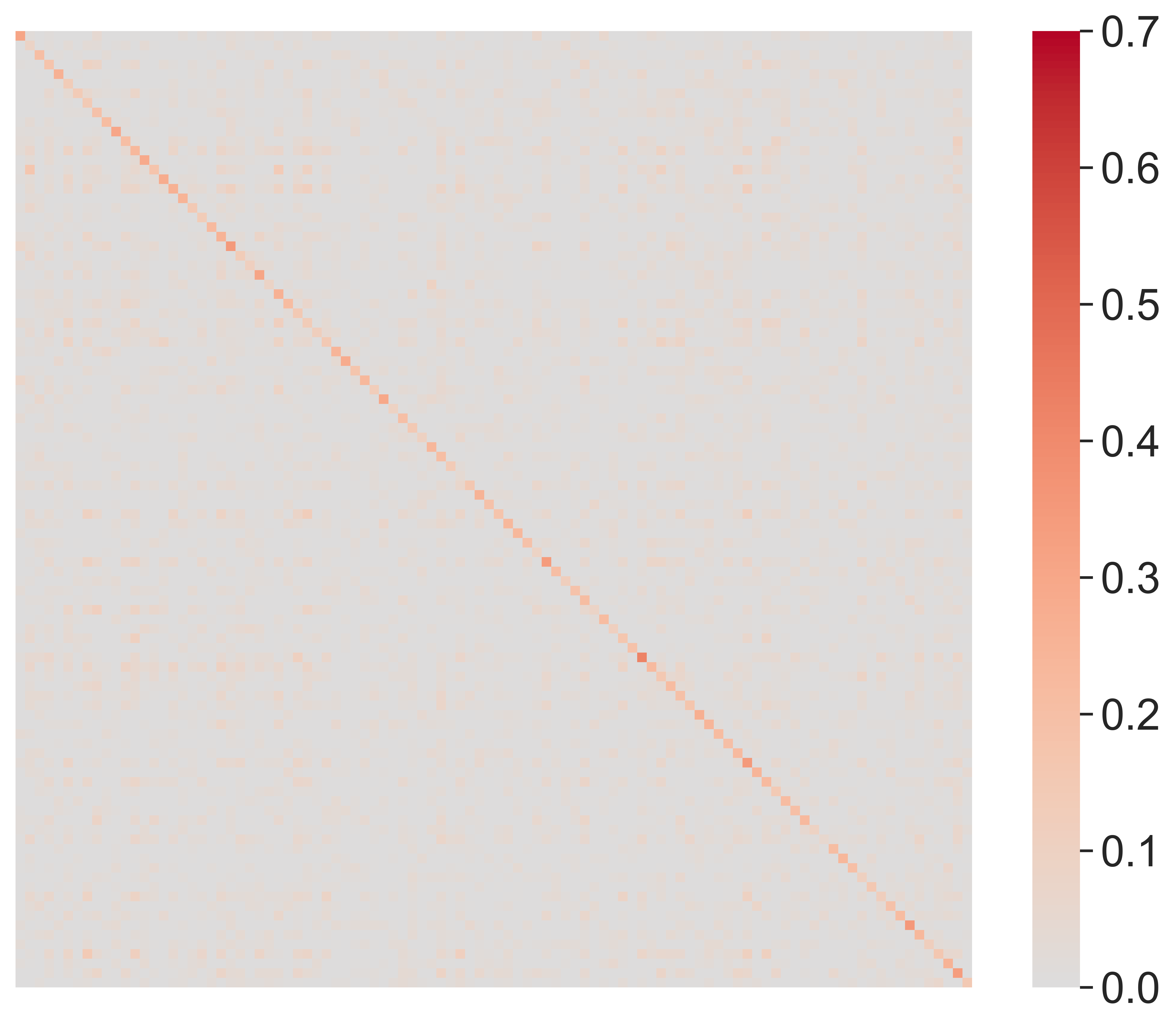}
        \caption{\cite{lu2024brain}, AICHA}
        \label{intra_inter_subj_sims_vae_aicha}
    \end{subfigure}
    \begin{subfigure}{0.24\textwidth}
        \includegraphics[width=\textwidth]{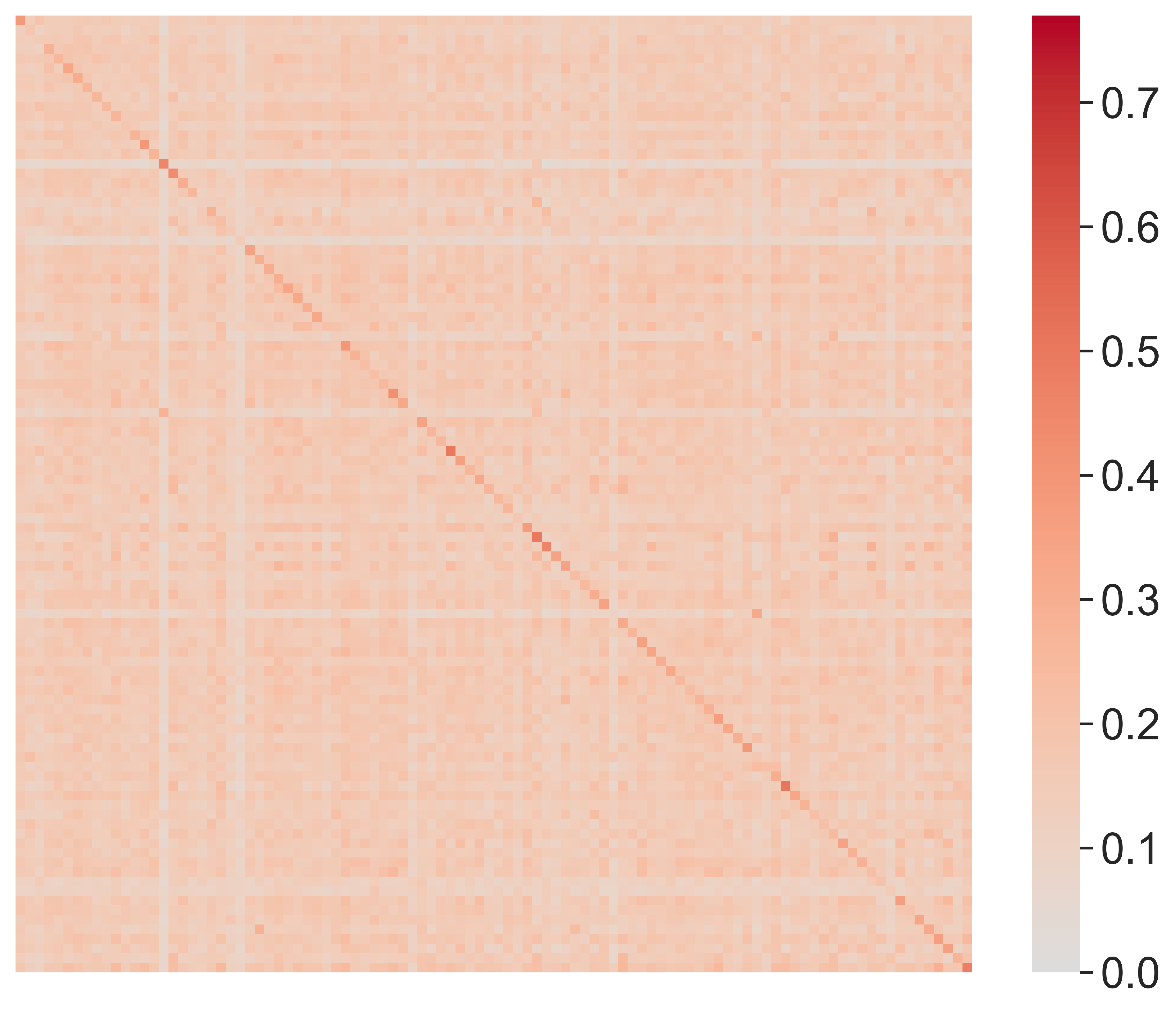}
        \caption{VarCoNet, AAL}
        \label{intra_inter_subj_sims_varconet_aal}
    \end{subfigure}
    \begin{subfigure}{0.24\textwidth}
        \includegraphics[width=\textwidth]{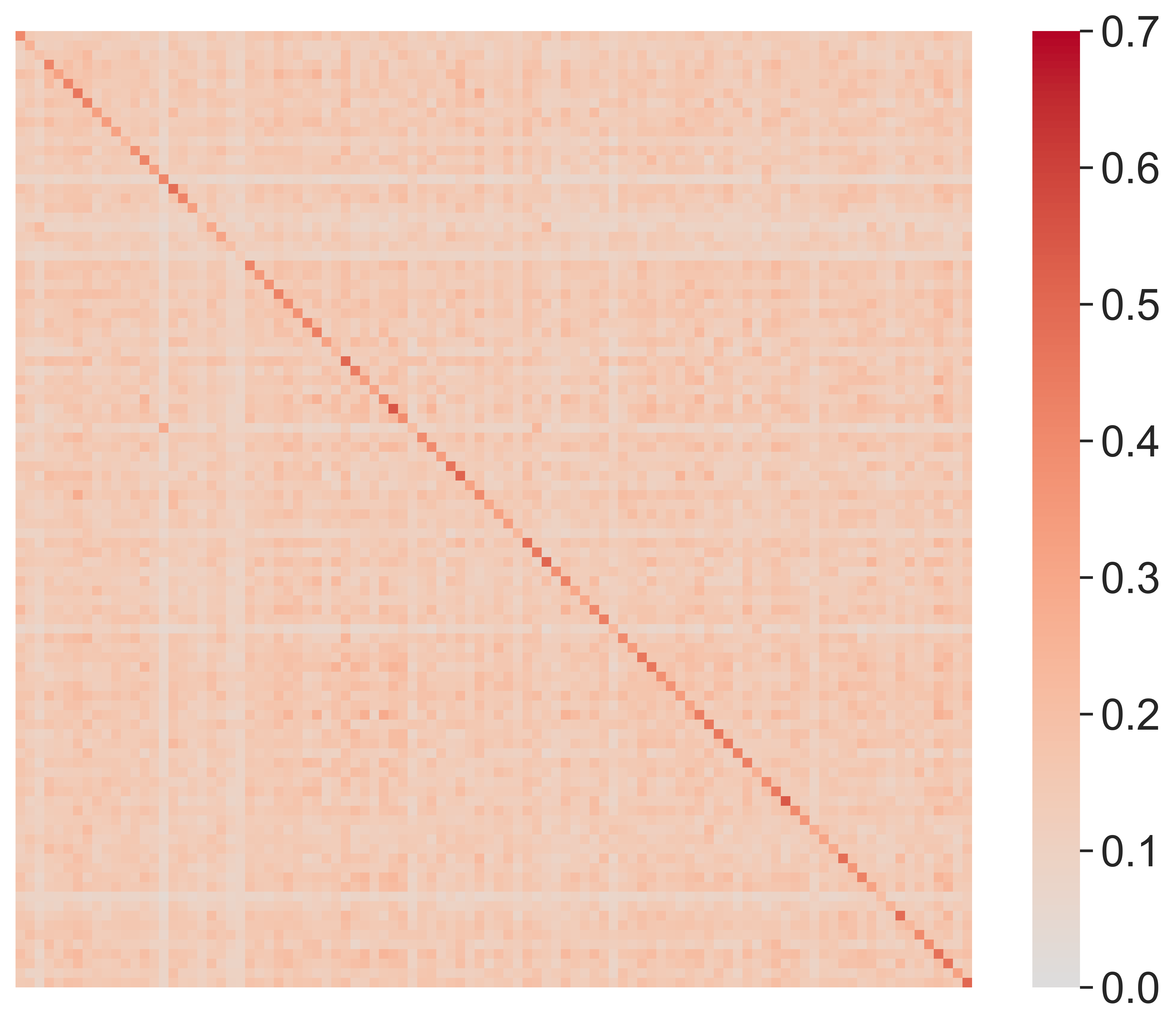}
        \caption{VarCoNet, AICHA}
        \label{intra_inter_subj_sims_varconet_aicha}
    \end{subfigure}
    
    \caption{Intra- and inter-subject similarity matrices of VarCoNet and competing methods for both atlases (AAL3 and AICHA). Each matrix contains 100 rows and 100 columns, representing 100 subjects from the test set. Rows correspond to the first session and columns to the second session; therefore, diagonal elements represent intra-subject similarities, while off-diagonal elements represent inter-subject similarities.}
    \label{intra_inter_subject_similarities}
\end{figure*}

\autoref{fingerprinting_results} presents the subject fingerprinting performance of VarCoNet, alongside two competing DL-based methods and the conventional PCC across all duration combinations and atlases. Further insights are provided in \autoref{intra_inter_subject_similarities}, which displays intra- and inter-subject similarity matrices to determine whether improvements in fingerprinting arise from reduced intra-subject variability, increased inter-subject variability, or both. These matrices are the average of 60 matrices obtained by performing fingerprinting 10 times for each of the six duration combinations. For illustration purposes, the correlation matrices in \autoref{intra_inter_subject_similarities} show results for 100 subjects from the test set, rather than all 393. In each matrix, the rows correspond to session 1 embeddings and the columns to session 2 embeddings for the same 100 subjects. These matrices are not symmetric, as the similarity between subject $i$ in session 1 and subject $j$ in session 2 is not necessarily equal to the reverse.

As shown in \autoref{fingerprinting_results}, compared to the methods of \cite{lu2024brain, cai2021functional} and the standard PCC, VarCoNet consistently outperforms across all duration combinations and atlases, with particularly strong performance in combinations involving shorter time courses (e.g., 2 minutes). Two noteworthy trends emerge from \autoref{fingerprinting_results}. First, fingerprinting accuracy improves as the duration increases. Second, except for the method of \cite{cai2021functional}, the more fine-grained functional AICHA atlas generally yields better results than the coarser anatomical AAL3 atlas. Perhaps most importantly, VarCoNet exhibits the lowest variation in fingerprinting accuracy across duration combinations, a desirable property when working with multi-site datasets, where differences in scan lengths must not introduce significant biases.

Further insights are provided by examining intra- and inter-subject similarities in \autoref{intra_inter_subject_similarities}. The methods of \cite{cai2021functional, lu2024brain} appear to improve fingerprinting primarily by reducing inter-subject similarities. In contrast, VarCoNet improves fingerprinting by both reducing inter-subject similarities and increasing intra-subject similarities. As a result, the VarCoNet-based FCs preserve inter-individual variability that reflects meaningful population-level characteristics.

\subsection{ASD classification performance} \label{ASD_results_section}
\autoref{ASD_results} presents the ASD classification results of VarCoNet in comparison to competing DL methods. VarCoNet consistently outperforms all other approaches, achieving superior performance across evaluation settings and atlases. Specifically, the methods of \cite{riaz2020deepfmri,nguyen2020attend,wang2023unsupervised,zhang2023gcl,meng2024cvformer,wang2025self}, namely DeepFMRI, BAnD, UCGL, A-GCL, CVFormer, and GCDA perform poorly on ASD classification, with AUC scores close to chance level in both the 10-fold CV and the external test set. Possible explanations for their low performance are discussed in Section \ref{discussion_baselines}.

The Transformer model trained with the CSM framework \cite{thomas2022self}, fine-tuned for ASD classification, shows moderate performance, consistently lower than VarCoNet across both evaluation settings and atlases. Similar moderate results are observed for FBNETGEN \cite{kan2022fbnetgen}, BNT \cite{kan2022brain}, and BrainIB \cite{zheng2025brainib}. Among existing approaches, BolT \cite{bedel2023bolt} achieves the closest performance to VarCoNet, though the performance of BolT remains lower than that of VarCoNet.

Using the AAL3 atlas, VarCoNet achieves lower BCE loss than BolT in both the 10-fold CV and the external test set. In the 10-fold CV, VarCoNet improves AUC and F1-score by approximately 3\%. These gains become more pronounced in the external test set, reaching around 12\% in AUC and 9\% in F1-score. Similar trends are observed with the AICHA atlas: under 10-fold CV, VarCoNet shows improvements of roughly 3\% in both AUC and F1-score compared to BolT, which increase to about 5.5\% and 4\%, respectively, in the external test set. In all cases, VarCoNet also maintains consistently lower BCE loss than BolT. While both methods appear to benefit from the more fine-grained AICHA atlas, these improvements are not statistically significant given the standard deviations.

\begin{table*}[t!]
\scriptsize
\centering
\caption{ASD classification results on the ABIDE I dataset (mean $\pm$ standard deviation). Boldface indicates the best-performing method for each metric}
\label{ASD_results}
\resizebox{\textwidth}{!}{\begin{tabular}{ccccccc}
\hline \hline

& \multicolumn{3}{c}{{10-Fold CV}} & \multicolumn{3}{c}{{Ext. Test (Caltech)}}\\ \hline

Method & BCE Loss (\%) & AUC (\%) & F1-score (\%) & BCE Loss (\%) & AUC (\%) & F1-score (\%) \\ \hline

& \multicolumn{6}{c}{\multirow{1}{*}{{DiFuMo}}}\\ \hline

CSM \cite{thomas2022self} & 0.664 $\pm$ 0.036 & 66.00 $\pm$ 4.91 & 64.02 $\pm$ 8.09 & 0.716 $\pm$ 0.009 & 55.32 $\pm$ 1.89 & 59.60 $\pm$ 9.12 \\ \hline

& \multicolumn{6}{c}{\multirow{1}{*}{{Volumetric}}}\\ \hline

BAnD \cite{nguyen2020attend} & 0.694 $\pm$ 0.004 & 51.62 $\pm$ 5.66 & 58.10 $\pm$ 15.2 & 0.700 $\pm$ 0.007 & 45.85 $\pm$ 5.99 & 57.38 $\pm$ 12.6 \\ \hline

& \multicolumn{6}{c}{{AAL3}}\\ \hline

DeepFMRI \cite{riaz2020deepfmri} & 0.692 $\pm$ 0.010 & 49.52 $\pm$ 4.66 & 52.24 $\pm$ 13.9 & 0.695 $\pm$ 0.020 & 47.89 $\pm$ 9.53 & 49.20 $\pm$ 19.2 \\

FBNETGEN \cite{kan2022fbnetgen}  & 0.688 $\pm$ 0.028 & 58.02 $\pm$ 6.04 & 57.65 $\pm$ 10.6 & 0.716 $\pm$ 0.054 & 55.32 $\pm$ 8.95 & 63.38 $\pm$ 9.46 \\

BNT \cite{kan2022brain}     & 0.703 $\pm$ 0.053 & 55.57 $\pm$ 6.55 & 56.74 $\pm$ 11.2 & 0.698 $\pm$ 0.017 & 61.82 $\pm$ 4.05 & 42.73 $\pm$ 26.4 \\

UCGL \cite{wang2023unsupervised} & 0.703 $\pm$ 0.021 & 50.93 $\pm$ 6.03 & 52.10 $\pm$ 15.3 & 0.712 $\pm$ 0.038 & 52.8 $\pm$ 10.24 & 53.92 $\pm$ 12.2 \\

A-GCL \cite{zhang2023gcl}     & 0.691 $\pm$ 0.007 & 52.31 $\pm$ 7.61 & 51.53 $\pm$ 17.1 & 0.697 $\pm$ 0.011 & 47.78 $\pm$ 11.68 & 45.56 $\pm$ 27.7 \\
 
BolT \cite{bedel2023bolt}    & 0.663 $\pm$ 0.026 & 66.36 $\pm$ 4.70 & 64.72 $\pm$ 5.75 & 0.728 $\pm$ 0.067 & 55.70 $\pm$ 4.32 & 64.83 $\pm$ 4.11 \\

CVFormer \cite{meng2024cvformer} & 0.730 $\pm$ 0.036 & 52.53 $\pm$ 5.94 & 50.88 $\pm$ 16.1 & 0.713 $\pm$ 0.048 & 53.52 $\pm$ 10.4 & 51.47 $\pm$ 10.7 \\

BrainIB \cite{zheng2025brainib} & 0.682 $\pm$ 0.021 & 59.54 $\pm$ 6.21 & 58.63 $\pm$ 12.8 & 0.720 $\pm$ 0.029 & 46.49 $\pm$ 4.10 & 40.37 $\pm$ 25.7 \\

GCDA \cite{wang2025self} & 0.708 $\pm$ 0.039 & 54.10 $\pm$ 6.70 & 54.05 $\pm$ 13.1 & 0.754 $\pm$ 0.053 & 41.61 $\pm$ 10.8 & 48.3 $\pm$ 24.2 \\

VarCoNet                 & 0.633 $\pm$ 0.030 & 69.91 $\pm$ 4.74 & 67.54 $\pm$ 5.25 & 0.653 $\pm$ 0.025 & \textbf{67.22 $\pm$ 3.71} & \textbf{73.63 $\pm$ 3.33} \\ \hline

& \multicolumn{6}{c}{\multirow{1}{*}{{AICHA}}}\\ \hline

DeepFMRI \cite{riaz2020deepfmri} & 0.698 $\pm$ 0.010 & 48.62 $\pm$ 7.11 & 45.82 $\pm$ 20.1 & 0.703 $\pm$ 0.010 & 51.40 $\pm$ 13.9 & 61.43 $\pm$ 16.4 \\

FBNETGEN \cite{kan2022fbnetgen}  & 0.691 $\pm$ 0.042 & 60.43 $\pm$ 6.63 & 60.87 $\pm$ 7.72 & 0.791 $\pm$ 0.116 & 51.64 $\pm$ 10.1 & 53.92 $\pm$ 19.8 \\

BNT \cite{kan2022brain}     & 0.696 $\pm$ 0.025 & 51.76 $\pm$ 7.55 & 54.11 $\pm$ 14.5 & 0.702 $\pm$ 0.016 & 59.78 $\pm$ 10.3 & 50.97 $\pm$ 21.8 \\

UCGL \cite{wang2023unsupervised} & 0.715 $\pm$ 0.029 & 51.38 $\pm$ 6.37 & 49.02 $\pm$ 16.9 & 0.732 $\pm$ 0.032 & 47.78 $\pm$ 6.49 & 53.69 $\pm$ 18.1 \\

A-GCL \cite{zhang2023gcl}     & 0.692 $\pm$ 0.007 & 51.00 $\pm$ 5.99 & 51.89 $\pm$ 16.9 & 0.693 $\pm$ 0.003 & 54.06 $\pm$ 16.0 & 47.53 $\pm$ 19.8 \\

BolT \cite{bedel2023bolt}    & 0.655 $\pm$ 0.036 & 69.23 $\pm$ 4.59 & 66.08 $\pm$ 6.57 & 0.685 $\pm$ 0.018 & 61.40 $\pm$ 5.31 & 67.27 $\pm$ 6.84 \\

CVFormer \cite{meng2024cvformer} & 0.742 $\pm$ 0.053 & 52.87 $\pm$ 5.88 & 52.96 $\pm$ 13.7 & 0.744 $\pm$ 0.066 & 51.55 $\pm$ 8.40 & 51.85 $\pm$ 8.32 \\

BrainIB\cite{zheng2025brainib} & 0.677 $\pm$ 0.021 & 61.07 $\pm$ 6.20 & 60.42 $\pm$ 11.42 & 0.701 $\pm$ 0.021 & 53.89 $\pm$ 4.90 & 50.55 $\pm$ 18.7 \\

GCDA \cite{wang2025self} & 0.710 $\pm$ 0.039 & 52.13 $\pm$ 7.02 & 51.19 $\pm$ 16.1 & 0.751 $\pm$ 0.082 & 52.72 $\pm$ 8.52 & 62.86 $\pm$ 6.49 \\

VarCoNet                 & \textbf{0.612 $\pm$ 0.036} & \textbf{72.62 $\pm$ 4.70} & \textbf{68.47 $\pm$ 5.62} & \textbf{0.650 $\pm$ 0.032} & 67.02 $\pm$ 5.04 & 71.41 $\pm$ 4.05 \\

\hline\hline 
\end{tabular}}
\end{table*}

\begin{table}[t!]
\scriptsize
\setlength\extrarowheight{1.5pt}
\centering
\caption{ASD classification results of BolT and VarCoNet on the ABIDE II dataset (mean $\pm$ standard deviation), along with the prediction change percentage on the Caltech subset of ABIDE I. Boldface indicates the best-performing method for each metric.}
\label{ASD_results_ABIDEII}
\resizebox{\columnwidth}{!}{\begin{tabular}{ccccc}
\hline \hline
Method & BCE Loss (\%) & AUC (\%) & F1-score (\%) & Pred. change (\%) \\ \hline

& \multicolumn{4}{c}{\multirow{1}{*}{{AAL}}}\\ \hline

BolT \cite{bedel2023bolt} & 1.243 $\pm$ 0.076 & 61.58 $\pm$ 0.77 & 52.53 $\pm$ 4.74 & 21.53 \\

VarCoNet              & 0.639 $\pm$ 0.007 & 68.42 $\pm$ 1.12 & 61.57 $\pm$ 2.17 & \textbf{11.40} \\ \hline

& \multicolumn{4}{c}{\multirow{1}{*}{{AICHA}}}\\ \hline

BolT \cite{bedel2023bolt} & 1.050 $\pm$ 0.050 & 67.28 $\pm$ 1.15 & 59.10 $\pm$ 3.40 & 23.96 \\

VarCoNet              & \textbf{0.609 $\pm$ 0.011} & \textbf{73.10 $\pm$ 1.30} & \textbf{65.01 $\pm$ 2.08} & 18.06 \\

\hline\hline 
\end{tabular}}
\end{table}

\autoref{ASD_results_ABIDEII} presents the performance of VarCoNet and its closest competitor, BolT, on 579 subjects from the independent ABIDE II dataset. Without fine-tuning, VarCoNet consistently outperforms BolT across both atlases, achieving a BCE loss nearly half that of BolT, along with AUC improvements of approximately 6.5\% (AAL3) and 5.5\% (AICHA), and F1-score gains of about 9\% (AAL3) and 6\% (AICHA). The much higher BCE loss of BolT indicates frequent high-confidence misclassifications, a critical weakness in medical applications where predictive reliability is essential. By contrast, VarCoNet maintains BCE loss values consistent with those observed in 10-fold CV and Caltech test evaluations. In addition, VarCoNet shows around 10\% (AAL3) and 6\% (AICHA) greater consistency when tested on different rs-fMRI signal segments, further underscoring its robustness.

\subsection{Ablations}
\begin{figure*}
    \centering
    \begin{subfigure}{0.32\textwidth}
        \centering
        \includegraphics[width=\textwidth]{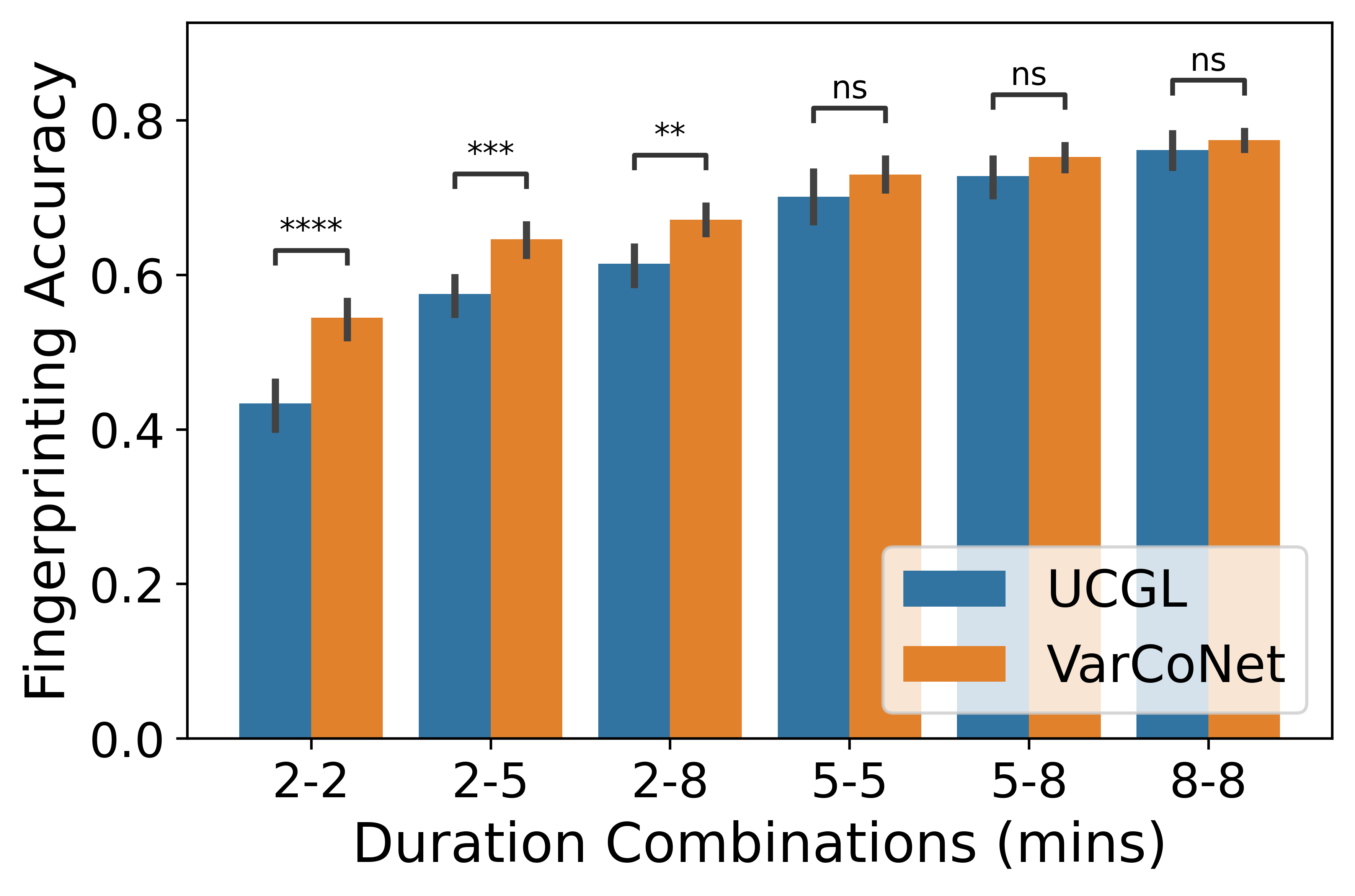}
        \caption{AAL3: Augmentation}
        \label{ablation_hcp_aal_augment}
    \end{subfigure}
    \centering
    \begin{subfigure}{0.32\textwidth}
        \centering
        \includegraphics[width=\textwidth]{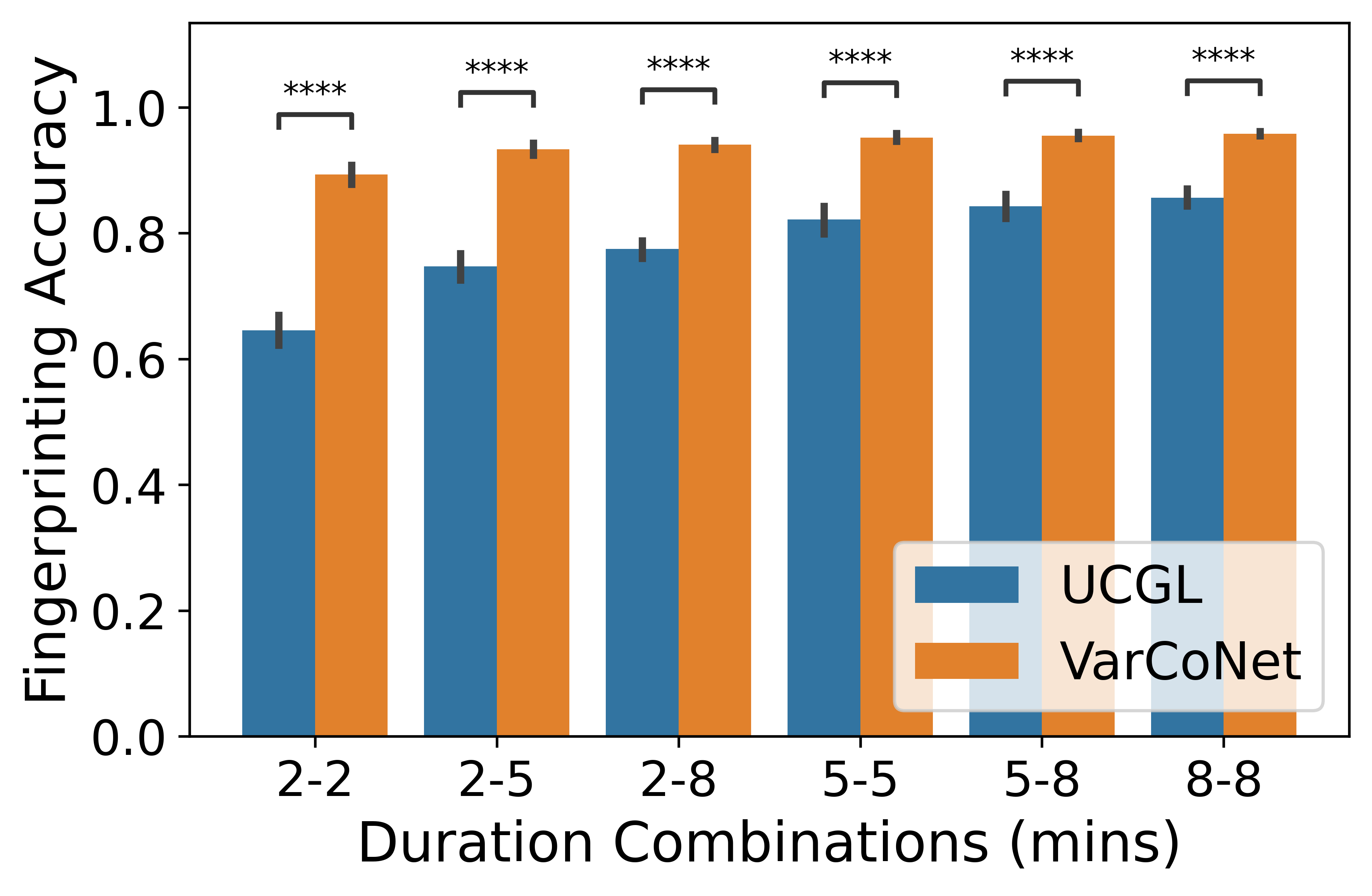}
        \caption{AICHA: Augmentation}
        \label{ablation_hcp_aicha_augment}
    \end{subfigure}
    \centering
    \begin{subfigure}{0.32\textwidth}
        \centering
        \includegraphics[width=\textwidth]{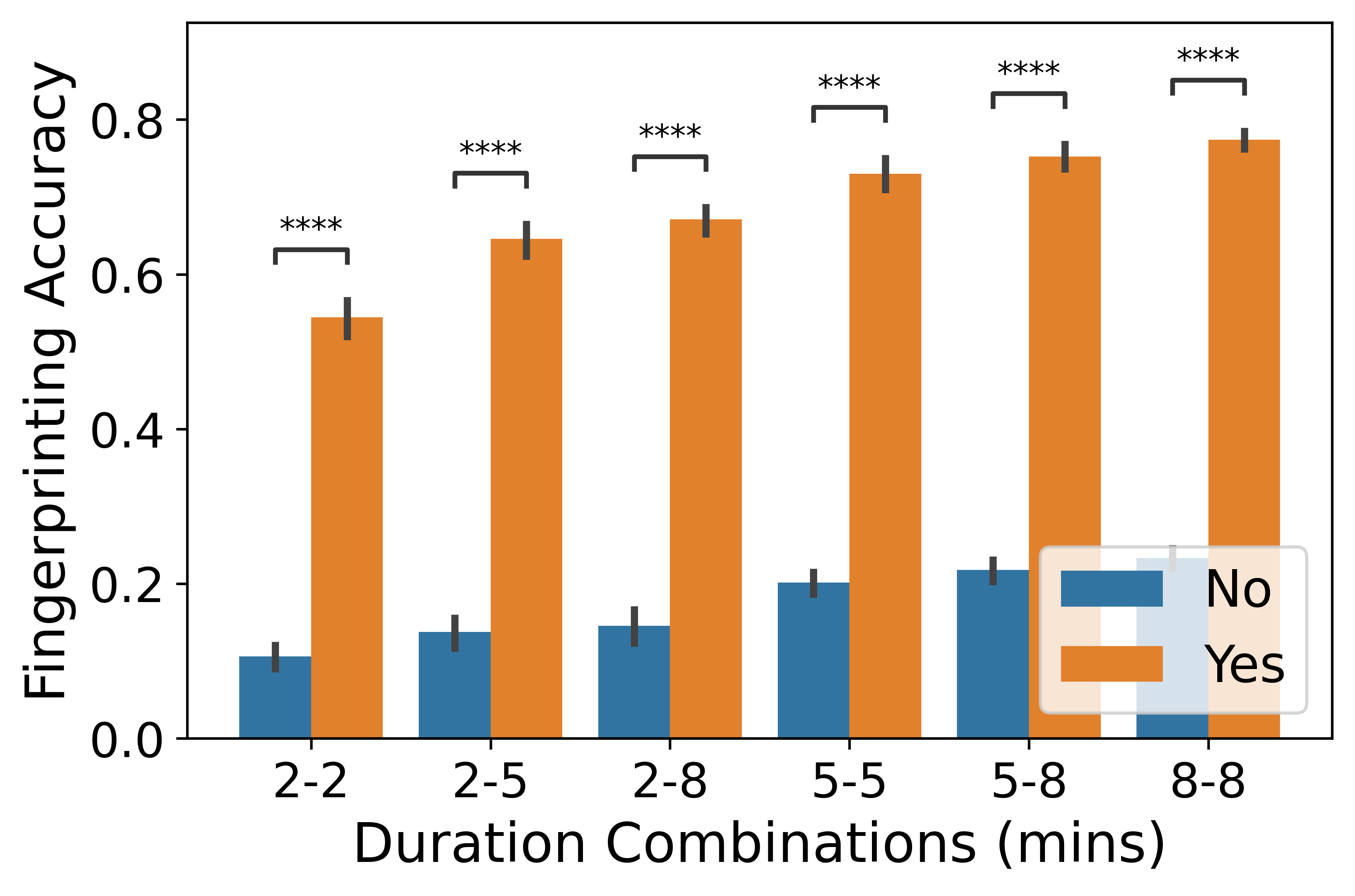}
        \caption{AAL3: Transformer}
        \label{ablation_hcp_aal_transformer}
    \end{subfigure}
    \centering
    \begin{subfigure}{0.32\textwidth}
        \centering
        \includegraphics[width=\textwidth]{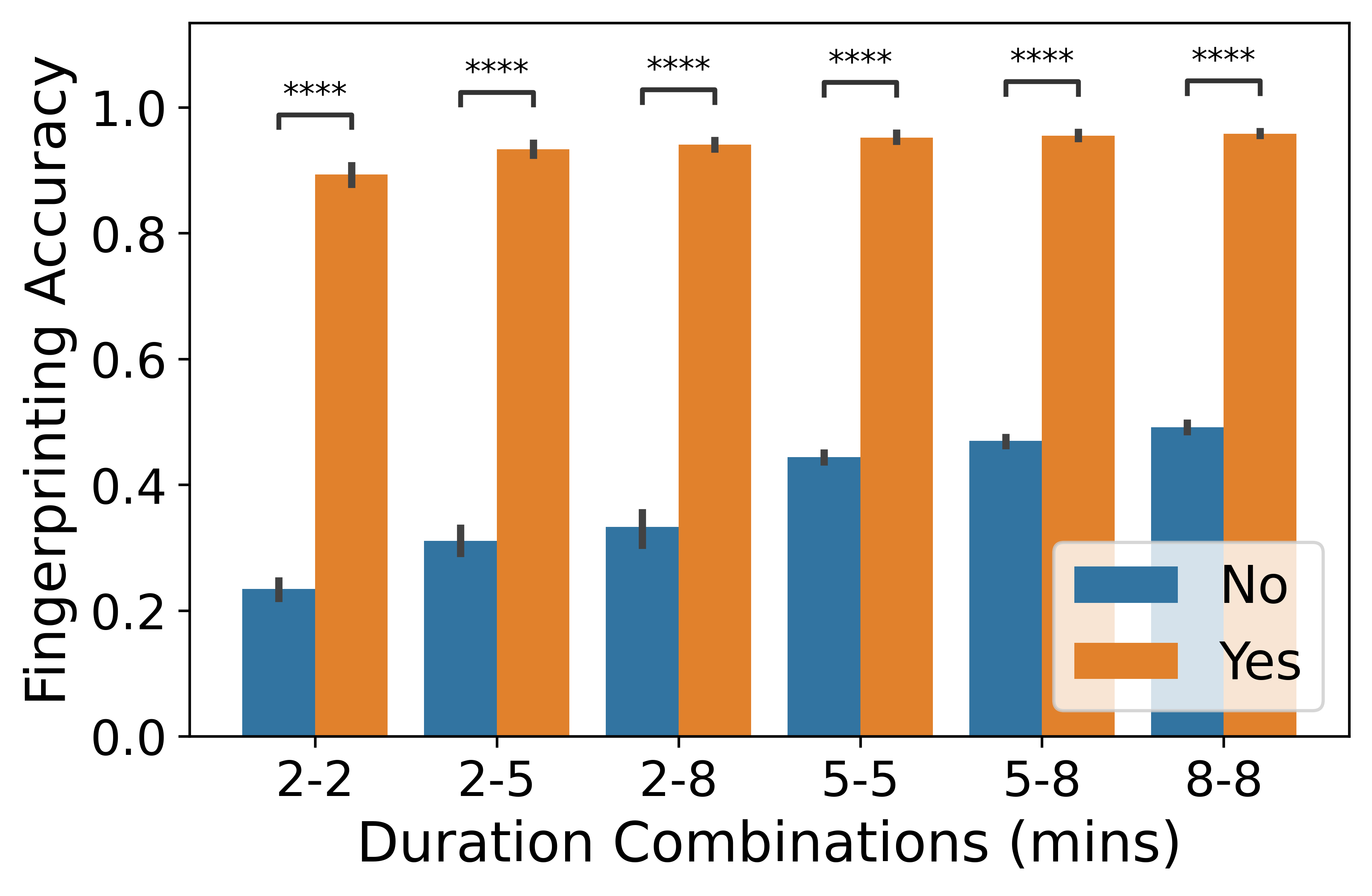}
        \caption{AICHA: Transformer}
        \label{ablation_hcp_aicha_transformer}
    \end{subfigure}
    \centering
    \begin{subfigure}{0.32\textwidth}
        \centering
        \includegraphics[width=\textwidth]{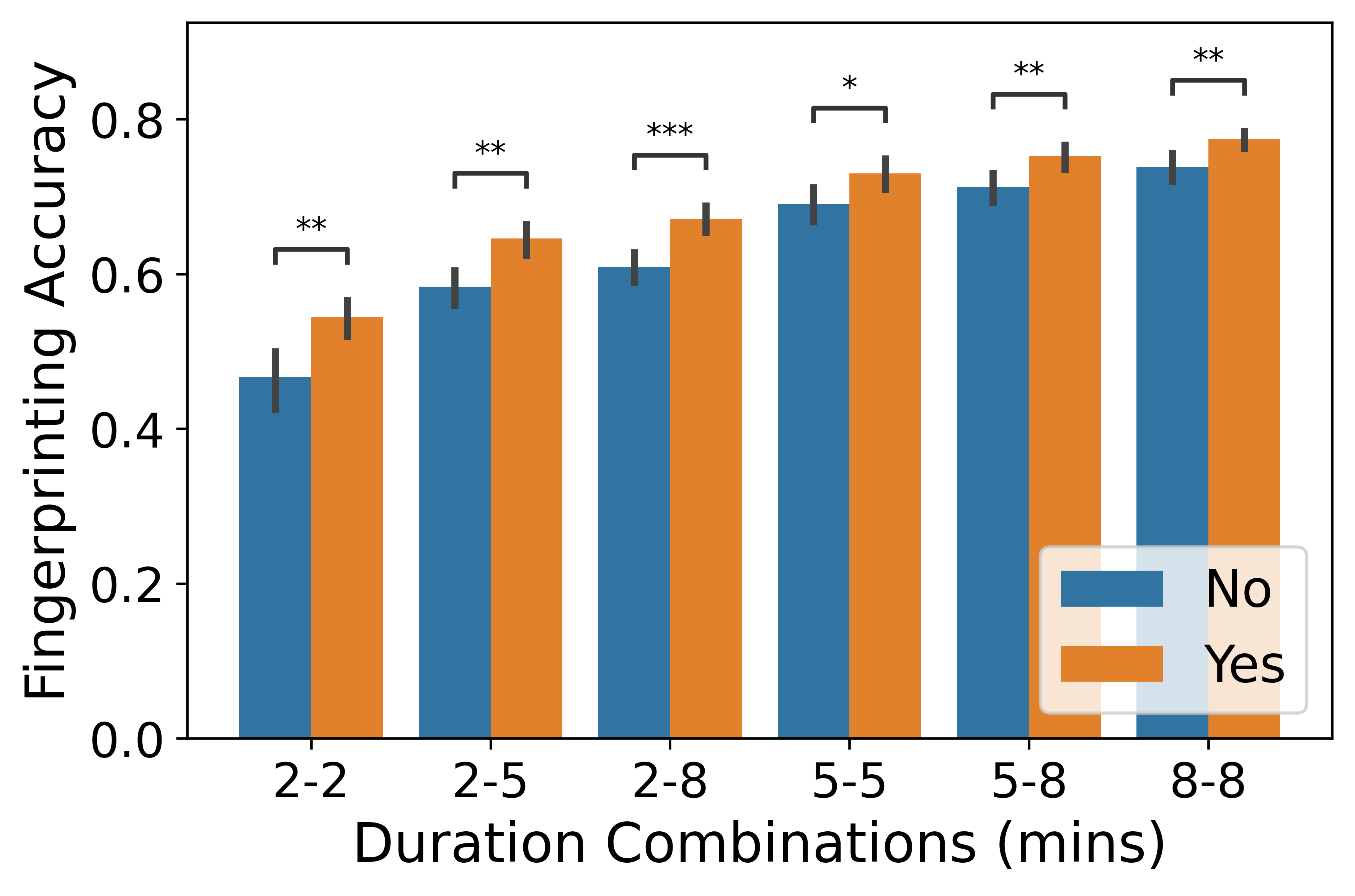}
        \caption{AAL3: 1D-CNN}
        \label{ablation_hcp_aal_1dcnn}
    \end{subfigure}
    \centering
    \begin{subfigure}{0.32\textwidth}
        \centering
        \includegraphics[width=\textwidth]{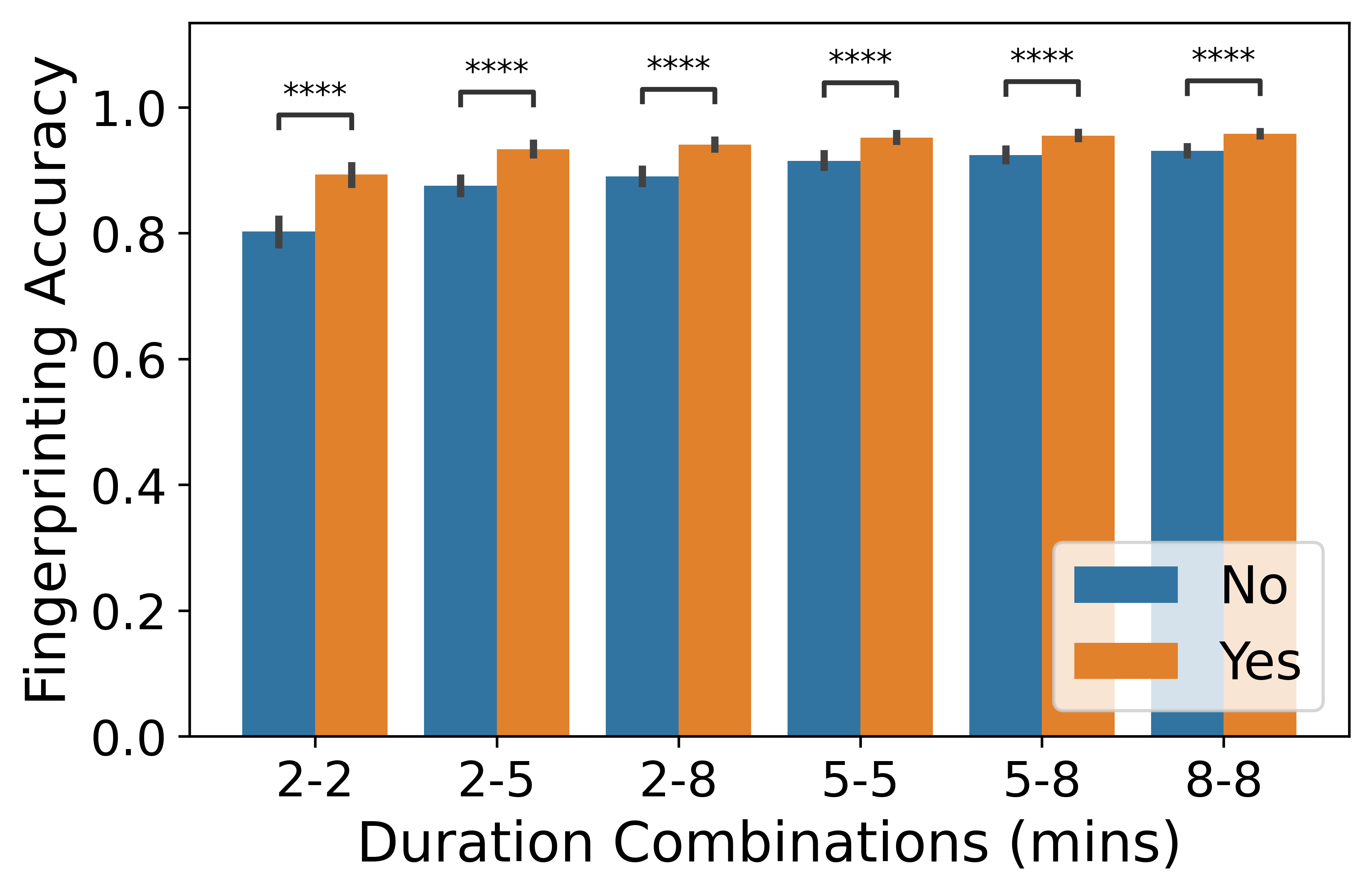}
        \caption{AICHA: 1D-CNN}
        \label{ablation_hcp_aicha_1dcnn}
    \end{subfigure}
    \caption{Ablation studies on HCP. (a,b) compare the proposed augmentation method with that of Wang \textit{et al.} \cite{wang2023unsupervised}; (c,d) evaluate the effect of removing the Transformer (yielding a 1D-CNN–only encoder); (e,f) assess the effect of removing the 1D-CNN (yielding a Transformer-only encoder). Asterisks above the bars indicate statistical significance: **** $p<0.0001$, *** $p<0.001$, ** $p<0.01$, * $p<0.05$, ns $p>0.05$.}
    \label{ablation_plots_hcp}
\end{figure*}

\begin{figure*}
    \centering
    \begin{subfigure}{0.24\textwidth}
        \centering
        \includegraphics[width=\textwidth]{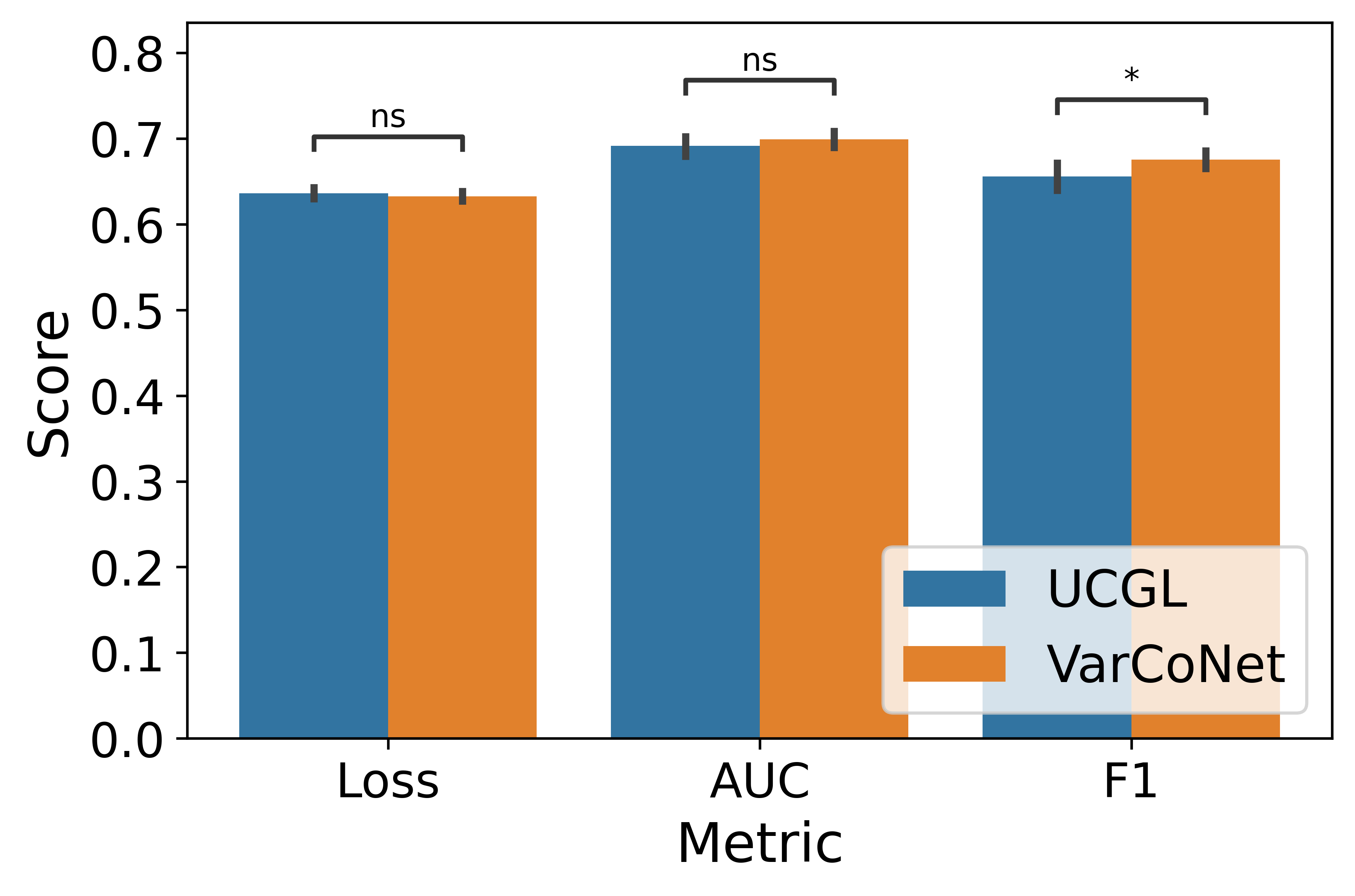}
        \caption{AAL3: Augmentation}
        \label{ablation_abide_aal_augment}
    \end{subfigure}
    \centering
    \begin{subfigure}{0.245\textwidth}
        \centering
        \includegraphics[width=\textwidth]{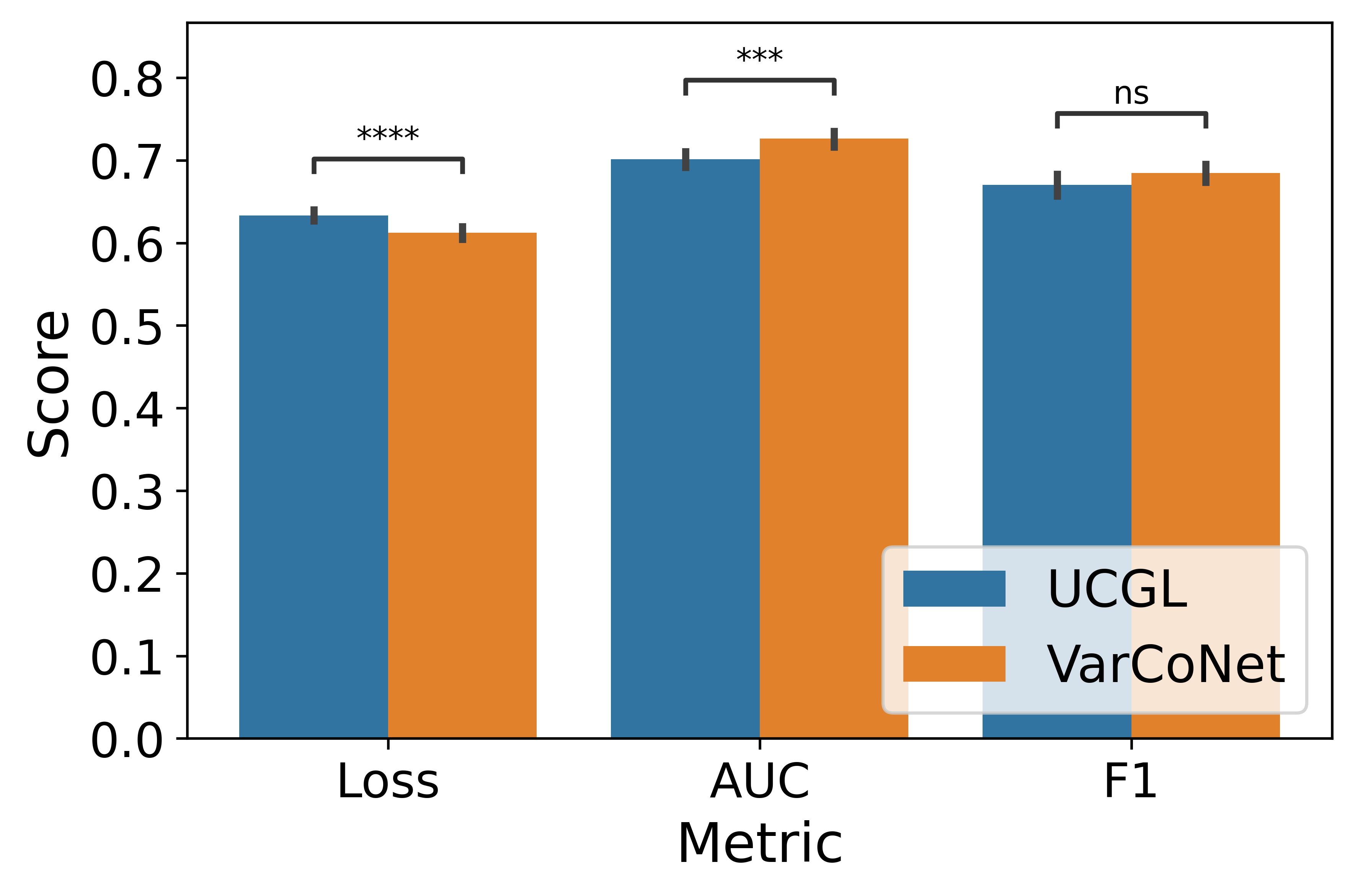}
        \caption{AICHA: Augmentation}
        \label{ablation_abide_aicha_augment}
    \end{subfigure}
    \centering
    \begin{subfigure}{0.245\textwidth}
        \centering
        \includegraphics[width=\textwidth]{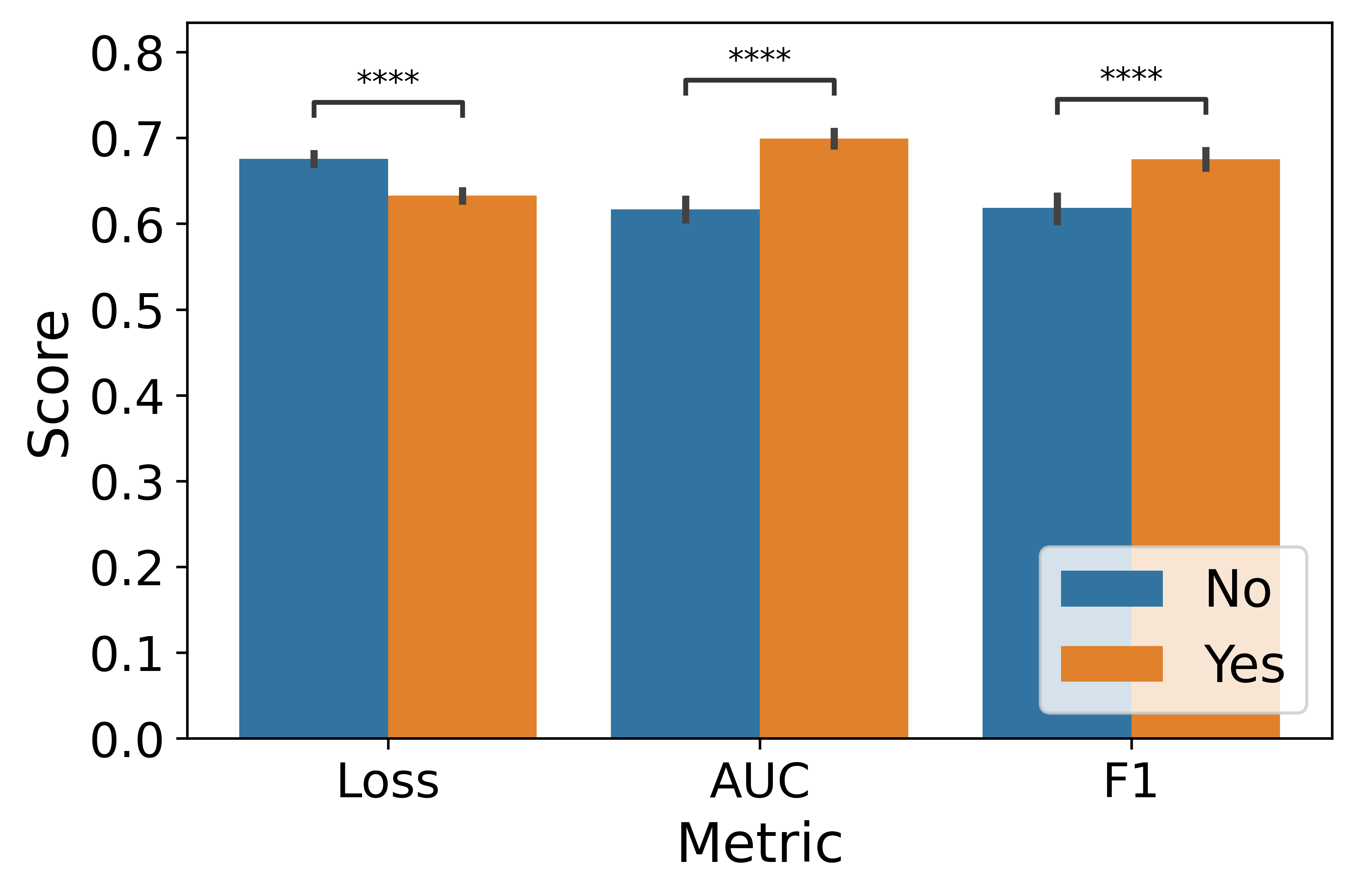}
        \caption{AAL3: Transformer}
        \label{ablation_abide_aal_transformer}
    \end{subfigure}
    \centering
    \begin{subfigure}{0.245\textwidth}
        \centering
        \includegraphics[width=\textwidth]{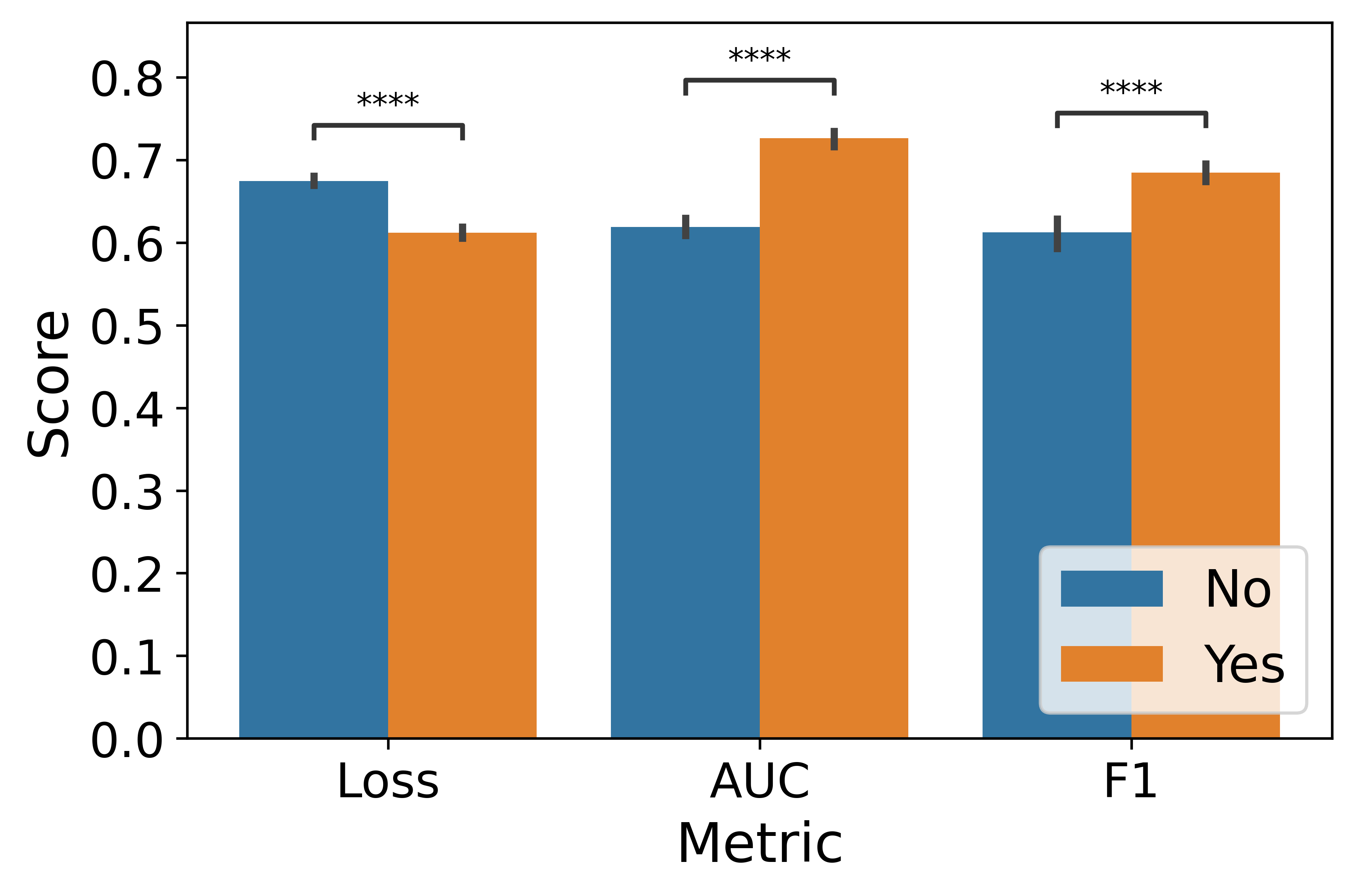}
        \caption{AICHA: Transformer}
        \label{ablation_abide_aicha_transformer}
    \end{subfigure}
    \centering
    \begin{subfigure}{0.245\textwidth}
        \centering
        \includegraphics[width=\textwidth]{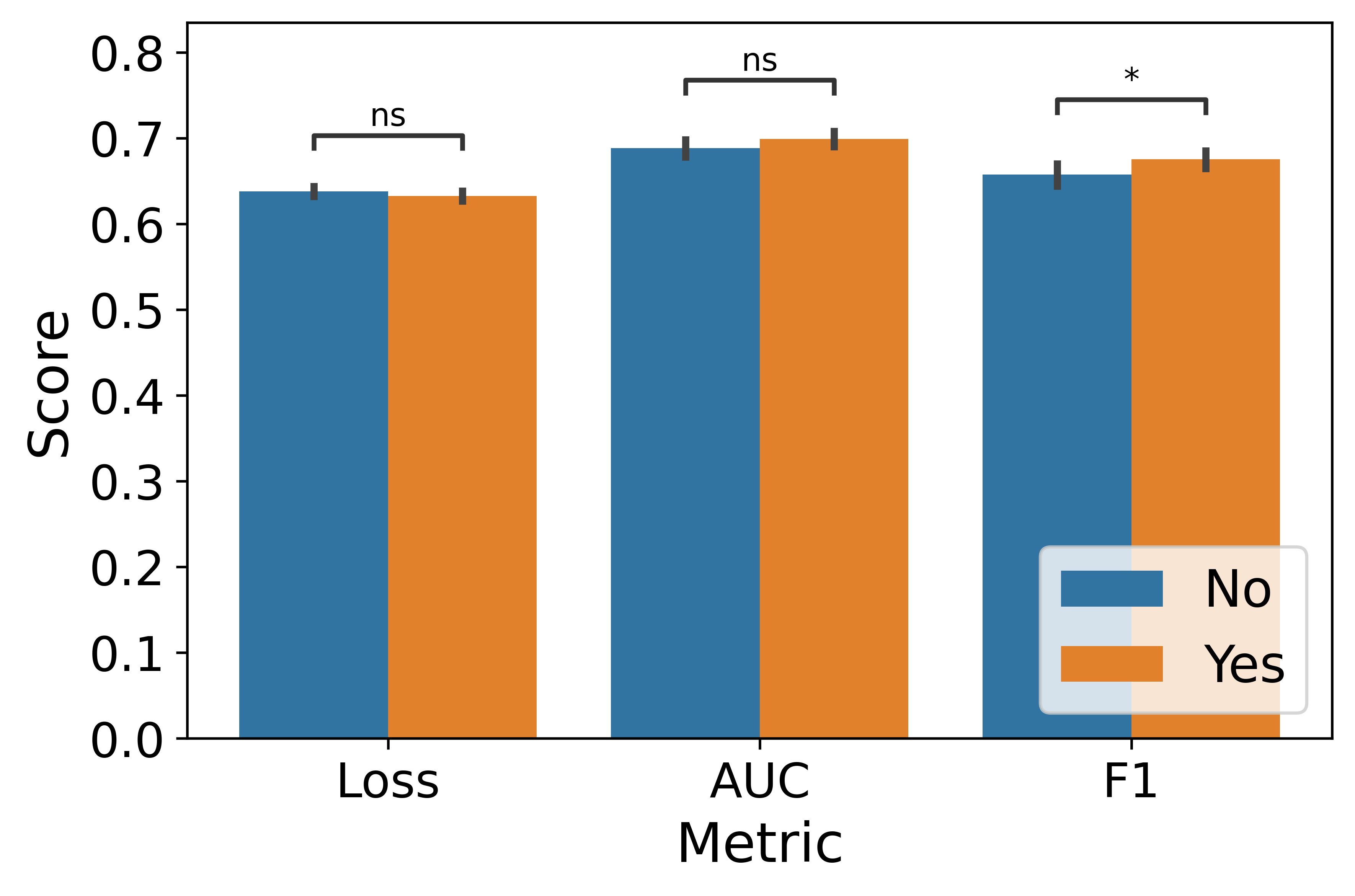}
        \caption{AAL3: 1D-CNN}
        \label{ablation_abide_aal_1dcnn}
    \end{subfigure}
    \centering
    \begin{subfigure}{0.245\textwidth}
        \centering
        \includegraphics[width=\textwidth]{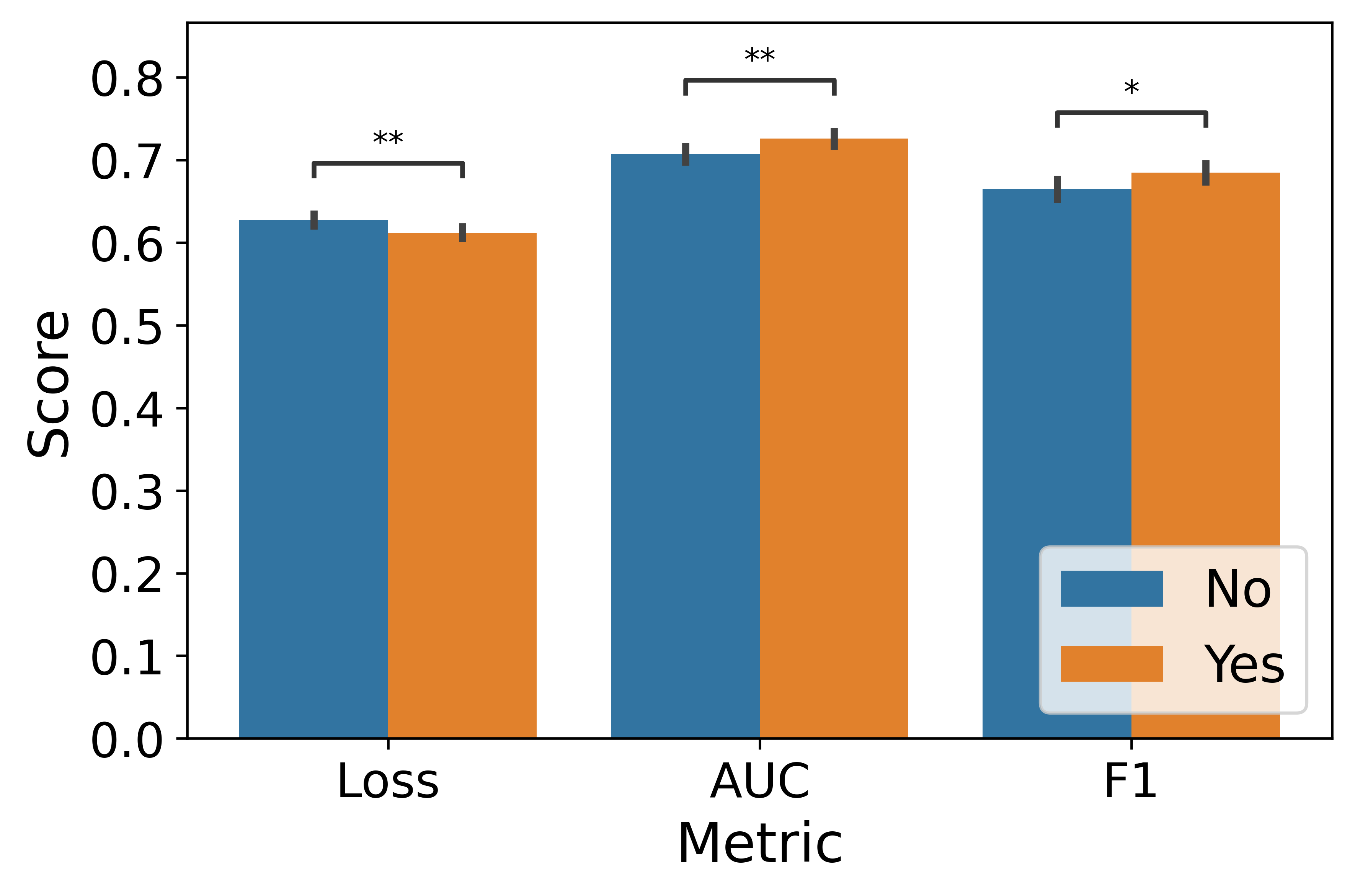}
        \caption{AICHA: 1D-CNN}
        \label{ablation_abide_aicha_1dcnn}
    \end{subfigure}
    \centering
    \begin{subfigure}{0.245\textwidth}
        \centering
        \includegraphics[width=\textwidth]{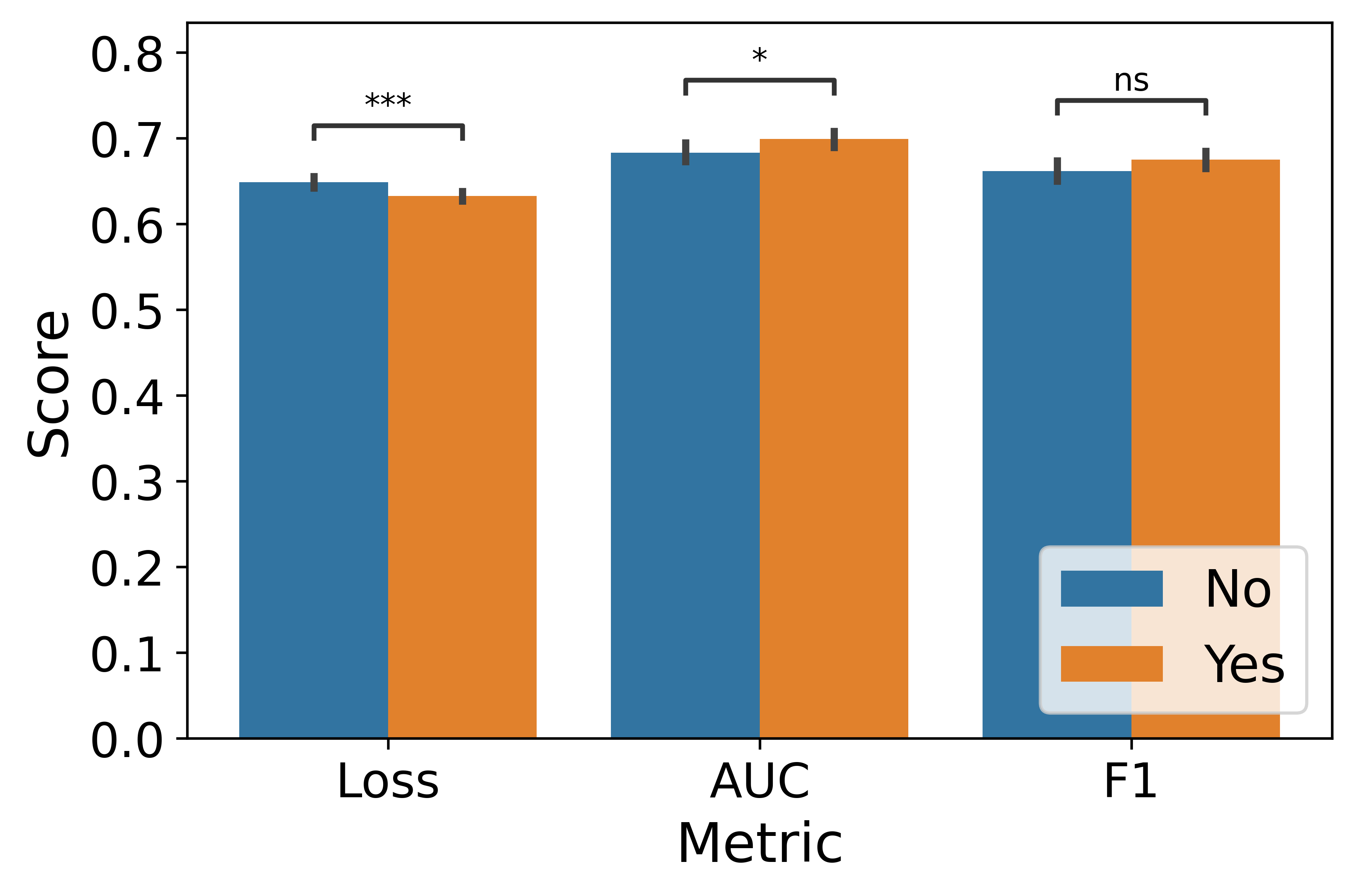}
        \caption{AAL3: SSL}
        \label{ablation_abide_aal_ssl}
    \end{subfigure}
    \centering
    \begin{subfigure}{0.245\textwidth}
        \centering
        \includegraphics[width=\textwidth]{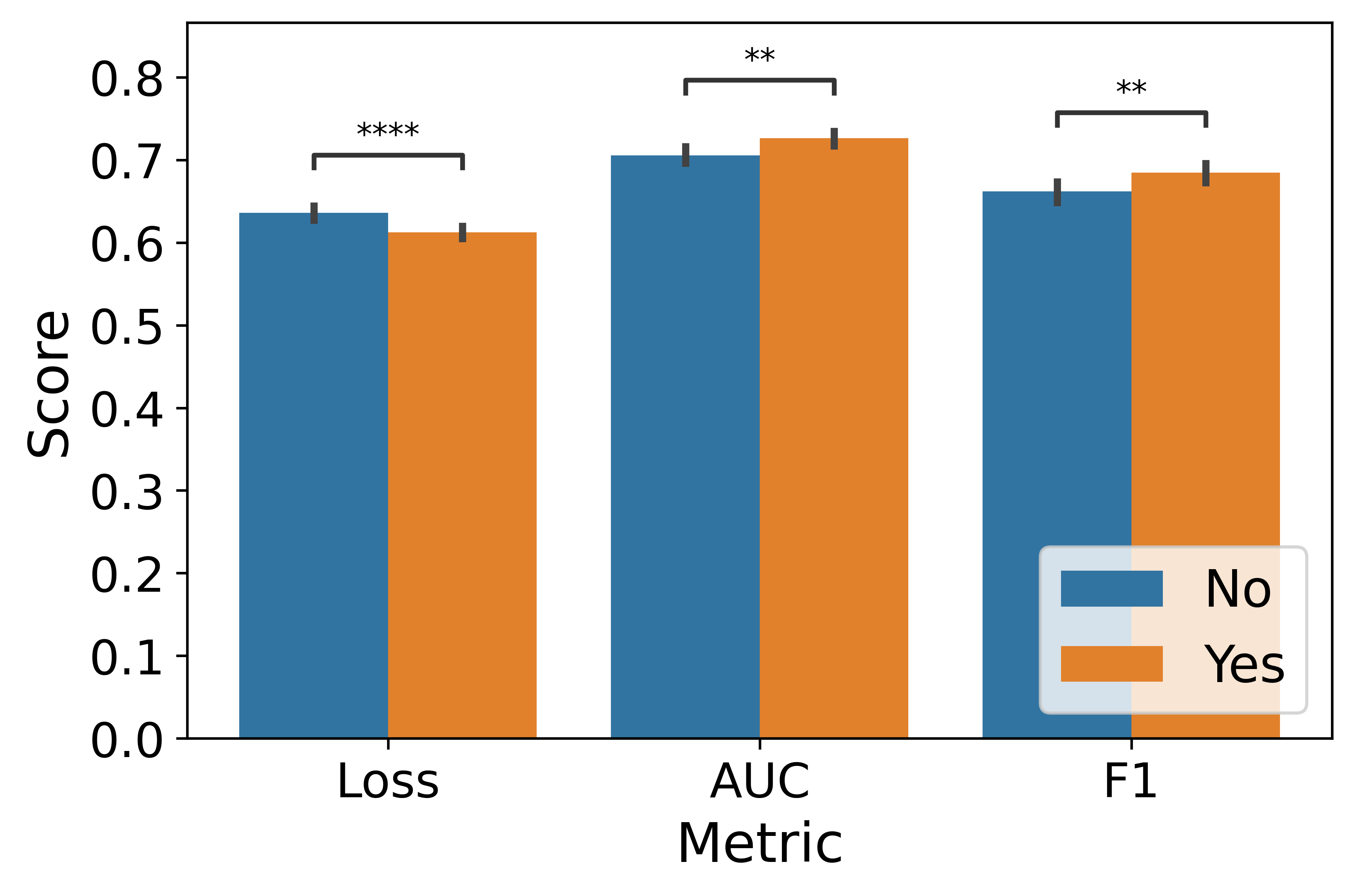}
        \caption{AICHA: SSL}
        \label{ablation_abide_aicha_ssl}
    \end{subfigure}
    \caption{Ablation studies on ABIDE I. (a,b) compare the proposed augmentation method with that of Wang \textit{et al.} \cite{wang2023unsupervised}; (c,d) assess the effect of removing the Transformer (1D-CNN-based encoder); (e, f) evaluate the effect of removing the 1D-CNN (Transformer-only encoder); (g, h) test supervised training of the 1D-CNN–Transformer encoder instead of using the VarCoNet framework. The asterisks above the bars indicate statistical significance (****: p-val$<$0.0001, ***: p-val$<$0.001, **: p-val$<$0.01, *: p-val$<$0.05, ns: p-val$>$0.05}
    \label{ablation_plots_abide}
\end{figure*}

\autoref{ablation_plots_hcp} and \autoref{ablation_plots_abide} show the ablation study results for the HCP and ABIDE I datasets, respectively. Detailed findings for each ablated parameter are provided below.

\subsubsection{Influence of Augmentation Method}
\autoref{ablation_hcp_aal_augment} and \autoref{ablation_hcp_aicha_augment} show the effect of replacing the proposed augmentation method with that of the UCGL framework \cite{wang2023unsupervised} on subject fingerprinting accuracy using the AAL3 and AICHA atlases, respectively. Likewise, \autoref{ablation_abide_aal_augment} and \autoref{ablation_abide_aicha_augment} present the corresponding results for ASD classification.

For subject fingerprinting, the proposed augmentation provides clear gains in accuracy across both atlases and all duration combinations. With the AICHA atlas, improvements are statistically significant for all durations, while with the AAL3 atlas, significance is observed for combinations involving the shortest duration (2 mins). The augmentation also benefits ASD classification: for the AAL3 atlas, all metrics improve, with a statistically significant gain in F1-score; for the AICHA atlas, all metrics improve, with significance in loss and AUC.

Overall, this ablation confirms the importance of the proposed augmentation scheme. By effectively handling variability in acquisition duration, it addresses the first research gap (“Inefficient handling of variability in acquisition duration,” see Section \ref{research_gaps}) and establishes augmentation as a core element of VarCoNet.

\subsubsection{Influence of Transformer}
\autoref{ablation_hcp_aal_transformer} and \autoref{ablation_hcp_aicha_transformer} show the effect of replacing the combined 1D-CNN-Transformer encoder with a 1D-CNN architecture (as in \cite{lamprou2025Improved} and \cite{kan2022fbnetgen}) on subject-fingerprinting accuracy using the AAL3 and AICHA atlases, respectively. Similarly, \autoref{ablation_abide_aal_1dcnn} and \autoref{ablation_abide_aicha_1dcnn} present the corresponding results for ASD classification.

For subject fingerprinting, incorporating the Transformer is critical, as it substantially increases accuracy across all duration combination settings. Significant gains are also observed in ASD classification, where the combined 1D-CNN–Transformer consistently outperforms the 1D-CNN baseline across all metrics and atlases, with results that are statistically significant. This ablation study directly addresses the second research gap (“Limited integration of local and global temporal features,” see Section \ref{research_gaps}), establishing the Transformer as an essential component of VarCoNet.

\subsubsection{Influence of 1D-CNN}
\autoref{ablation_hcp_aal_1dcnn} and \autoref{ablation_hcp_aicha_1dcnn} illustrate the effect of using a combined 1D-CNN–Transformer encoder, as opposed to a Transformer alone, on subject-fingerprinting accuracy with the AAL3 and AICHA atlases, respectively. The corresponding impact on ASD classification performance is shown in \autoref{ablation_abide_aal_1dcnn} and \autoref{ablation_abide_aicha_1dcnn}.

For subject fingerprinting, adding the 1D-CNN before the Transformer is particularly advantageous for shorter time courses (2–2, 2–5, and 2–8 minutes), where it yields substantial gains in identification accuracy. For ASD classification, the 1D-CNN also provides significant improvements: across all metrics when using the AICHA atlas, and specifically for F1-score when using the AAL3 atlas. Beyond accuracy, the 1D-CNN improves training and inference efficiency. By reducing the temporal dimension, it decreases the computational load of the Transformer by limiting the number of attention operations, which leads to faster training and inference and lower memory usage. Empirical observations further indicate that models incorporating the 1D-CNN converge more quickly and consistently.

In summary, inserting a 1D-CNN before the Transformer enhances performance while also improving efficiency and stability. This directly addresses the third research gap (“Inefficient attention computation on raw time-series,” see Section \ref{research_gaps}) and establishes the 1D-CNN as a key component of VarCoNet.

\subsubsection{Influence of contrastive SSL}
\autoref{ablation_abide_aal_ssl} and \autoref{ablation_abide_aicha_ssl} show the effect of training the 1D-CNN–Transformer encoder within the contrastive SSL-based VarCoNet framework, compared to conventional SL, for ASD classification with the AAL3 and AICHA atlases. Across both atlases and all metrics, VarCoNet consistently outperforms its supervised counterpart, with statistical significance except for the F1-score on the AAL3 atlas.

The superior performance is attributed to the richer embeddings produced by VarCoNet (i.e., VarCoNet-based FCs). These embeddings are not optimized solely for ASD classification but instead capture broader neurophysiological features that distinguish individuals, including, but not limited to, the presence of ASD. As a result, the learned representations are more generalizable and versatile, reflecting individual variability in brain connectivity.

Finally, by relying on SSL, VarCoNet can be extended to diverse downstream tasks beyond brain disorder classification without requiring task-specific retraining. These findings address the fourth research gap identified in this work (“Overfitting to task-specific labels,” see Section \ref{research_gaps}) and align with prior studies \cite{zagoruyko2016paying,ericsson2021well,karthik2021tradeoffs}.

\subsection{Interpretation}
\subsubsection{Subject fingerprinting interpretation} 
\begin{figure*}
    \centering
    \begin{subfigure}{0.325\textwidth}
        \includegraphics[width=\textwidth]{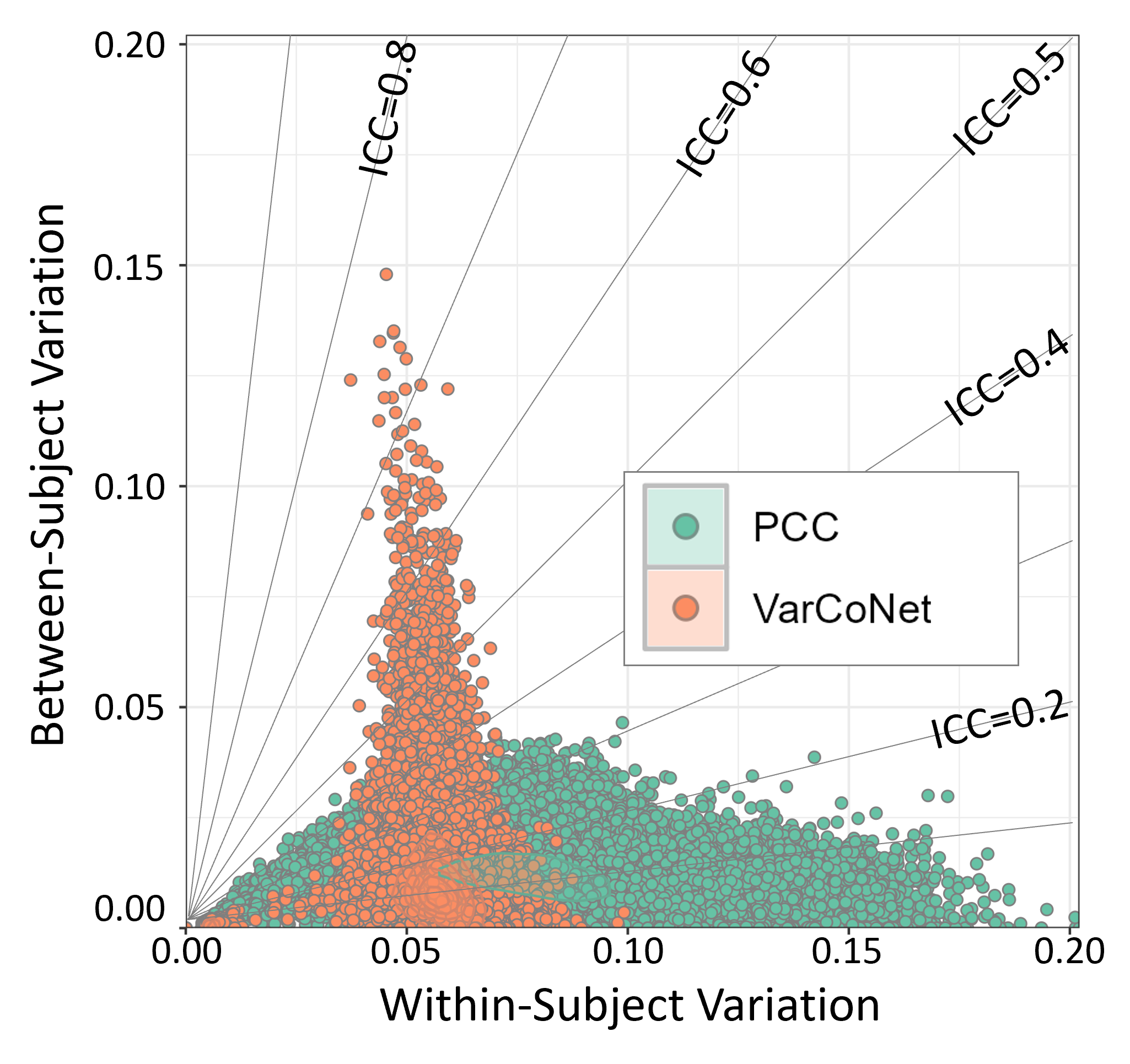}
        \caption{ICC: VarCoNet vs. PCC}
        \label{ICC_varconet_vs_pcc}
    \end{subfigure}
    \centering
    \begin{subfigure}{0.325\textwidth}
        \includegraphics[width=\textwidth]{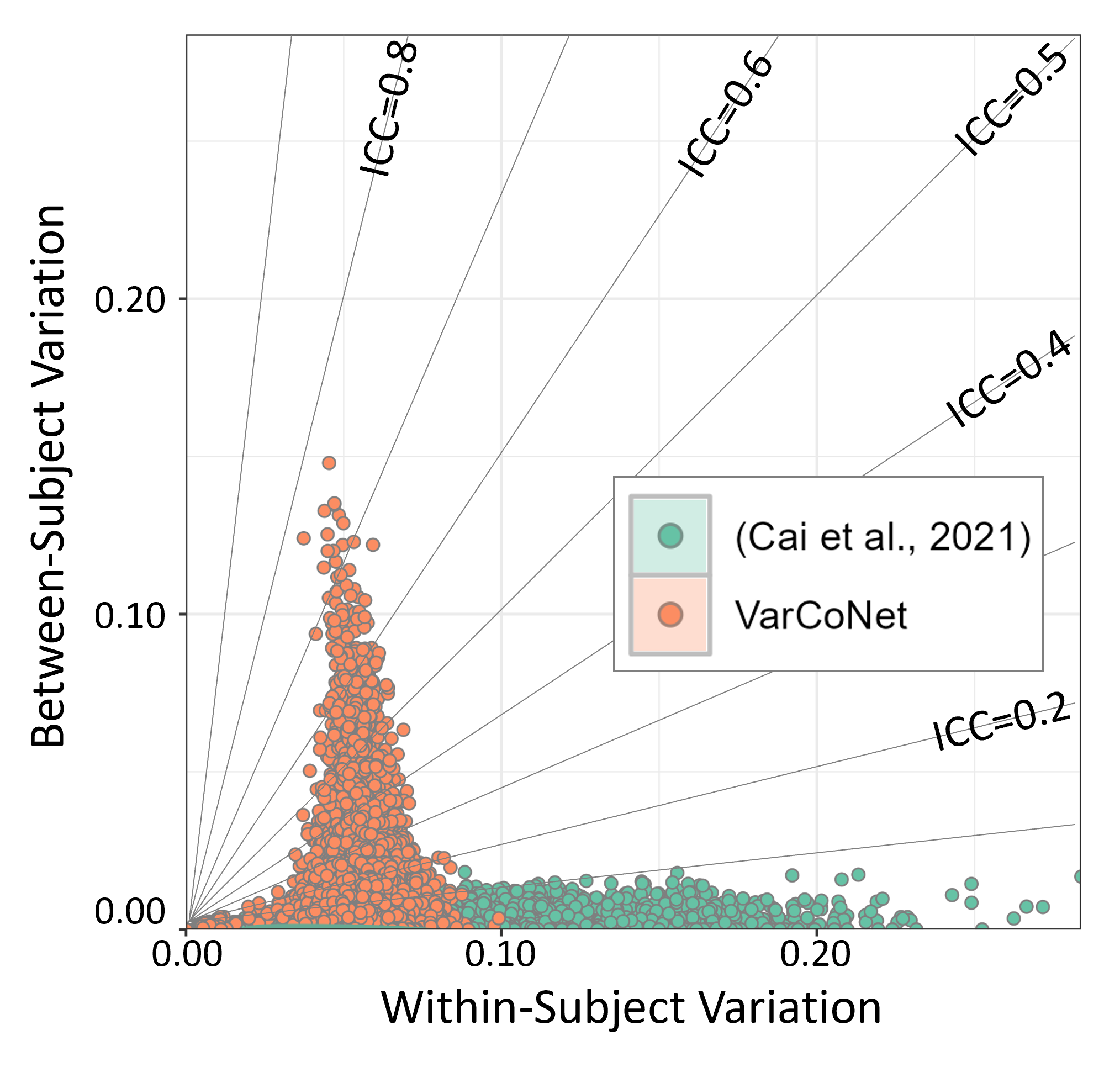}
        \caption{ICC: VarCoNet vs. \cite{cai2021functional}}
        \label{ICC_varconet_vs_ae}
    \end{subfigure}
    \centering
    \begin{subfigure}{0.325\textwidth}
        \includegraphics[width=\textwidth]{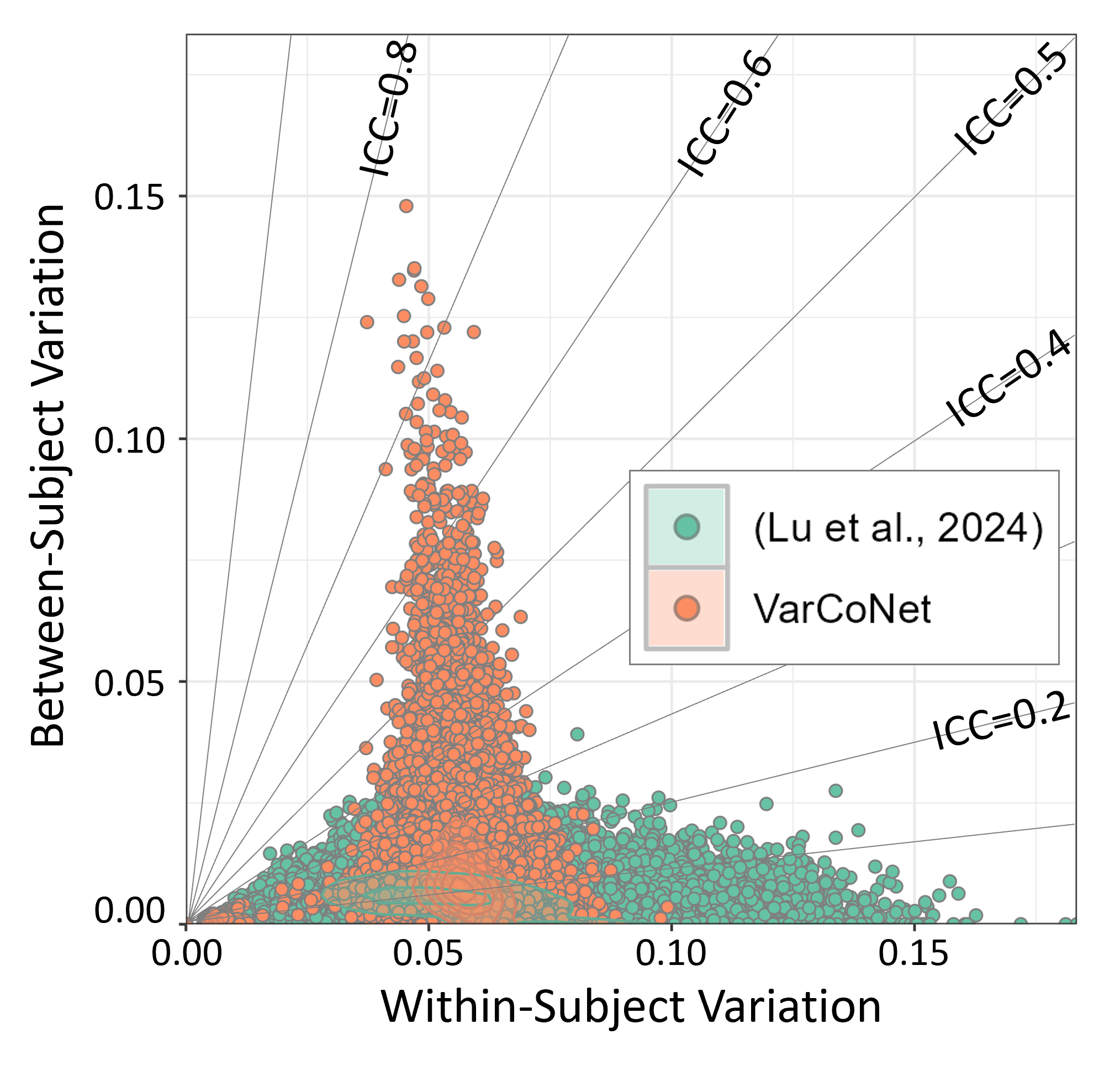}
        \caption{ICC: VarCoNet vs. \cite{lu2024brain}}
        \label{ICC_varconet_vs_vae}
    \end{subfigure}
    \centering
    \begin{subfigure}{0.325\textwidth}
        \includegraphics[width=\textwidth]{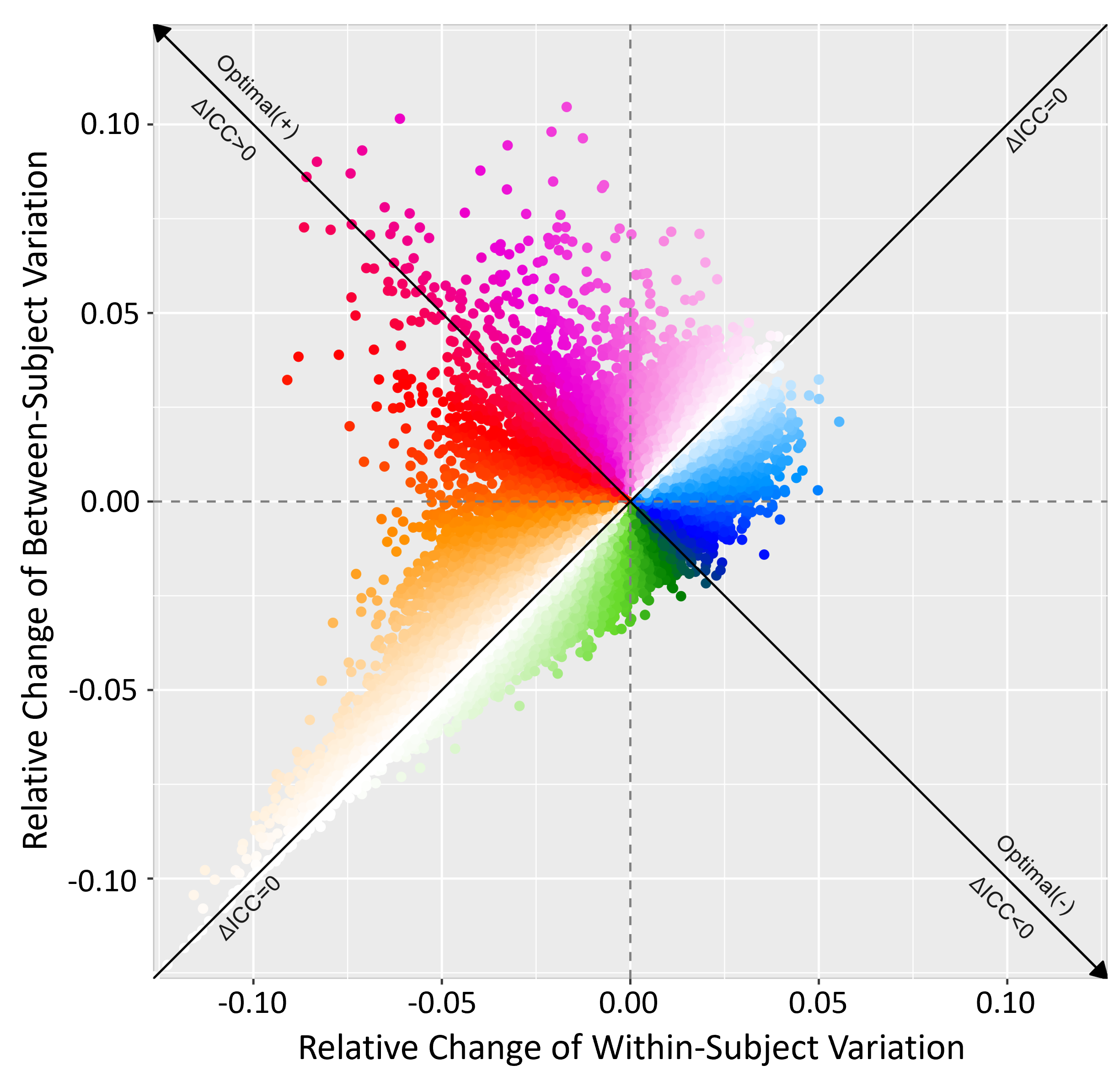}
        \caption{$\Delta$ICC: VarCoNet vs. PCC}
        \label{grad_varconet_vs_pcc}
    \end{subfigure}
    \centering
    \begin{subfigure}{0.325\textwidth}
        \includegraphics[width=\textwidth]{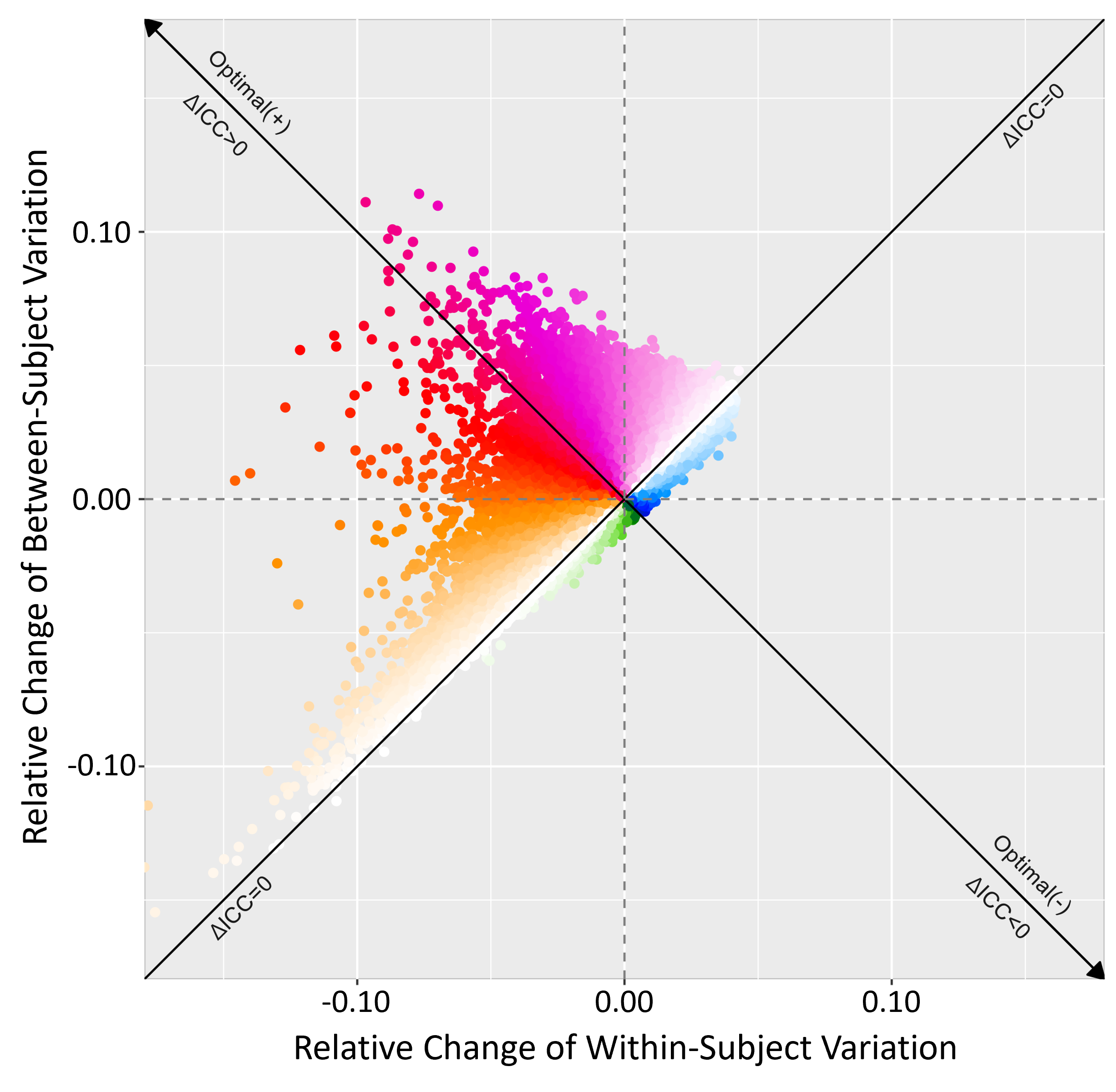}
        \caption{$\Delta$ICC: VarCoNet vs. \cite{cai2021functional}}
        \label{grad_varconet_vs_ae}
    \end{subfigure}
    \centering
    \begin{subfigure}{0.325\textwidth}
        \includegraphics[width=\textwidth]{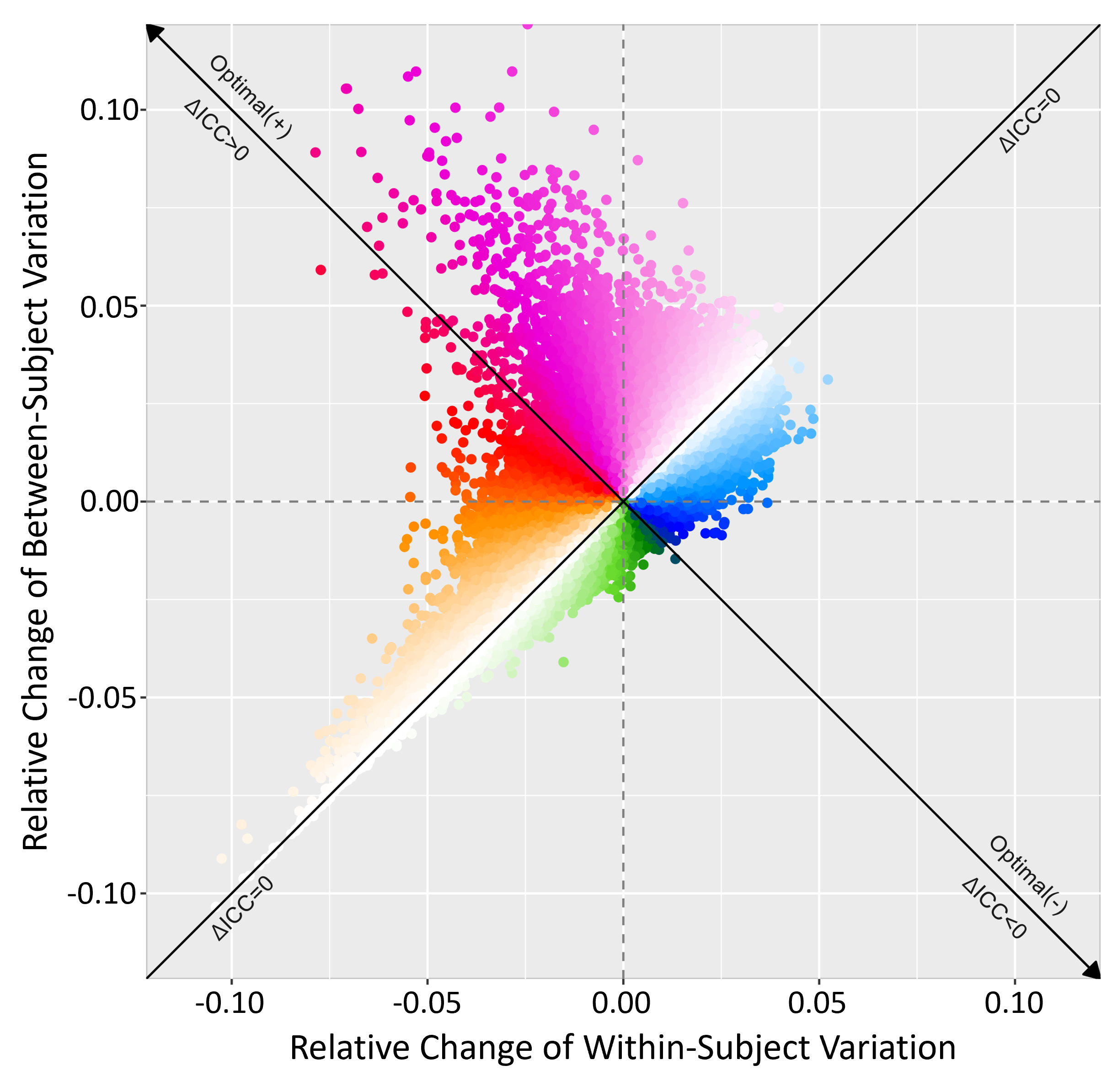}
        \caption{$\Delta$ICC: VarCoNet vs. \cite{lu2024brain}}
        \label{grad_varconet_vs_vae}
    \end{subfigure}
    \caption{Subject fingerprinting interpretation using the Rex toolbox. Panels (a)–(c) show the individual variation field and ICC of VarCoNet (VCN) compared to PCC, \cite{cai2021functional}, and \cite{lu2024brain}. Panels (d)–(f) illustrate the relative change in variation, normalized along the optimal direction, using PCC, \cite{cai2021functional}, and \cite{lu2024brain} as baselines and VarCoNet as the target method.}
    \label{rex_plots}
\end{figure*}

To better understand the factors underlying VarCoNet’s success in subject fingerprinting, we used the ReX toolbox \cite{rex}, which enables visual comparison of different methods in terms of intra-class correlation (ICC) and the gradient flow map of the individual variation field, key measures of methods' ability to capture individual differences. 

ICC typically quantifies the proportion of inter-individual variation that can be reliably captured. For example, an ICC of 0.5 indicates that the variable in question captures only 50\% of the inter-subject variation from an observed total that is confounded by within-subject variation. ICC can be improved by either reducing within-subject variation or increasing between-subject variation. Given a variable, such as the connectivity between two brain regions computed by two distinct methods across $N$ subjects with at least two sessions each, this field allows visualization of how methods differ in their ability to capture meaningful variation. One can then determine which method yields better ICC and whether improvements are driven by reduced intra-subject noise or enhanced inter-subject signal.

Moreover, given a baseline and a target method, the gradient flow map reveals whether switching methods improves the variable’s capacity to capture individual differences. Points in the lower right quadrant indicate a decline in ICC (higher intra-subject and lower inter-subject variation), while points in the upper left quadrant indicate an improvement (lower intra-subject and higher inter-subject variation). For the lower left and upper right quadrants, points above the diagonal reflect improvements in ICC, whereas points below the diagonal indicate deterioration.

To compare VarCoNet with PCC and the methods of \cite{cai2021functional, lu2024brain}, we use embeddings from all 393 subjects in the test set (with two sessions per subject), parcellated using the AICHA atlas. A segment length of 2 minutes is used for each session to create a challenging setting. Given the 384 regions in the AICHA atlas, this yields $(384 \times 383)/2 = 73,536$ unique connectivity variables.

\autoref{ICC_varconet_vs_pcc}, \autoref{ICC_varconet_vs_ae}, and \autoref{ICC_varconet_vs_vae} present the individual variation fields and ICC values comparing VarCoNet to PCC, \cite{cai2021functional}, and \cite{lu2024brain}, respectively. Across all comparisons, VarCoNet substantially reduces intra-subject variation while enhancing inter-subject variation, leading to a more favorable intra- to inter-subject variation ratio. These effects are more clearly observed in the corresponding gradient flow maps in \autoref{grad_varconet_vs_pcc}, \autoref{grad_varconet_vs_ae}, and \autoref{grad_varconet_vs_vae}. Across all three comparisons, the ICC consistently improves when switching to VarCoNet, as evidenced by the majority of points lying above the diagonal, indicating meaningful gains in capturing individual-specific signal.

\subsubsection{ASD classification interpretation}
\begin{figure}
    \centering
    \begin{subfigure}{0.22\columnwidth}
        \centering
        \includegraphics[width=\columnwidth]{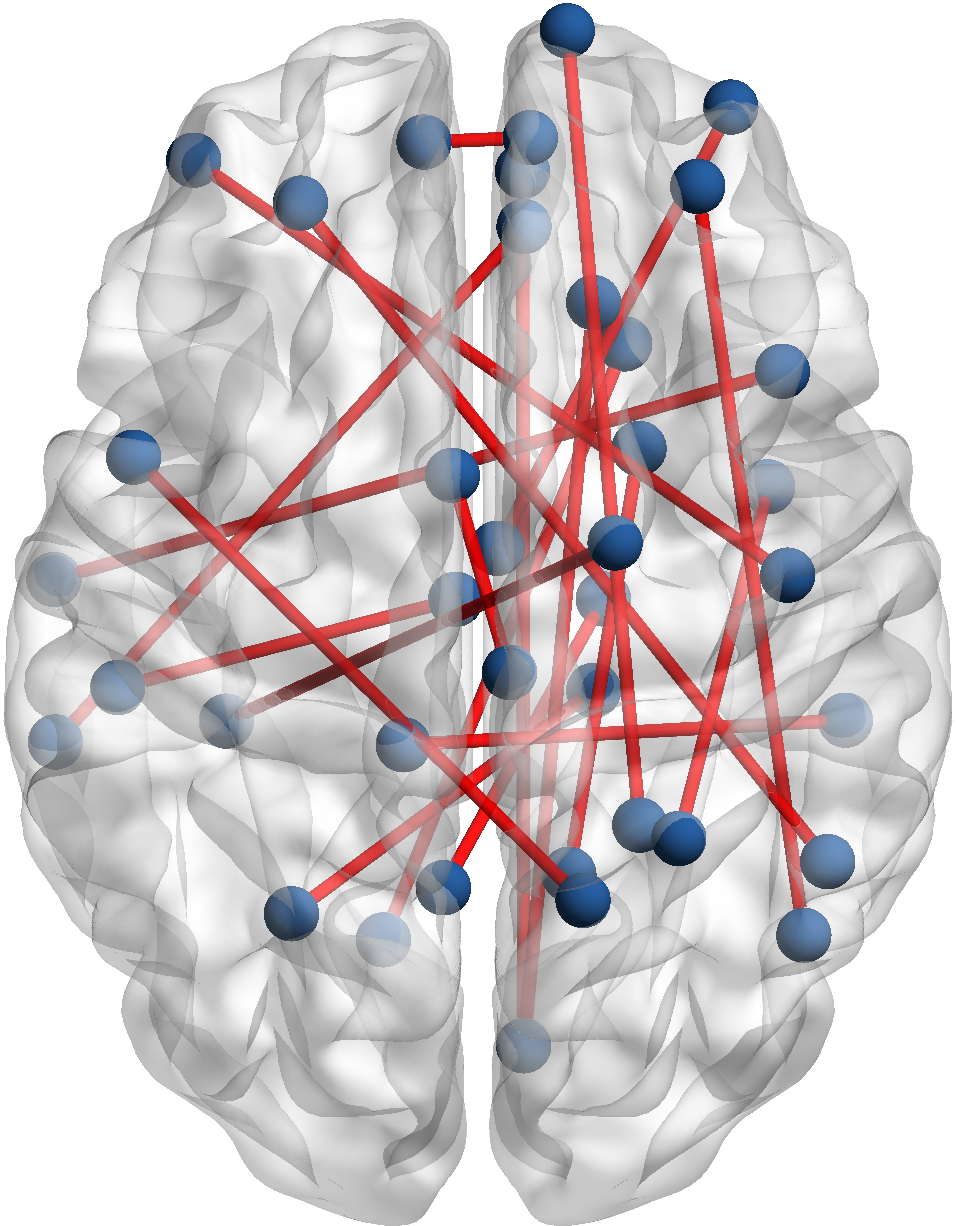}
        \caption{Axial}
        \label{aicha_asd_axial}
    \end{subfigure}
    \begin{subfigure}{0.30\columnwidth}
        \centering
        \includegraphics[width=\columnwidth]{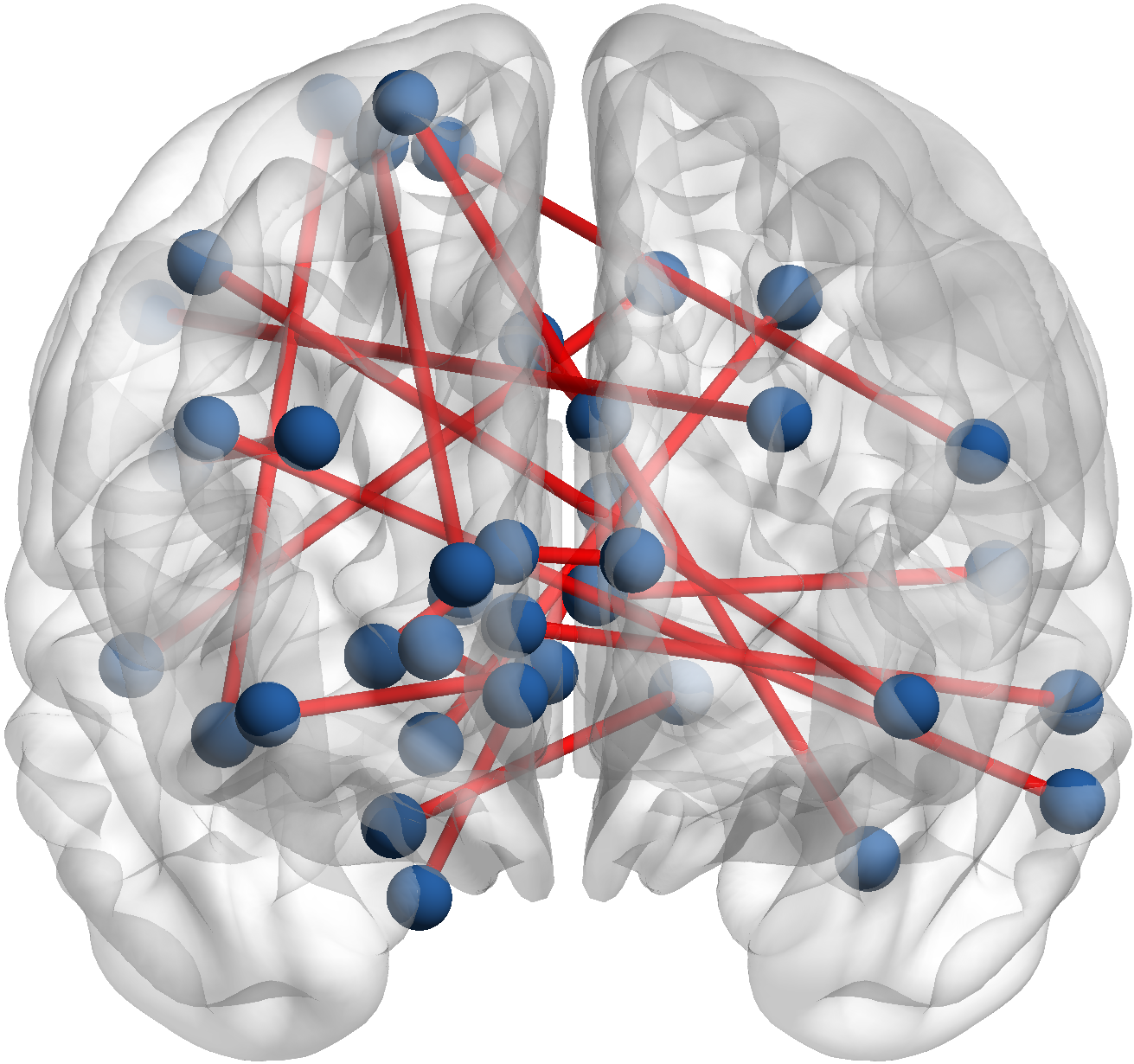}
        \caption{Coronal}
        \label{aicha_asd_coronal}
    \end{subfigure}
    \begin{subfigure}{0.36\columnwidth}
        \centering
        \includegraphics[width=\columnwidth]{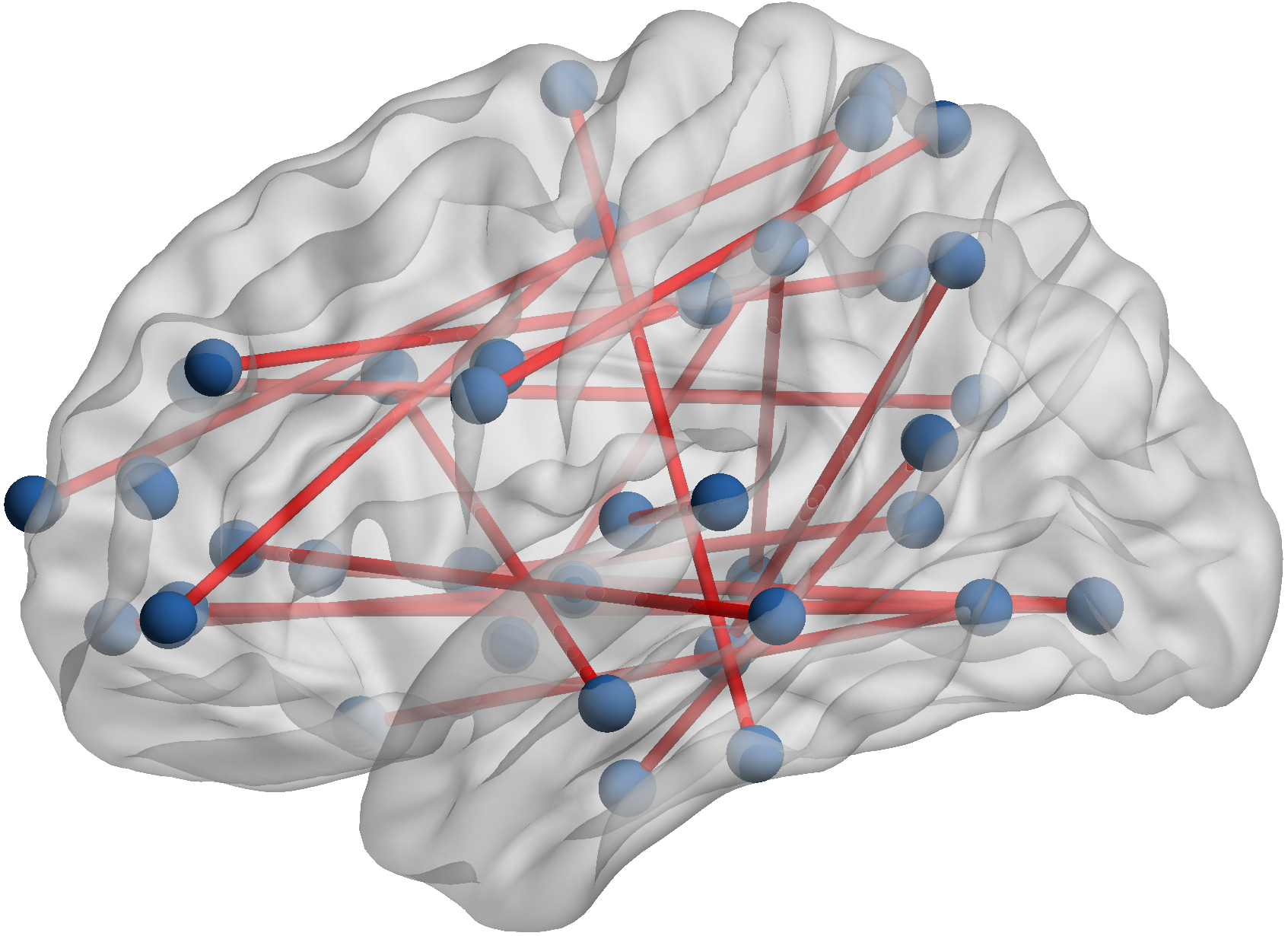}
        \caption{Sagittal}
        \label{aicha_asd_sagittal}
    \end{subfigure}
    \caption{Visualization of the most informative brain region connections for ASD classification. Panels (a)–(c) show the top 20 connections identified using the AICHA atlas, visualized in axial, coronal, and sagittal views, respectively.}
    \label{ASD_important_connections}
\end{figure}

We further investigate the brain connections most strongly associated with ASD based on VarCoNet's learned representations. The importance of each connection is derived from the weights of the linear classification layer. This layer has shape $(R \times (R - 1)/2, 2)$, representing connections between all pairs of regions and their contributions to the two output classes (NC and ASD). These weights are split into two vectors, $w_0$ and $w_1$, of shape $(R \times (R - 1)/2, 1)$, corresponding to the NC and ASD classes, respectively. Feature importance is then computed as $I = w_1 - w_0$, where a positive value indicates that a connection contributes toward classifying a sample as ASD, while a negative value favors the NC class. The absolute value of $I$ reflects the strength of that influence.

To ensure robust estimation, this importance score is computed for each of the 110 trained linear layers (100 from 10-fold CV repeated 10 times, and 10 from the external test set, also repeated 10 times). The scores are then averaged across all repetitions, and the top 20 most influential connections are selected based on their absolute importance values. These connections are visualized using BrainNet Viewer \cite{xia2013brainnet} for the best-performing atlas (AICHA), as shown in \autoref{ASD_important_connections}. Overall, the relevant connections for a successful distinction between a neurotypical and an ASD subject are in line with previous rs-fMRI studies, including functional connections between temporo-parietal regions involved in social cognition (e.g. \cite{63591,23675}) and other frontal cortices identified in previous neuroimaging studies \cite{1010412107,horien2022functional}.

\section{Discussion and conclusion}

\subsection{From contrastive learning to precision medicine}
While a few studies applied contrastive learning to extract embeddings from rs-fMRI data for ASD classification \cite{zhang2023gcl,peng2022gate}, our work presents the first empirical demonstration of the success of contrastive learning in diverse tasks. By design, contrastive learning aims to increase the similarity between augmented views of the same sample while decreasing similarity across samples, a concept that naturally aligns with intra- and inter-subject variability in brain function. In this context, subject fingerprinting serves as an ideal task to validate whether the model has effectively captured individual-specific patterns.

VarCoNet is a robust framework underpinned by the principle that accounting for inter-individual variability is key to achieving precision medicine \cite{seghier2018interpreting}. It leverages SimCLR-based contrastive learning, which inherently promotes decreased intra-subject variation and increased inter-subject variation, both essential for learning individualized representations. The model is first evaluated on subject fingerprinting to confirm that the VarCoNet-based FCs are meaningful. It is then assessed on ASD classification to test the central hypothesis that models which effectively capture inter-individual variability can also distinguish clinical conditions. Following a comprehensive analysis on several datasets, VarCoNet outperforms its supervised counterpart as well as several DL methods, providing strong empirical evidence for its robustness and versatility. 

\subsection{Performance of ASD baselines} \label{discussion_baselines}
As presented in Section \ref{ASD_results_section}, some competing methods performed poorly in ASD classification, with AUC scores close to chance level or underperforming relative to their original implementations. Several factors may explain these outcomes. First, for methods that did not provide codes \cite{riaz2020deepfmri, meng2024cvformer}, implementation discrepancies may have contributed to the reduced performance. Although considerable effort was made to accurately replicate the models based on the descriptions provided in the original publications, ensuring an exact replication is not straightforward given the large number of hyperparameters and implementation details that can significantly affect the performance of DL models.

Another possible factor is that some methods \cite{riaz2020deepfmri,bedel2023bolt,wang2023unsupervised,meng2024cvformer,wang2025self} were trained and tested only on single-site or two-site subsets of the ABIDE I dataset rather than on the entire dataset. Such subsets are often more "friendly" to DL models, as they exhibit less variability compared to multi-site data. This reduced variability can improve performance for models trained on a single site, as they do not need to generalize across the greater variability present in multi-site datasets.

Finally, the use of the ICA-AROMA pipeline \cite{pruim2015ica} for motion artifact removal may have influenced the performance gap between VarCoNet and competing methods. Unlike our approach, which employed this advanced procedure, none of the competing methods incorporated sophisticated motion correction in their original implementations. Motion is well known to degrade rs-fMRI data quality, while standard corrections such as 24-parameter regression or spike regression are often inadequate. This issue is particularly critical in ASD studies, as individuals with ASD typically exhibit higher motion during scanning than NCs. Consequently, models lacking advanced motion correction strategies, such as ICA-AROMA or ICA-FIX \cite{griffanti2014ica}, risk learning motion-related features rather than true neural patterns, thereby introducing bias into ASD classification results.

\subsection{VarCoNet's ASD classification performance}
The performance of VarCoNet in ASD classification is particularly notable given the heterogeneity of the datasets and the fact that it operates without label information during training. As highlighted by our ablation studies, VarCoNet’s contrastive SSL-based framework consistently outperforms traditional SL. While this might initially seem counterintuitive, prior research has demonstrated that SSL frameworks often outperform SL in various downstream tasks, including classification \cite{ericsson2021well, karthik2021tradeoffs}. This advantage stems from the tendency of SL models to overfit to label-specific features, a phenomenon known as "attentive overfitting" \cite{zagoruyko2016paying}, whereas SSL models are less susceptible to this issue \cite{ericsson2021well}.

From a clinical perspective, ASD represents a broad spectrum of deficit types and severity, and thus a single cutoff-based label may not adequately capture the heterogeneity of these deficits across individuals \cite{chapman2020representing}. This challenge can limit the effectiveness of SL models, which depend on clearly defined classes. In contrast, VarCoNet employs a contrastive learning approach focused on distinguishing individual subjects, enabling it to learn subtle differences across various factors such as age, cognitive behavior, and the presence of disorders like ASD. The VarCoNet-based FCs thus provide an ASD-aware representation, which is further refined by the linear layer for classification. This approach enhances the model’s ability to robustly distinguish between ASD and NC.

\subsection{Implication, limitations and future work}
Our work has several important implications. First, due to its SSL design, VarCoNet provides a general framework for extracting robust FC from rs-fMRI data without relying on labels. This is particularly valuable in medicine, where collecting large volumes of labeled rs-fMRI data is often difficult, expensive, and requires expert annotation. VarCoNet addresses this challenge by enabling training on unlabeled rs-fMRI data and subsequent application to downstream tasks with minimal or no fine-tuning. To support this, we release our trained model weights and implementation code. Second, as our results show, VarCoNet is significantly more robust than both conventional PCC and other DL methods when processing short rs-fMRI time courses. This makes it especially promising for time-varying FC (TVFC) analysis, where FC is computed over short windows—sometimes as brief as 30 seconds \cite{allen2014tracking}. Reliable TVFC is essential for accurate brain state estimation, which enhances our understanding of dynamic functional coordination, behavioral shifts, and adaptive processes. Finally, VarCoNet's resilience to short rs-fMRI durations makes it well-suited for scenarios such as those described by \cite{fair2007method}, where brief "interleaved" resting blocks within task-based fMRI are repurposed for resting-state analysis. Since the residual signal following removal of task-related segments is often limited, VarCoNet’s ability to extract meaningful FC from short time series enables effective use of these datasets.

We acknowledge two limitations in the current implementation of our framework. First, although VarCoNet outputs FC matrices that can highlight important connections between brain regions, it still faces inherent issues associated with interpreting DL models applied to rs-fMRI signals, namely, their complexity, non-stationarity, existing artifacts, high dimensionality, and lack of meaning (unlike, for example, task-induced variations in fMRI \cite{seghier2009dissociating}. These issues hinder direct human interpretability. The use of Transformer architectures compounds this issue, as the underlying attention mechanisms are more opaque and harder to intuitively interpret than more straightforward models such as 1D-CNNs. The second limitation is that VarCoNet was implemented to process rs-fMRI data with a specific repetition time (TR) of 1.5 seconds. Standardizing all input data to this TR via resampling was needed because Transformers require input tokens of consistent temporal meaning. However, this resampling may introduce small interpolation errors or slightly alter the temporal properties of fMRI signals acquired at TRs very different from our selected TR of 1.5 seconds.   

Future work will focus on integrating VarCoNet with GNNs, leveraging the graph-like structure of its output. We also plan to incorporate a trainable binary mask that selectively discards irrelevant or noisy connections, enabling the model to concentrate on the most informative features. The ultimate goal is to build an end-to-end framework trained with our contrastive learning approach, which takes parcellated rs-fMRI time-series as input, extracts robust, individualized FCs, and processes them with GNNs to generate embeddings suitable for downstream tasks such as subject fingerprinting and brain disorder classification. This integration is expected to enhance the model’s ability to capture higher-order interactions between brain regions while improving interpretability through the learned  mask. Additionally, future efforts will aim to eliminate the need for resampling input data to a fixed TR. Potential solutions include incorporating TR as an explicit input to the model or designing multiple 1D-CNNs with varied kernel sizes to ensure consistent temporal interpretation across different sampling rates.

\section{Conclusion}
This study presents VarCoNet, a contrastive self-supervised learning framework designed to extract robust functional connectomes in the light of the existing functional inter-individual variability. VarCoNet integrates a 1D-CNN–Transformer encoder, optimized via Bayesian search, for advanced processing of rs-fMRI time-series, along with a refined augmentation strategy that favors the learning of meaningful representations. Extensive evaluations on the HCP, ABIDE I, and ABIDE II datasets show that VarCoNet consistently outperforms prior methods in both subject fingerprinting and autism spectrum disorder (ASD) classification. In-depth ablation studies confirm the model’s robustness and stability, while interpretability analyses identify brain region connections linked to enhanced discrimination between individuals with ASD and neurotypicals. Taken together, these findings underscore the importance of modeling inter-individual variability in functional connectomes to advance precision medicine. By emphasizing subject-specific connectivity patterns, VarCoNet can serve as a robust and versatile rs-fMRI encoder for diverse identification and classification tasks.

\section*{Acknowledgments}
This study was supported by Khalifa University of Science and Technology (Grant number RC2-2018-022).

\bibliographystyle{IEEEtran}
\bibliography{main}

\end{document}